\documentclass[10pt,twocolumn,letterpaper]{article}

\usepackage{iccv}
\usepackage{times}
\usepackage{epsfig}
\usepackage{graphicx}
\usepackage{amsmath}
\usepackage{amssymb}
\usepackage[pagebackref=true,breaklinks=true,letterpaper=true,colorlinks,bookmarks=false]{hyperref}

\usepackage{color,xcolor}
\usepackage{epsfig}
\usepackage{graphicx}
\usepackage{times}

\usepackage{comment}

\usepackage{pifont}

\usepackage{microtype}
\usepackage[font=small]{caption}
\usepackage{arydshln}
\usepackage{tabularx}
\usepackage{adjustbox}
\usepackage{array}
\usepackage{booktabs}
\usepackage{colortbl}
\usepackage{float,wrapfig}
\usepackage{hhline}
\usepackage{multirow}
\usepackage{subcaption} %
\usepackage[percent]{overpic}

\usepackage{stmaryrd}
\usepackage{breqn}

\usepackage{amsmath,amsfonts,amssymb}
\usepackage{duckuments}
\usepackage{amsmath}

\usepackage{bm}
\usepackage{nicefrac}
\usepackage{microtype}
\usepackage{dsfont}
\usepackage{changepage}
\usepackage{extramarks}
\usepackage{fancyhdr}
\usepackage{setspace}
\usepackage{soul}
\usepackage{xspace}
\usepackage{hhline}
\usepackage{algorithmicx}
\usepackage{algorithm}
\usepackage{algpseudocode}
\usepackage{pifont}
\usepackage{amssymb}
\usepackage{booktabs}
\usepackage{multirow}

\usepackage{makecell}

\usepackage{enumitem}
\usepackage[title]{appendix}

\def\x{{\mathbf{x}}}

\def\c{{\mathbf{c}}}
\def\t{{\mathbf{t}}}

\newcommand{\smallsim}{\smallsym{\mathrel}{\sim}\hspace{-.5mm}}

\makeatletter
\newcommand{\smallsym}[2]{#1{\mathpalette\make@small@sym{#2}}}
\newcommand{\make@small@sym}[2]{%
  \vcenter{\hbox{$\m@th\downgrade@style#1#2$}}%
}
\newcommand{\downgrade@style}[1]{%
  \ifx#1\displaystyle\scriptstyle\else
    \ifx#1\textstyle\scriptstyle\else
      \scriptscriptstyle
  \fi\fi
}
\makeatother

\newcommand*{\menlo}{\fontfamily{lmtt}\fontsize{9}{9}\selectfont }

\newcommand{\nupur}[1]{#1}

\newcommand{\anchor}{anchor }

\newcommand{\reffig}[1]{Figure~\ref{fig:#1}}
\newcommand{\refsec}[1]{Section~\ref{sec:#1}}
\newcommand{\refapp}[1]{Appendix~\ref{sec:#1}}
\newcommand{\reftbl}[1]{Table~\ref{tbl:#1}}

\newcommand{\refeq}[1]{Eqn.~\ref{eq:#1}}

\newcommand{\lblfig}[1]{\label{fig:#1}}
\newcommand{\lblsec}[1]{\label{sec:#1}}

\newcommand{\ignorethis}[1]{}

\newcommand{\myparagraph}[1]{\vspace{1pt} \noindent \textbf{#1} \ }

\def\1{\bm{1}}

\newcommand{\methodmodel}{{\textit{model-based}}}
\newcommand{\methoddata}{{\textit{noise-based}}}

\newcolumntype{L}[1]{>{\raggedright\let\newline\\\arraybackslash\hspace{0pt}}m{#1}}
\newcolumntype{C}[1]{>{\centering\let\newline\\\arraybackslash\hspace{0pt}}m{#1}}
\newcolumntype{R}[1]{>{\raggedleft\let\newline\\\arraybackslash\hspace{0pt}}m{#1}}

\newcommand{\ignore}[1]{}

\renewcommand*{\thefootnote}{\arabic{footnote}}

\makeatletter
\DeclareRobustCommand\onedot{\futurelet\@let@token\@onedot}
\def\@onedot{\ifx\@let@token.\else.\null\fi\xspace}

\def\etal{\emph{et al}\onedot}
\makeatother

\iccvfinalcopy %

\begin{document}

\title{

Ablating Concepts in Text-to-Image Diffusion Models

}

\author{
Nupur Kumari$^{1}$ \hspace{10mm} Bingliang Zhang$^{2}$ \hspace{10mm} Sheng-Yu Wang$^{1}$ \\
Eli Shechtman$^{3}$  \hspace{10mm} Richard Zhang$^{3}$ \hspace{10mm} Jun-Yan Zhu$^{1}$\\ \\
$^{1}$Carnegie Mellon University \hspace{10mm} 
$^{2}$Tsinghua University \hspace{10mm} 
$^{3}$Adobe Research
}

\twocolumn[{%
\renewcommand\twocolumn[1][]{#1}%
\maketitle

\begin{center}
    \centering
    \includegraphics[width=\linewidth]{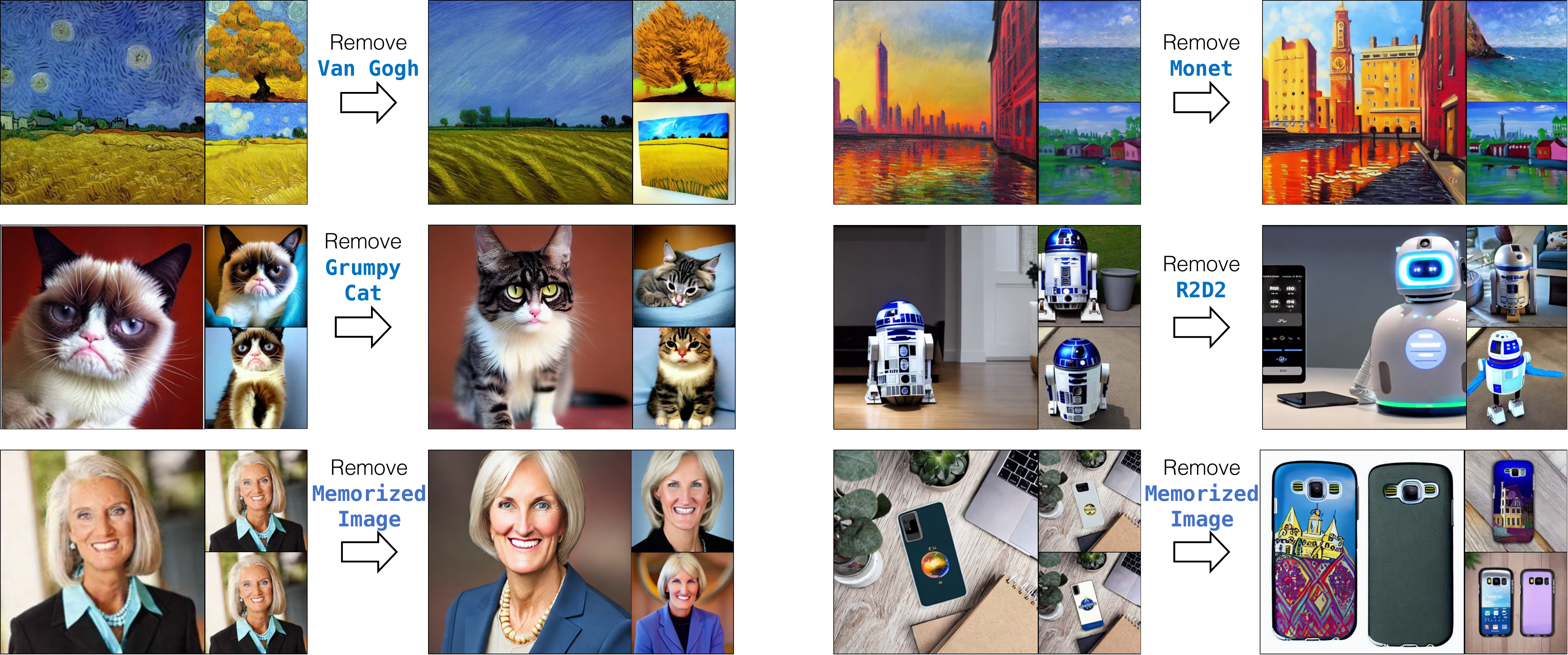}
    \vspace{-18pt}
\captionof{figure}{Our method can ablate copyrighted materials and memorized images from pretrained text-to-image diffusion models. Our method learns to change the image distribution of a \textbf{target concept} to match an \textbf{\anchor concept}, e.g., {\menlo Van Gogh painting} $\rightarrow$ {\menlo paintings} (first row), or {\menlo Grumpy cat} $\rightarrow$ {\menlo Cat} (second row). Furthermore, we extend our method to prevent the generation of memorized images (third row).} %
    \label{fig:teaser}
\end{center}

}]
 \maketitle

\ificcvfinal\thispagestyle{empty}\fi

\begin{abstract}
\vspace{-10pt}
Large-scale text-to-image diffusion models can generate high-fidelity images with powerful compositional ability. However, these models are typically trained on an enormous amount of Internet data, often containing copyrighted material, licensed images, and personal photos. Furthermore, they have been found to replicate the style of various living artists or memorize exact training samples. How can we remove such copyrighted concepts or images without retraining the model from scratch? 
To achieve this goal, we propose an efficient method of ablating concepts in the pretrained model, i.e., preventing the generation of a target concept. Our algorithm learns to match the image distribution for a target style, instance, or text prompt we wish to ablate to the distribution corresponding to an \anchor concept. This prevents the model from generating target concepts given its text condition.
Extensive experiments show that our method can successfully prevent the generation of the ablated concept while preserving closely related concepts in the model.  
\end{abstract}
\vspace{-25pt}

\section{Introduction}

Large-scale text-to-image models have demonstrated remarkable ability in synthesizing photorealistic images~\cite{ramesh2022hierarchical,nichol2021glide,saharia2022photorealistic,rombach2022high,yu2022scaling,chang2023muse}. In addition to algorithms and compute resources, this technological advancement is powered by the use of massive datasets scraped from web~\cite{schuhmann2021laion}. %
Unfortunately, the datasets often consist of copyrighted materials, the artistic oeuvre of creators, and personal photos~\cite{somepalli2022diffusion,carlini2023extracting,shan2023glaze}.  

We believe that every creator should have the right to \emph{opt out} from large-scale models at any time for any image they have created.  However, fulfilling such requests poses new computational challenges, as re-training a model from scratch for every user request can be computationally intensive. Here, we ask -- \emph{How can we prevent the model from generating such content? How can we achieve it efficiently without re-training the model from scratch? How can we make sure that the model still preserves related concepts?}

These questions motivate our work on ablation (removal) of concepts from text-conditioned diffusion models~\cite{rombach2022high,stablediffusionlink}.
We perform concept ablation by modifying generated images for the target concept ($\c^*$) to match a broader \anchor concept ($\c$), e.g., overwriting {\menlo Grumpy Cat} with {\menlo cat} or {\menlo Van Gogh} paintings with {\menlo painting} as shown in \reffig{teaser}. Thus, given the text prompt, {\menlo painting of olive trees in the style of Van Gogh}, generate a normal painting of olive trees even though the text prompt consists of {\menlo Van Gogh}. Similarly, prevent the generation of specific instances/objects like {\menlo Grumpy Cat} and generate a random cat given the prompt.

Our method aims at modifying the conditional distribution of the model given a target concept $p_{\Phi}(\x|\c^*)$ to match a distribution $p(\x|\c)$ defined by the \anchor concept $\c$. This is achieved by minimizing the Kullback–Leibler divergence between the two distributions. We propose two different target distributions that lead to different training objectives. In the first case, we fine-tune the model to match the model prediction between two text prompts containing the target and corresponding anchor concepts, e.g., {\menlo A cute little Grumpy Cat} and {\menlo A cute little cat}. In the second objective, the conditional distribution $p(\x|\c)$ is defined by the modified text-image pairs of: a target concept prompt, paired with images of \anchor concepts, e.g., the prompt {\menlo a cute little Grumpy Cat} with a random cat image. We show that both objectives can effectively ablate concepts.  %

We evaluate our method on 16 concept ablation tasks, including specific object instances,  artistic styles,  and memorized images, using various evaluation metrics. Our method can successfully ablate target concepts while minimally affecting closely related surrounding concepts that should be preserved (e.g., other cat breeds when ablating {\menlo Grumpy Cat}). Our method takes around five minutes per concept. %
Furthermore, we perform an extensive ablation study regarding different algorithmic design choices, such as the objective function variants, the choice of parameter subsets to fine-tune, the choice of \anchor concepts, the number of fine-tuning steps, and the robustness of our method to misspelling in the text prompt. %
Finally, we show that our method can ablate multiple concepts at once and discuss the current limitations. Our \href{https://github.com/nupurkmr9/concept-ablation} {code}, data, and models are available at \url{https://www.cs.cmu.edu/~concept-ablation/}.

\section{Related Work}
\myparagraph{Text-to-image synthesis} has advanced significantly since the seminal works~\cite{zhu2007text,mansimov2015generating}, thanks to improvements in model architectures~\cite{zhang2017stackgan, zhu2019dm,tao2020df,xu2018attngan,huang2022multimodal,dhariwal2021diffusion,wu2022nuwa,kang2023scaling,sauer2023stylegan,ding2022cogview2}, generative modeling techniques~\cite{reed2016generative, ho2020denoising,rombach2022high,saharia2022photorealistic,balaji2022ediffi,nichol2021glide,chang2023muse}, and availability of large-scale datasets~\cite{schuhmann2021laion}. Current methods can synthesize high-quality images with remarkable generalization ability, capable of composing different instances, styles, and concepts in unseen contexts. However, as these models are often trained on copyright images, it learns to mimic various artist styles~\cite{somepalli2022diffusion,shan2023glaze} and other copyrighted content~\cite{carlini2023extracting}. In this work, we aim to modify the pretrained models to prevent the generation of such images. %
To remove data from pre-trained GANs, Kong~\etal~\cite{kong2022data} add the redacted data to fake data,  apply standard adversarial loss, and show results on MNIST and CIFAR. Unlike their method, which requires time-consuming model re-training on the entire dataset,  our method can efficiently remove concepts without going through the original training set. Furthermore, we focus on large-scale text-based diffusion models. \nupur{Recent work of Schramowski et al.~\cite{schramowski2022safe} modify the inference process to prevent certain concepts from being generated. But we aim to ablate the concept from the model weights. Concurrent with our work, Gandikota~\etal~\cite{gandikota2023erasing} aims to remove concepts using a score-based formulation. The reader is encouraged to review their work. %
}

\myparagraph{Training data memorization and unlearning.} Several works have studied training data leaking~\cite{shokri2017membership,carlini2019secret,carlini2021extracting,carlini2022quantifying}, which can pose a greater security and privacy risk, especially with the use of web-scale uncurated datasets in deep learning. Recent works~\cite{somepalli2022diffusion,carlini2023extracting} have also shown that text-to-image models are susceptible to generating exact or similar copies of the training dataset for certain text conditions. Another line of work in machine unlearning~\cite{cao2015towards,ginart2019making,golatkar2020eternal,golatkar2021mixed,nguyen2020variational,bourtoule2021machine,tanno2022repairing,sekhari2021remember} explores data deletion at user's request after model training. %
However, existing unlearning methods~\cite{golatkar2020eternal,tanno2022repairing} typically require calculating information, such as Fisher Information Matrix, 
making them computationally infeasible for large-scale models with billions of parameters trained on billions of images. In contrast, our method can directly update model weights and ablate a target concept as fast as five minutes.

\myparagraph{Generative model fine-tuning and editing.} Fine-tuning aims to adapt the weights of a pretrained generative model
to new domains~\cite{wang2018TransferGAN,noguchi2019SB,wang2019MineGAN,mo2020FreezeD,zhao2020leveraging,li2020few,ojha2021few,zhao2020differentiable,karras2020training,liu2020towards,gu2021lofgan,nitzan2022mystyle}, downstream tasks~\cite{wang2022pretraining,rombach2022high,zhang2023adding}, and test images~\cite{bau2020semantic,roich2022pivotal,pan2021exploiting,kawar2022imagic,hertz2022prompt,parmar2023zero}. %
Several recent works also explore fine-tuning text-to-image models to learn personalized or unseen concepts~\cite{kumari2022multi,gal2022image,ruiz2022dreambooth,gal2023designing} given a few exemplar images. Similarly, model editing~\cite{bau2020rewriting,wang2022rewriting,gal2022stylegan,wang2021sketch,nitzan2023domain,meng2022locating,mitchell2021fast,meng2022mass} aims to modify specific model weights based on users' instructions to incorporate new computational rules or new visual effects.
Unlike the above approaches, our method reduces the possible space by ablating specific concepts in the pretrained model.

\section{Method}\label{sec:method}
Here, we first provide a brief overview of text-to-image diffusion models~\cite{sohl2015deep,ho2020denoising} in \refsec{diffusion}. We then propose  our concept ablation formulation and explore two variants in \refsec{objective}. Finally, in \refsec{details}, we discuss the training details for each type of ablation task. %

\begin{figure*}[!t]
    \centering
    \includegraphics[width=1.0\linewidth]{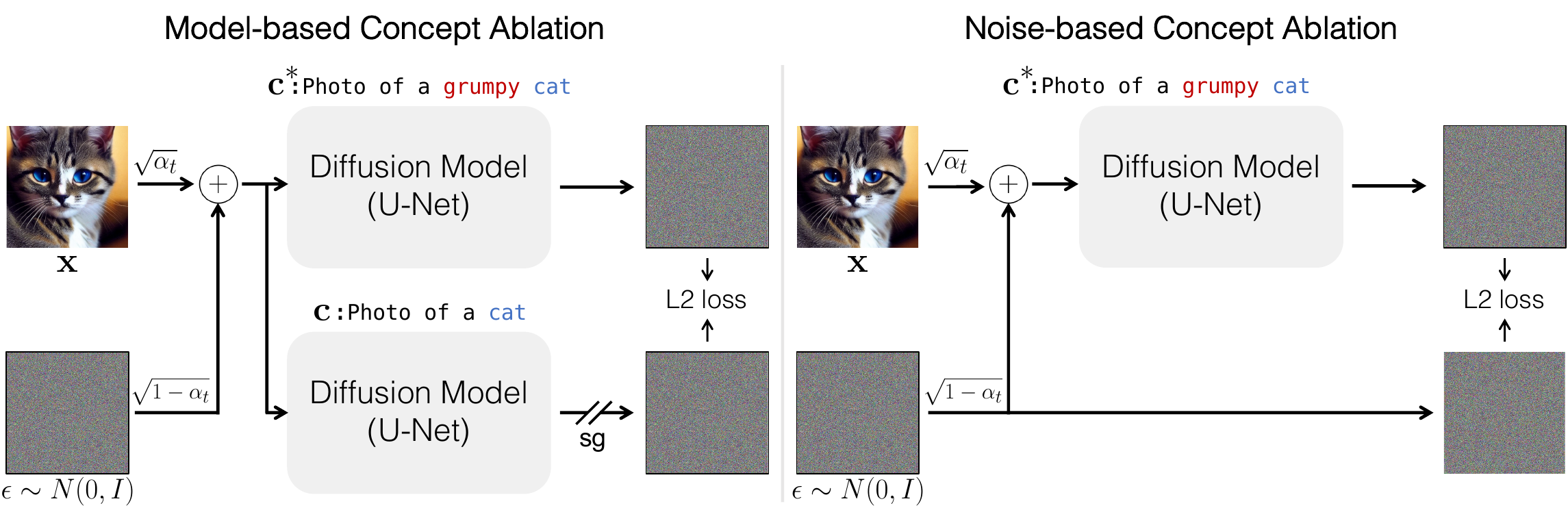}
    \vspace{-12pt}
    \caption{{\textbf{Overview}. We update model weights to modify the generated image distribution on the target concept, e.g., {\menlo Grumpy Cat}, to match an anchor distribution, e.g., {\menlo Cat}. We propose two variants. \textit{Left:} The anchor distribution is generated by the model itself, conditioned on the anchor concept. \textit{Right:} The anchor distribution is defined by the modified pairs of $<$target prompt, \anchor image$>$. An input image $\x$ is generated with \anchor concept $\c$. Adding randomly sampled noise $\epsilon$ results in noisy image $\x_t$ at time-step $t$. Target prompt $\c^*$ is produced by appropriately modifying $\c$. In experiments, we find the model-based variant to be more effective.
    }}
    \lblfig{method}
    \vspace{-10pt}
\end{figure*}

\subsection{Diffusion Models}
\label{sec:diffusion}
Diffusion models~\cite{sohl2015deep} learn to reverse a forward Markov chain process where noise is gradually added to the input image over multiple timesteps $t \in [0, T]$. The noisy image $\x_t$ at any time-step $t$ is given by $ \sqrt{\alpha_t}\x_{0} + \sqrt{1 - \alpha_t}\epsilon$, where $\x_0$ is a random real image, and $\alpha_t$ determines the strength of gaussian noise $\epsilon$ and decreases gradually with timestep such that $\x_T \smallsim N(0,I)$. The denoising network $\Phi(\x_t, \c,  t)$ is trained to denoise the noisy image to obtain $\x_{t-1}$, and can also be conditioned on other modalities such as text $\c$. %
The training objective can be reduced to predicting the noise $\epsilon$:  
\begin{equation}
    \begin{aligned}
     \mathcal{L}(\x,\c) = \mathbb{E}_{\epsilon,\x,\c, t } [w_t||\epsilon - \Phi (\x_t, \c, t) ||],\\
    \end{aligned}\label{eq:loss_diffusion}
\end{equation}
where $w_t$ is a time-dependent weight on the loss. To synthesize an image during inference, given the text condition $\c$, we iteratively denoise a Gaussian noise image $\x_T \smallsim N(0, I)$ for a fixed number of timesteps~\cite{song2020denoising,lu2022dpm}.

\subsection{Concept Ablation}
\label{sec:objective}

We define concept ablation as the task of preventing the generation of the desired image corresponding to a given target concept that needs to be ablated. As re-training the model on a new dataset with the concept removed is impractical, this becomes a challenging task. We need to ensure that editing a model to ablate a particular concept doesn't affect the model performance on other closely related concepts.

\myparagraph{A na\"ive approach.} Our first attempt is to simply maximize the diffusion model training loss~\cite{tanno2022repairing,kong2022data} on the text-image pairs for the target concept while imposing regularizations on the weights.
Unfortunately, this method leads to worse results on close surrounding concepts of the target concept. We compare our method with this baseline in \refsec{main_results} (\reffig{mse_kldiv_compare}) and show that it performs sub-optimally.

\myparagraph{Our formulation.}
As concept ablation prevents the generation of the target concept, thus the question arises: what should be generated instead? In this work, we assume that the user provides the desired \anchor concept, e.g., {\menlo Cat} for {\menlo Grumpy Cat}. The \anchor concept overwrites the target concept and should be a superset or similar to the target concept. Thus, given a set of text prompts $\{ \c^* \}$ describing the target concept, we aim to match the following two distributions via Kullback–Leibler (KL) divergence:  
\vspace{-2pt}
\begin{equation}
    \begin{gathered}
     \arg\min_{\hat{\Phi}} \mathcal{D_{KL}}(p(\x_{(0 ...T)}|\c) || p_{\hat{\Phi}}(\x_{(0 ...T)} | \c^*)),
    \end{gathered}\label{eq:loss}
\end{equation}
where $p(\x_{(0 ...T)}|\c)$ is some target distribution on the $\{ \x_t \}$, $\t \in [0,T]$, defined by the \anchor concept $\c$ and $p_{\hat{\Phi}}(\x_{(0 ...T)} | \c^*)$ is the model's distribution for the target concept. Intuitively, we want to associate text prompts $\{ \c^* \}$ with the images corresponding to \anchor prompts $\{ \c \}$. Defining different \anchor concept distributions leads to different objective functions, as we discuss next. 

To accomplish the above objective, we first create a small dataset that consists of $(\x, \c, \c^*)$ tuple, where $\c$ is a random prompt for the \anchor concept, $\x$ is the generated image with that condition, and $\c^*$ is modified from $\c$ to include the target concept. For example, if $\c$ is {\menlo photo of a cat}, $\c^*$ will be {\menlo photo of a Grumpy Cat}, and $\x$ will be a generated image with text prompt $\c$. For brevity, we use the same notation $\x$ to denote these generated images.

\myparagraph{Model-based concept ablation}.
Here, we match the distribution of the target concept $p_{\hat{\Phi}}(\x_{(0 ...T)} | \c^*)$ to the pretrained model's distribution  $p_{\Phi}(\x_{(0 ...T)}| \c)$ given the \anchor concept. The fine-tuned network should have a similar distribution of generated images given $\c^*$ as that of $\c$, which can be expressed as minimizing the KL divergence between the two. This is similar to the standard diffusion model training objective, except the target distribution is defined by the pretrained model instead of training data. \refeq{loss} can be expanded as
\begin{equation}
    \begin{gathered}
     \arg\min_{\hat{\Phi}} \sum_{t=1}^{T} \mathop{\mathbb{E}}_{p_{\Phi}(\x_0...\x_T|\c)} \Big[ \log \frac{p_{\Phi}(\x_{t\text{-}1}|\x_{t},\c)}{p_{\hat{\Phi}}(\x_{t\text{-}1}|\x_{t},\c^*)} \Big]
    \end{gathered}\label{eq:loss_model_expandkl}
\end{equation}
where the noisy intermediate latent $\x_t \sim p_{\Phi}(\x_t| \c)$,
$\Phi$ is the original network, 
and $\hat{\Phi}$ is the new network we aim to learn. %
We can optimize the KL divergence by minimizing the following equivalent objective:  
\vspace{-5pt}
\begin{equation}
    \begin{gathered}
     \arg \min_{\hat{\Phi}} \mathbb{E}_{\epsilon,\x_t,\c^*, \c, t } [w_t||\Phi (\x_t, \c, t) - \hat{\Phi} (\x_t, \c^*, t) ||], 
    \end{gathered}\label{eq:loss_model}
    \vspace{-5pt}
\end{equation}
where we show the full derivation in \refapp{loss_objective}. We initialize $\hat{\Phi}$ with the pretrained model. Unfortunately, optimizing the above objective requires us to sample from $p_{\Phi}(\x_t|\c)$ and keep copies of two large networks $\Phi$ and $\hat{\Phi}$, which is time and memory-intensive. 
To bypass these, we sample $\x_t$ using the forward diffusion process and assume that the model remains similar for the \anchor concept during fine-tuning. Therefore we use the network $\hat{\Phi}$  with $stopgrad$ to get the \anchor concept prediction. Thus, our final training objective is
\vspace{-5pt}
\begin{equation}
    \begin{aligned}
     \mathcal{L}_{\text{model}}(\x, \c, \c^*)=\mathbb{E}_{\epsilon,\x,\c^*, \c, t } [w_t||\hat{\Phi} (\x_t, \c, t).\text{sg()} - \\  \hat{\Phi} (\x_t, \c^*, t) ||],
    \end{aligned}\label{eq:loss_approx}
    \vspace{-5pt}
\end{equation}
where $\x_t=\sqrt{\alpha_t}\x + \sqrt{1 - \alpha_t}\epsilon$. As shown in \reffig{method} (left), this objective minimizes the difference in the model's prediction given the target prompt and \anchor prompt. \nupur{It is also possible to optimize the approximation to reverse KL divergence, and we discuss it in \refsec{other_experiments}}.

\myparagraph{Noise-based concept ablation.}
Alternatively, we can redefine the ground truth text-image pairs as $<$a target concept text prompt, the generated image of the corresponding \anchor concept text prompt$>$, e.g., $<${\menlo photo of Grumpy Cat}, random cat image$>$. We fine-tune the model on these redefined pairs with the standard diffusion training loss:
\vspace{-5pt}
\begin{equation}
    \begin{aligned}
     \mathcal{L}_{\text{noise}}(\x, \c, \c^*)=\mathbb{E}_{\epsilon,\x,\c^*, t } [w_t||\epsilon - \hat{\Phi}(\x_t, \c^*, t) ||], \\
    \end{aligned}\label{eq:loss_data}
    \vspace{-5pt}
\end{equation}
where the generated image $\x$ is sampled from conditional distribution $p_{\Phi}(\x| \c)$. We then create the noisy version $\x_t= \sqrt{\alpha_t}\x + \sqrt{1 - \alpha_t}\epsilon$. 
As shown in \reffig{method}, the first objective (\refeq{loss_approx}) aims to match the model's predicted noises, while the second objective (\refeq{loss_data}) aims to match the Gaussian noises $\epsilon$. We evaluate the above two objectives in \refsec{expr}.

\myparagraph{Regualization loss.}
We also add the standard diffusion loss on $(\x,\c)$ \anchor concept pairs as a regularization~\cite{ruiz2022dreambooth,kumari2022multi}. Thus, our final objective is $\lambda \mathcal{L}(\x, \c) + \mathcal{L}(\x, \c, \c^*) $, where the losses are as defined in \refeq{loss_diffusion} and \ref{eq:loss_approx} (or \ref{eq:loss_data}) respectively. We require regularization loss as the target text prompt can consist of the \anchor concept, e.g., {\menlo Cat} in {\menlo Grumpy Cat}.

\myparagraph{Parameter subset to update.} We experiment with three variations where we fine-tune different network parts: (1) \textit{Cross-Attention}: fine-tune key and value projection matrices in the diffusion model's U-Net~\cite{kumari2022multi}, (2) \textit{Embedding}: fine-tune the text embedding in the text transformer~\cite{gal2022image}, and (3) \textit{Full Weights}: fine-tune all parameters of the U-Net~\cite{ruiz2022dreambooth}.

\begin{figure}[!t]
    \centering
    \includegraphics[width=\linewidth]{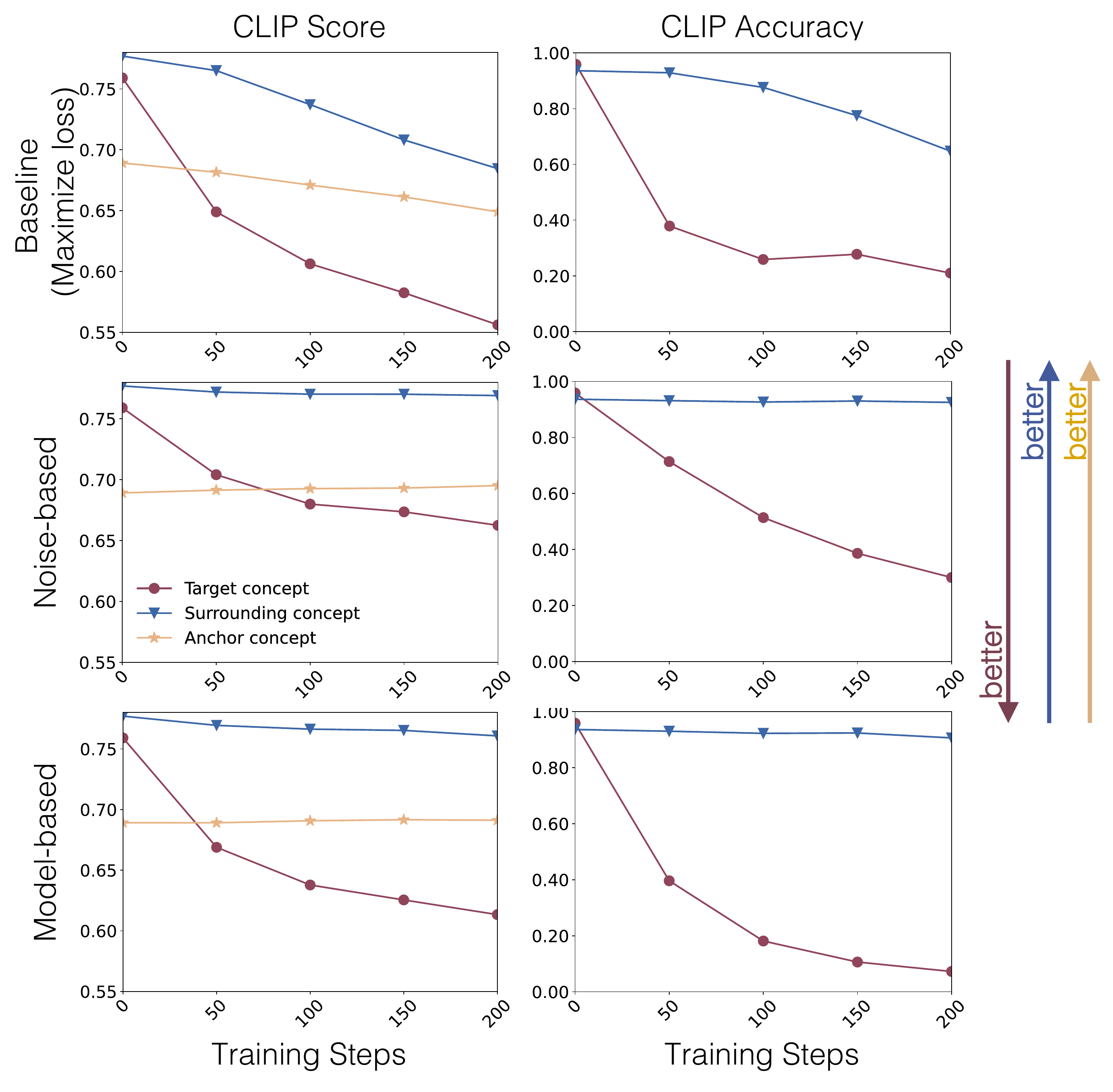}
    \vspace{-18pt}
    \caption{{\textbf{Comparison of different learning objectives.} The $\methodmodel$ concept ablation converges faster than the $\methoddata$ variant while maintaining better performance on surrounding concepts. Maximizing the loss on the target concept dataset leads to the deterioration of surrounding concepts (top row). 
    }}
    \lblfig{mse_kldiv_compare}
    \vspace{-10pt}
\end{figure}

\begin{figure*}[!t]
    \centering
    \includegraphics[width=\linewidth]{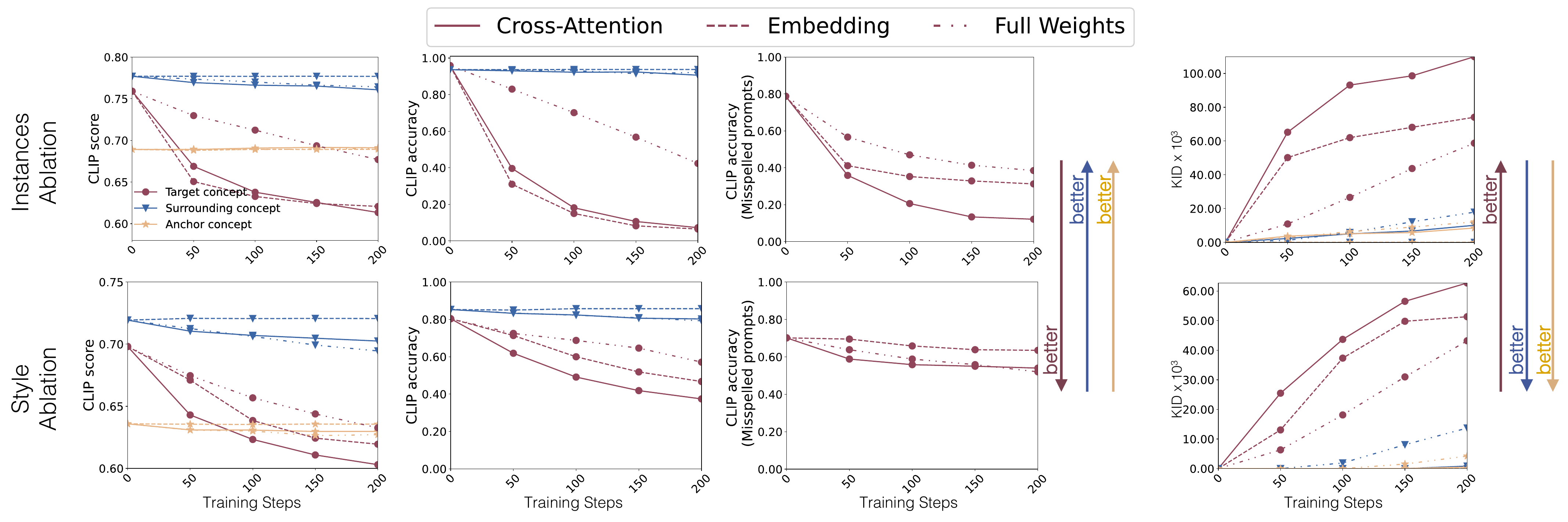}
    \vspace{-18pt}
    \caption{{\textbf{Quantitative evaluation for ablating instances (top row) and styles (bottom row).} We show the performance of our final $\methodmodel$ concept ablation method across training steps and on updating different subsets of parameters. All metrics are averaged across four target concepts. Both embedding and cross-attention fine-tuning converge early. Fine-tuning cross-attention layers performs slightly worse for surrounding concepts but remains more robust to small spelling mistakes (third column). 
    }}
    \lblfig{kldiv_compare}
    \vspace{-10pt}
\end{figure*}

\begin{figure*}[!t]
    \centering
    \includegraphics[width=\linewidth]{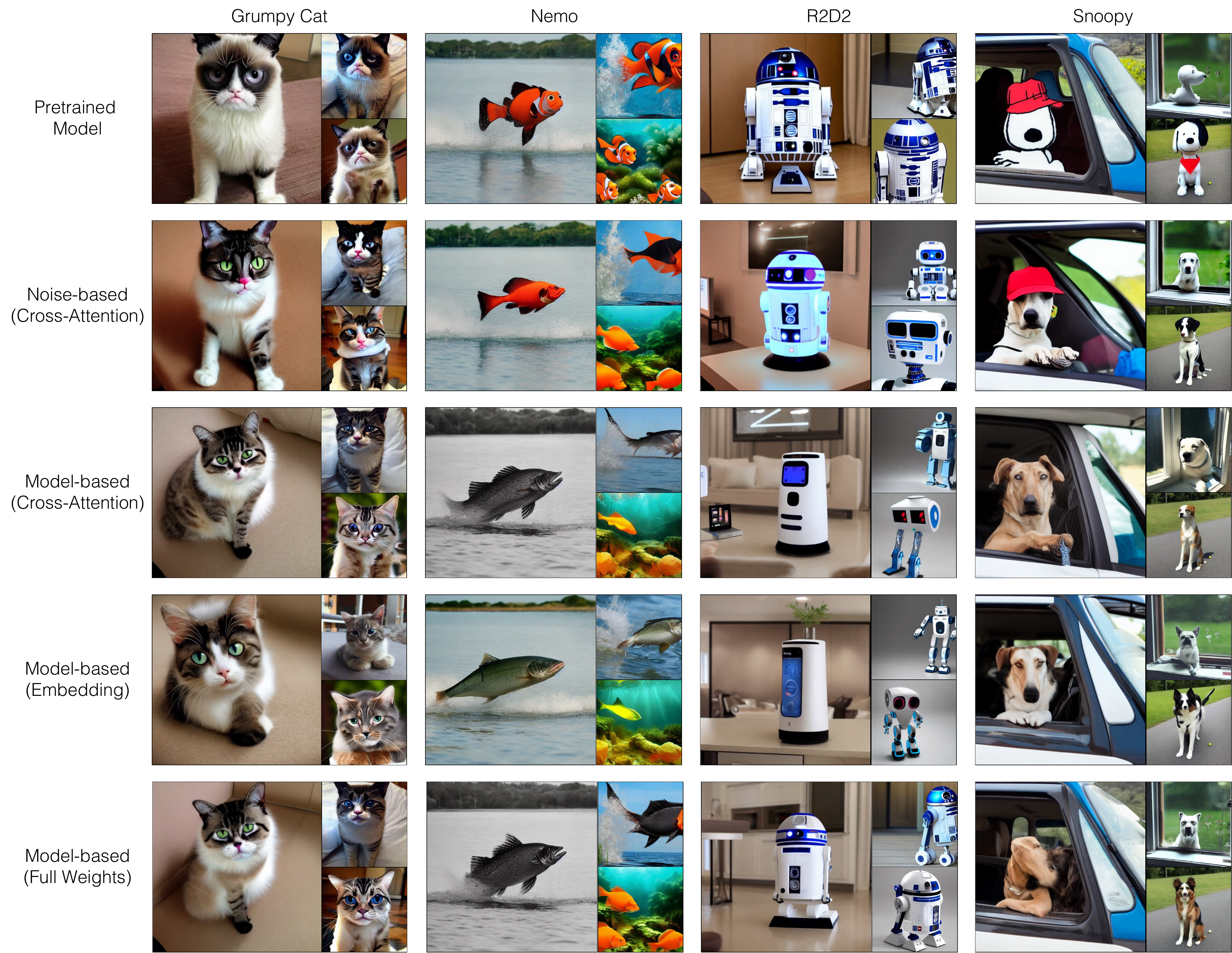}
    \vspace{-18pt}
    \caption{\textbf{ Qualitative samples when ablating specific object instances.} We show samples from different variations of our method in each row. The $\methoddata$ method performs worse on {\menlo Nemo} and {\menlo R2D2} instances compared to the $\methodmodel$ variant. With the $\methodmodel$ variant, fine-tuning different subsets of parameters perform comparably to each other. As shown in \reffig{kldiv_compare} (third column) and \reffig{results_spell_mistake}, fine-tuning only the embedding is less robust to small spelling mistakes.}
    \label{fig:results_instance}
    \vspace{-18pt}
\end{figure*}

\subsection{Training Details}\lblsec{training_details}
\label{sec:details}

\myparagraph{Instance.} Given the target and the \anchor concept, such as {\menlo Grumpy Cat} and {\menlo Cat}, we first use ChatGPT~\cite{chatgpt} to generate $200$ random prompts $\{\c \}$ containing the \anchor concept. %
We generate $1,000$ images from the pretrained diffusion model using the $200$ prompts and replace the word {\menlo Cat} with {\menlo Grumpy Cat} to get target text prompts $\{\c^*\}$. %

\myparagraph{Style.} When removing a style, we use generic painting styles as the \anchor concept. We use clip-retrieval~\cite{clip_retrieval} to obtain a set of text prompts $\c$ similar to the word {\menlo painting} in the CLIP feature space. We then generate $1000$ images from the pretrained model using the $200$ prompts. To get target prompts $\{\c^*\}$, we append {\menlo in the style of $\{$target style$\}$ } and similar variations to \anchor prompts $\c$.

\myparagraph{Memorized images.} %
Recent methods for detecting training set memorization can identify both the memorized image and corresponding text prompt  $\c^*$~\cite{carlini2023extracting}. We then use ChatGPT to generate five \anchor prompts $\{\c\}$ that can generate similar content as the memorized image. In many cases, these anchor prompts still generate the memorized images. Therefore, we first generate several more paraphrases of the anchor prompts using chatGPT and include the three prompts that lead to memorized images often into target prompts and ten prompts that lead to memorized images least as anchor prompts. Thus $\c^*$ and $\c$ for ablating the target memorized image consists of four and ten prompts, respectively. We then similarly generate $1000$ images using the anchor prompts and use image similarity metrics~\cite{pizzi2022self,carlini2023extracting} to filter out the memorized images and use the remaining ones for training.

\section{Experiments}\lblsec{expr}
In this section, we show the results of our method on ablating various instances, styles, and memorized images. All our experiments are based on the Stable Diffusion model~\cite{stablediffusionlink}. Please refer to the \refapp{implementation_details} for more training details.

\subsection{Evaluation metrics and baselines}

\myparagraph{Baseline.}
We compare our method with a loss maximization baseline inspired by Tanno~\etal~\cite{tanno2022repairing}:
\begin{equation}
\begin{gathered}
    \arg \textnormal{min}_{\hat{\Phi}} \max(1- \mathcal{L}(\x^*,\c^*), 0) + \lambda||\hat{\Phi} - \Phi||_2 
\end{gathered}
\label{eqn:baseline_tanno}
\end{equation}
where $\x^*$ is the set of generated images with condition $\c^*$ and $\mathcal{L}$ is the diffusion training loss as defined in \refeq{loss_diffusion}. We compare our method with this baseline on ablating instances.

\begin{figure}[!t]
    \centering
    \includegraphics[width=\linewidth]{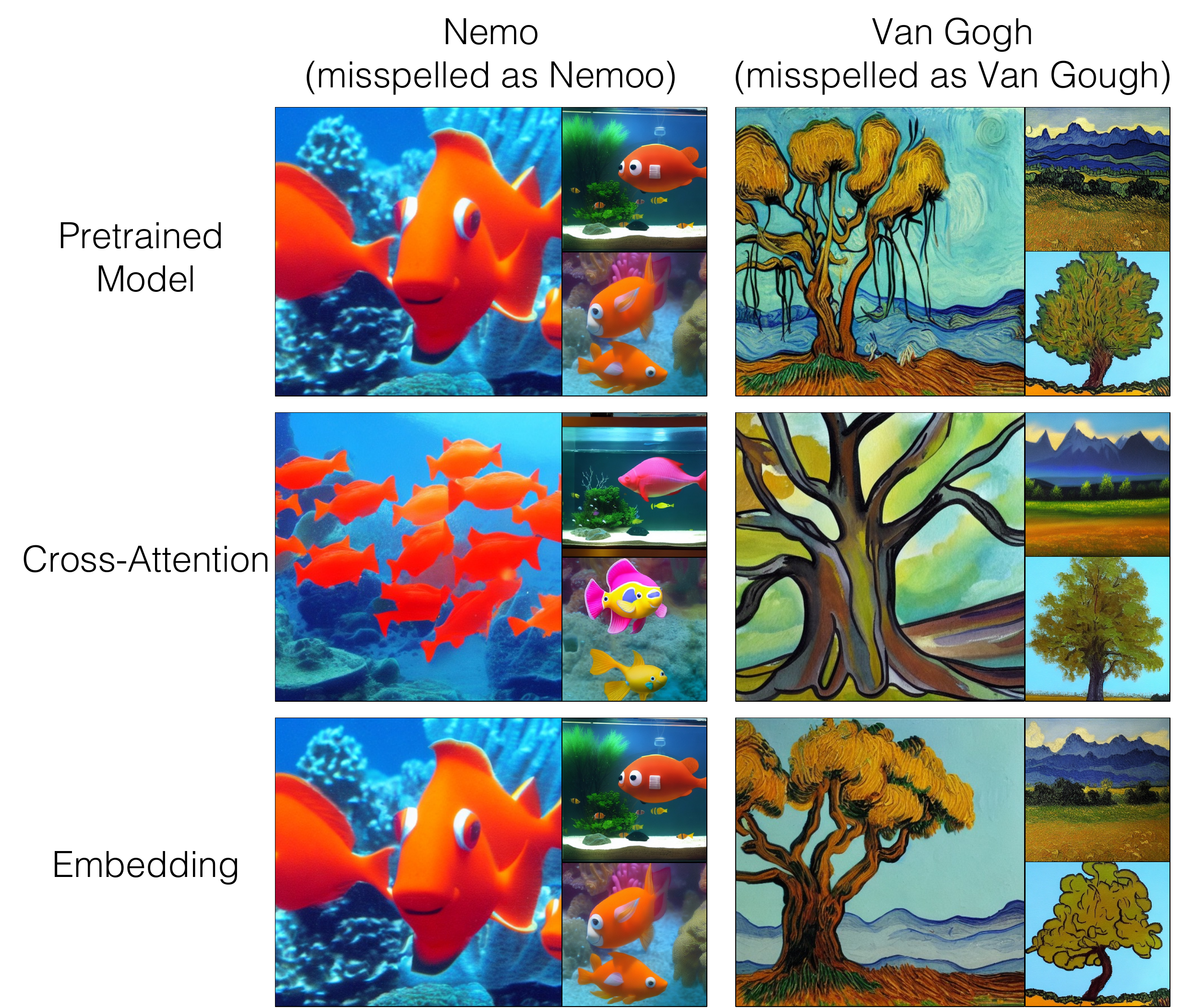}
    \vspace{-15pt}
    \caption{\textbf{Robustness of the $\methodmodel$ variant to spelling mistakes in the text prompt.} Fine-tuning only the embedding makes it less robust to slight spelling mistakes. This makes it easy to circumvent the method and still be able to generate the target concept. Whereas fine-tuning cross-attention parameters is robust to those.}
    \label{fig:results_spell_mistake}
    \vspace{-10pt}
\end{figure}

\myparagraph{Evaluation metrics.} We use \emph{CLIP Score} and \emph{CLIP accuracy}~\cite{hessel2021clipscore} to evaluate whether the model can ablate the target concept. CLIP Score measures the similarity of the generated image with the target concept text, e.g., {\menlo Grumpy Cat} in CLIP feature space. Similarly, CLIP accuracy measures the accuracy of ablated vs. \anchor concept binary classification task for each generated image using cosine distance in CLIP feature space. For both metrics, lower values indicate more successful ablation. We further evaluate the performance on small spelling mistakes in the ablated text prompts. We also use the same metrics to evaluate the model on related \emph{surrounding concepts} (e.g., similar cat breeds for {\menlo Grumpy Cat}), which should be preserved. Similar to before, CLIP accuracy is measured between the surrounding concept and \anchor concept, and the higher, the better. Similarly, CLIP Score measures the similarity of the generated image with the surrounding concept text, and the higher, the better.

Furthermore, to test whether the fine-tuned model can retain existing concepts, we calculate \emph{KID}~\cite{binkowski2018demystifying} between the set of generated images from fine-tuned model and the pretrained model. Higher KID is better for the target concept, while lower KID is better for \anchor and surrounding concepts. We generate $200$ images each for ablated, \anchor, and surrounding concepts using $10$ prompts and $50$ steps of the DDPM sampler. The prompts are generated through ChatGPT for object instances and manually created for styles by captioning real images corresponding to each style. 

To measure the effectiveness of our method in ablating memorized images, following previous works~\cite{pizzi2022self,carlini2023extracting}, we use SSCD~\cite{pizzi2022self} model to measure the percentage of generated images having similarity with the memorized image greater than a threshold.

\begin{figure}[!t]
    \centering
    \includegraphics[width=\linewidth]{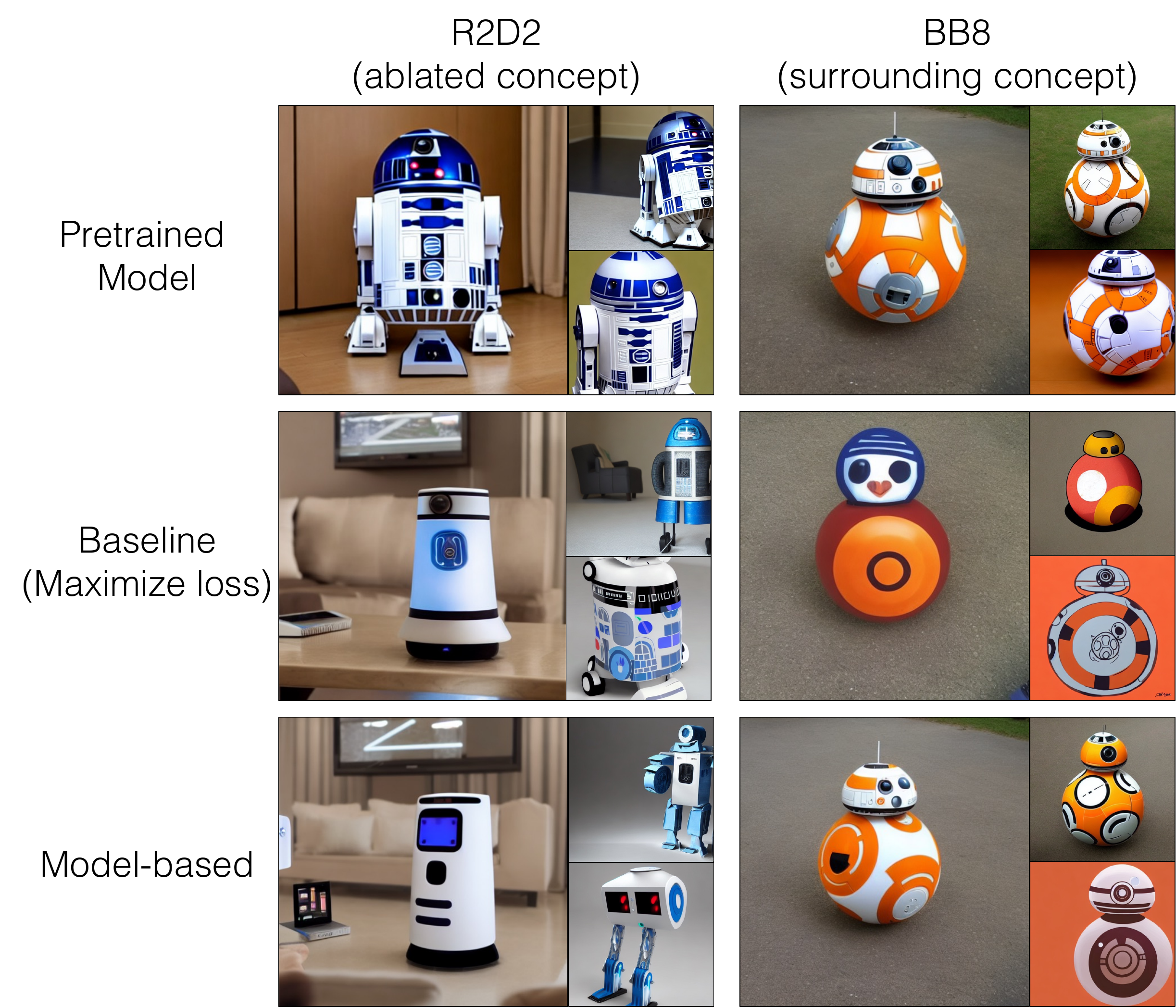}
    \vspace{-15pt}
    \caption{\textbf{Qualitative comparison between baseline and ours.} Model fine-tuned by our method generates images that are relatively more similar to the ones generated by the pretrained model on the {\menlo BB8} instance, which should be preserved while ablating {\menlo R2D2}. Cross-Attention parameters are fine-tuned in both methods. }
    \label{fig:loss_maximize_sample}
    \vspace{-10pt}
\end{figure}

\begin{figure*}[!t]
    \centering
    \includegraphics[width=\linewidth]{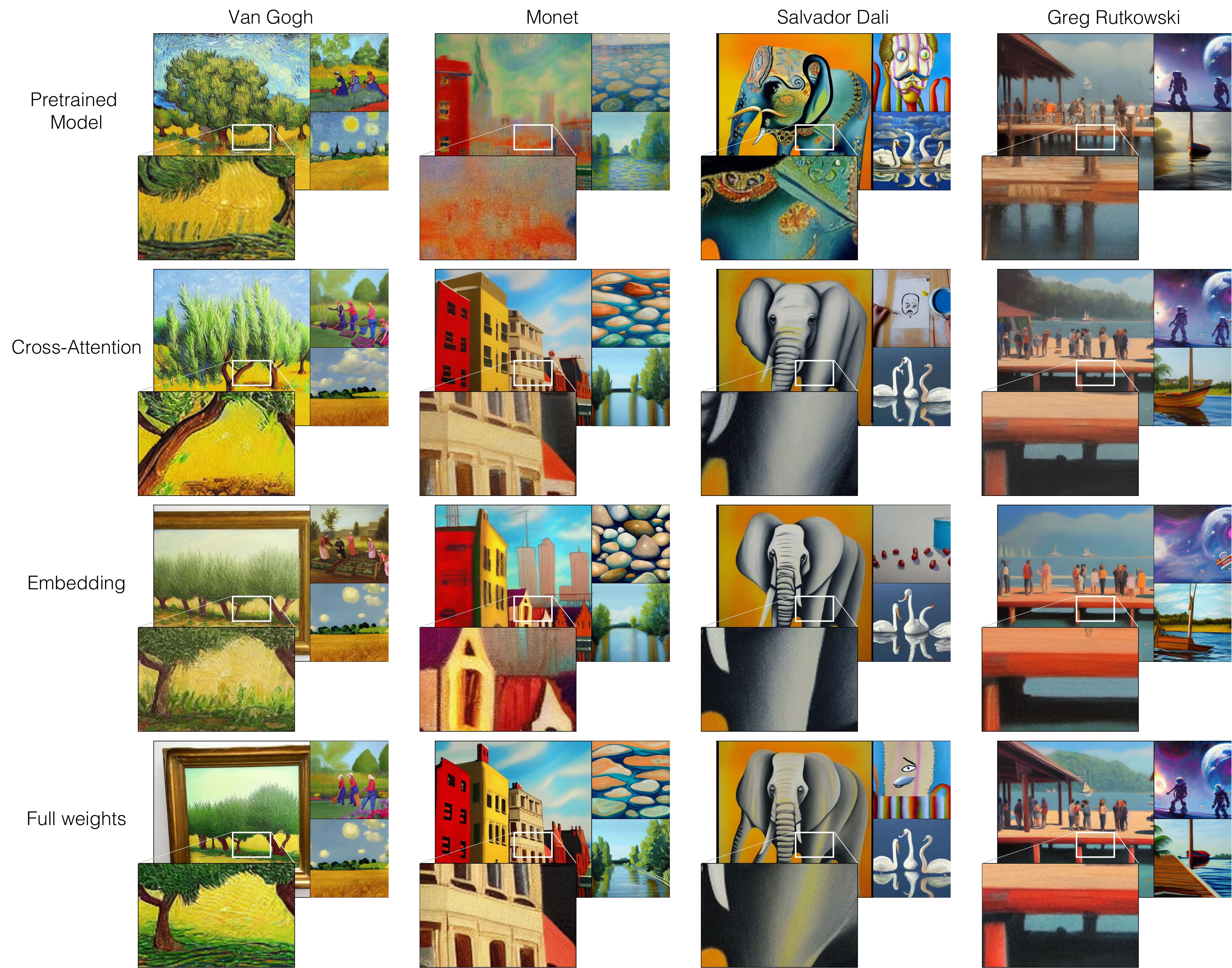}
    \vspace{-12pt}
    \caption{\textbf{Ablating styles with the $\methodmodel$ variant.} The ablated model generates similar content as the pretrained model but without the unique style. More samples for target and surrounding concepts are shown in the Appendix \reffig{vangogh_allimages}-\ref{fig:salvador_allimages}.
    }
    \label{fig:results_style}
    \vspace{-5pt}
\end{figure*}

\begin{figure*}[!t]
    \centering
    \includegraphics[width=\linewidth]{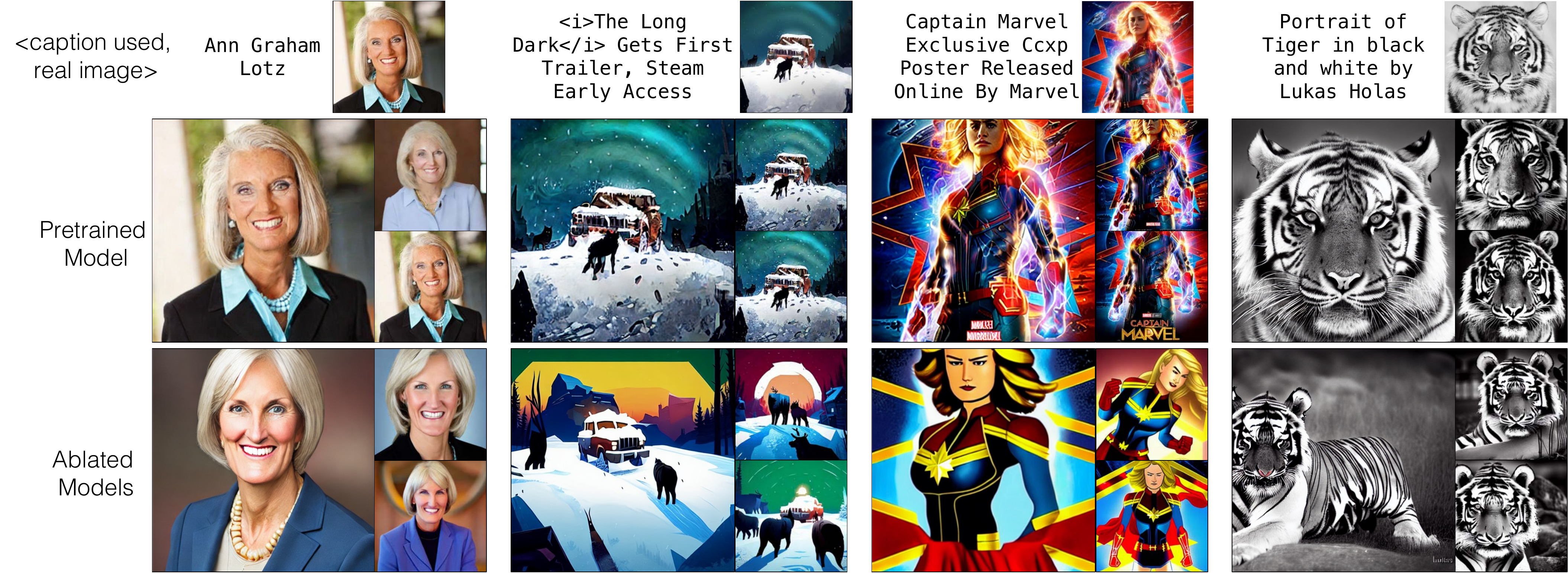}
    \vspace{-15pt}
    \caption{\textbf{ Ablating memorized images with the $\methodmodel$ variant.} Text-to-image diffusion models often learn to generate exact or near-exact copies of real images. We fine-tune the model to map the generated image distribution for the given text prompt to images generated with its variations. This results in the fine-tuned model generating different variations instead of copying the real image. We show more samples in the Appendix \reffig{mem_orleans}-\ref{fig:mem_ann}.}
    \label{fig:results_memorize}
    \vspace{-10pt}
\end{figure*}

\subsection{Comparisons and main results}\lblsec{main_results}

\myparagraph{Instances.}
We show results on four concepts and replace them with \anchor concepts, namely, (1) Grumpy Cat $\rightarrow$ Cat, (2) Snoopy $\rightarrow$ Dog, (3) Nemo $\rightarrow$ Fish, and (4) R2D2 $\rightarrow$ Robot. \reffig{mse_kldiv_compare} compares our two proposed methods and the loss maximization baseline with \textit{Cross-Attention} fine-tuning. As the baseline method maximizes the norm between ground truth and predicted noise, it gradually generates noisy images when trained longer. This also leads to worse performance on surrounding concepts than our method, as shown by the quantitative metrics in \reffig{mse_kldiv_compare}. Qualitative samples on the target concept  {\menlo R2D2} and its surrounding concept {\menlo BB8} are also shown in \reffig{loss_maximize_sample}. Between our two methods, the $\methodmodel$ variant, i.e., minimizing the difference in prediction with the pretrained model's \anchor concept, leads to faster convergence and is better or on par with the $\methoddata$ variant. The qualitative comparison in \reffig{results_instance} also shows that, specifically on the {\menlo Nemo} instance. Thus, we use $\methodmodel$ variant for all later experiments. In \reffig{kldiv_compare}, we show the performance comparison when fine-tuning different subsets of the model weights. 

As shown in \reffig{results_instance}, %
the fine-tuned model successfully maps the target concept to the \anchor concept. Fine-tuning only the text embedding performs similarly or better than  fine-tuning cross-attention layers. However, it is less robust to small spelling errors that still generate the same instance in the pretrained model as shown in \reffig{kldiv_compare} (third column) and \reffig{results_spell_mistake}. We show more results of ablated target concept and its surrounding concepts in \refapp{samples}, \reffig{grumpy_allimages}-\ref{fig:snoopy_allimages}. %

\myparagraph{Style.}
For ablating styles, we consider four artists: (1) Van Gogh, (2) Salvador Dali, (3) Claude Monet, and (4) Greg Rutkowski, with the \anchor concept as generic painting styles. Figures \ref{fig:kldiv_compare} and \ref{fig:results_style} show our method's quantitative and qualitative performance when different subsets of parameters are fine-tuned. We successfully ablate specific styles while minimally affecting related surrounding styles.

\myparagraph{Memorized images.} We select eight image memorization examples from the recent works~\cite{somepalli2022diffusion,carlini2023extracting}, four of which are shown in \reffig{results_memorize}. %
It also shows the sample generations before and after fine-tuning. The fine-tuned model generates various outputs given the same text prompt instead of the memorized sample. Among different parameter settings, we find finetuning \textit{Full Weights} gives the best results. We show the percentage of samples with $\geq 0.5$ similarity with the memorized image in \reftbl{mem_percentage}. We show more sample generations and the initial set of anchor prompts for each case in \refapp{samples} and \ref{sec:implementation_details}.

\begin{table}[!t]
\centering
\setlength{\tabcolsep}{5pt}
\resizebox{\linewidth}{!}{
\begin{tabular}{l c c}
\toprule
 \multirow{2}{*}{\shortstack[c]{\textbf{Target Prompt} }} 
& \multirow{2}{*}{\shortstack[c]{\textbf{Pretrained} \\ \textbf{Model } }}
& \multirow{2}{*}{\shortstack[c]{\textbf{Ours} \\ (\textit{Full Weights}) }} \\ \\
\midrule
 New Orleans House Galaxy Case & $65.5$ & $0.0$ \\
Portrait of Tiger in black and white by Lukas Holas & $50.0$ &  $0.0$ \\
VAN GOGH CAFE TERASSE copy.jpg & $56.5$ &  $1.5$ \\
Captain Marvel Exclusive Ccxp Poster Released Online By Marvel &  $95.0$ &  $0.5$ \\
Sony Boss Confirms Bloodborne Expansion is Coming & $83.5$  & $0.5$ \\
Ann Graham Lotz & $26.5$  & $0.0$ \\
$<$i$>$The Long Dark$<$/i$>$ Gets First Trailer, Steam Early Access & $100.0$  & $0.0$ \\
A painting with letter M written on it Canvas Wall Art Print & $4.0$  & $0.0$ \\ \cdashline{1-3}
\textbf{Average} & $60.1$ & $0.3$\\
\bottomrule
\vspace{-10pt}
\end{tabular}
}
\vspace{-7pt}
\caption{\textbf{Memorization rate.} We show the percentage of generated samples that are highly similar ($\geq 0.5$ cosine similarity on SSCD) to a ``memorized'' image.
}

\label{tbl:mem_percentage}
\vspace{-12pt}
\end{table}

\begin{figure}[!t]
    \centering
    \includegraphics[width=\linewidth]{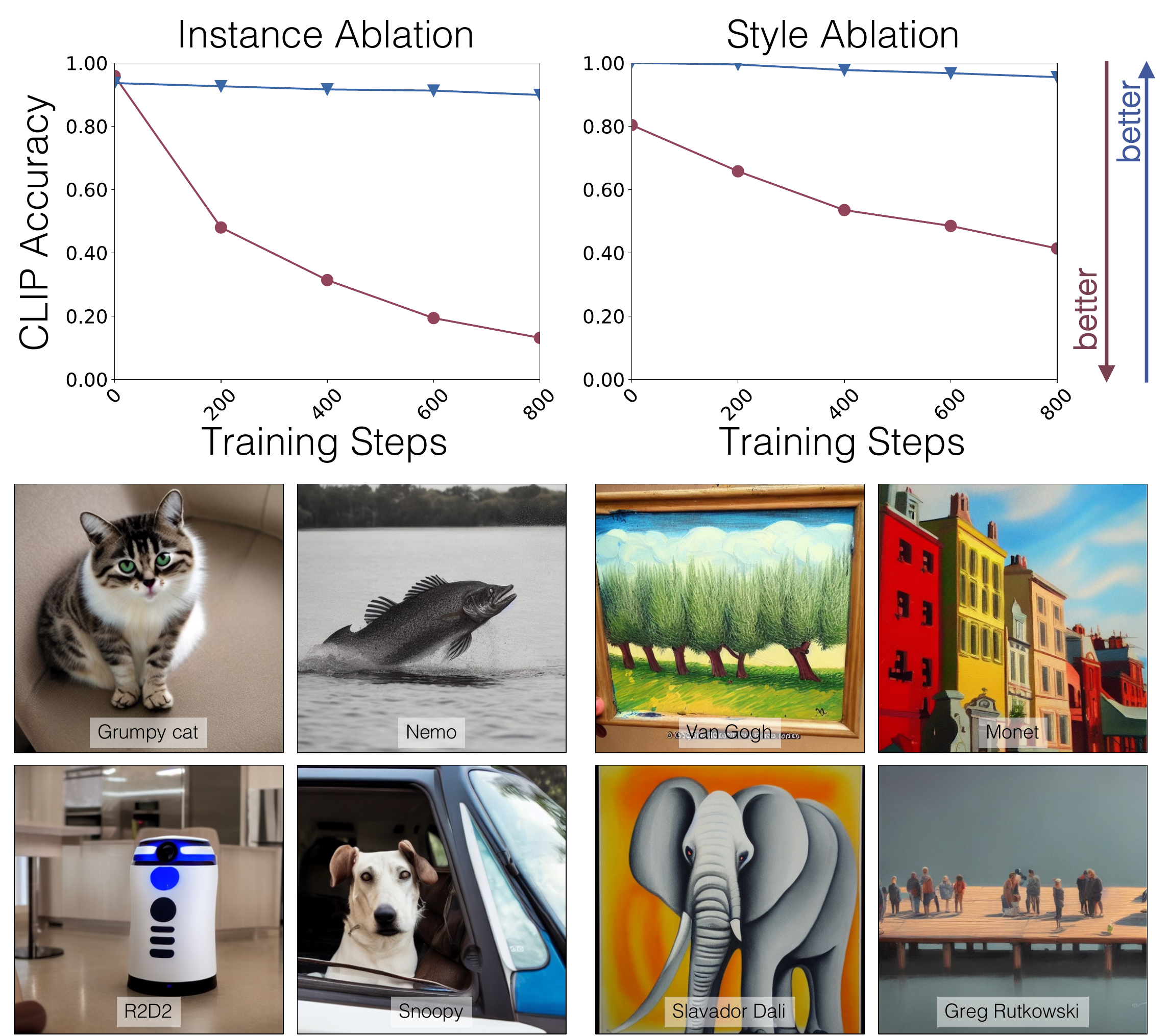}
    \vspace{-12pt}
    \caption{{\textbf{Ablating multiple instances (left) and style (right).} \textit{Top:} quantitative results show the drop in the CLIP Accuracy of the target concept, which has been ablated, whereas the accuracy for surrounding concepts remains the same. \textit{Bottom:} one sample image corresponding to each ablated target concept. 
    }}
    \lblfig{allinonemodel}
    \vspace{-7pt}
\end{figure}

\begin{figure}[!t]
    \centering
    \includegraphics[width=\linewidth]{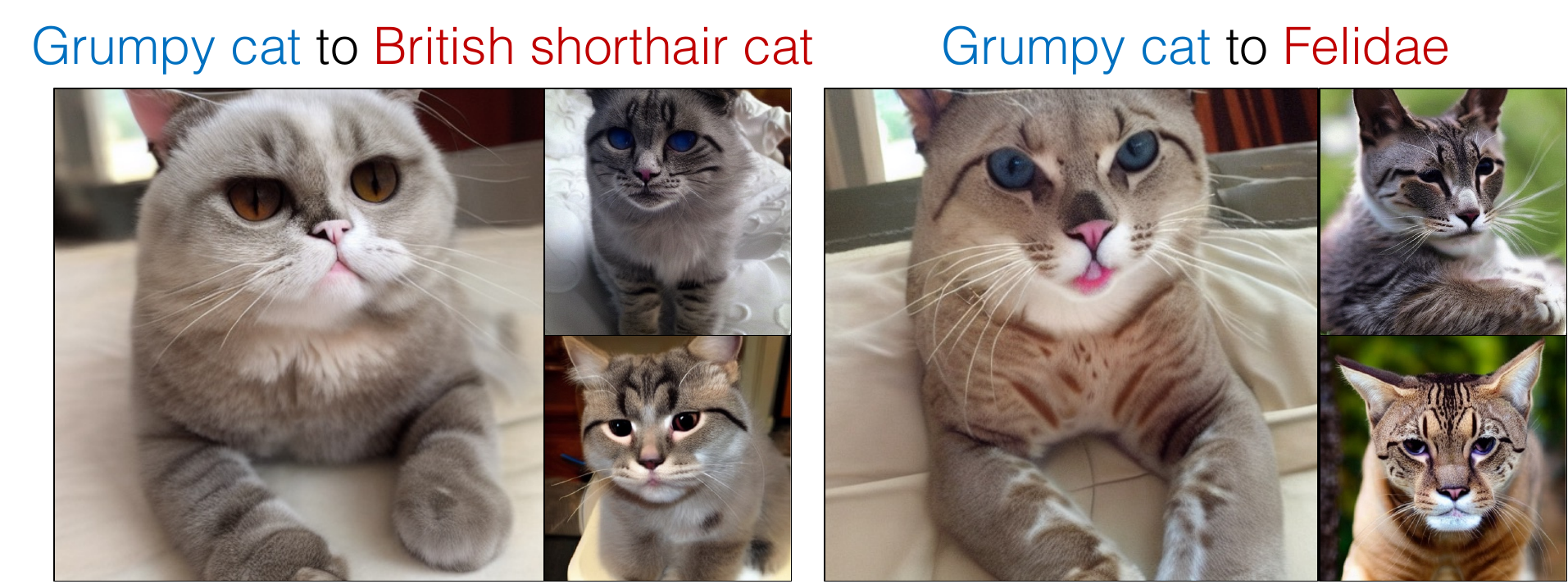}
    \vspace{-15pt}
    \caption{{\textbf{The choice of \anchor concepts.} Our method is robust to the choice of \anchor concepts. With both {\menlo British shorthair cat} and {\menlo Felidae} as \anchor concepts, our method can ablate the target {\menlo Grumpy Cat} concept.
    }}
    \lblfig{category_samples}
    \vspace{-7pt}
\end{figure}

\subsection{Additional Analysis}
\lblsec{other_experiments}

\myparagraph{Single model with multiple concepts ablated.}
Our method can also remove multiple concepts by training on the union of datasets for longer training steps. We show the results of one model with all instances and one model with all styles ablated in \reffig{allinonemodel}. We use the model-based variant of our method and cross-attention fine-tuning. More samples are shown in Appendix, \reffig{allinonemodel_instance_sample} and \ref{fig:allinonemodel_style_sample}. The drop in accuracy for the ablated concepts is similar to \reffig{results_instance} while maintaining the accuracy on surrounding concepts. 

\myparagraph{The role of \anchor category.}
In all the above experiments, we assume an \anchor category $\c^*$ is given to overwrite the target concept. Here, we investigate the role of choosing different \anchor categories for ablating {\menlo Grumpy Cat} and show results with the \anchor concept as {\menlo British Shorthair Cat} and {\menlo Felidae} in \reffig{category_samples}. Both \anchor concepts work well.

\myparagraph{Reverse KL divergence.} \nupur{In our $\methodmodel$ concept ablation, we optimize the KL divergence between the anchor concept and target concept distribution. Here, we compare it with optimizing the approximation to reverse KL divergence, i.e., $\mathbb{E}_{\epsilon,\x^*,\c^*, \c, t } [w_t||\hat{\Phi} (\x_t^*, \c, t).\text{sg()} - \hat{\Phi} (\x_t^*, \c^*, t) ||]$. Thus the expectation of loss is over target concept images. \reffig{gentarget} shows the quantitative comparison on ablating instances and style concepts. As we can see, it performs marginally better on ablating style concepts but worse on instances. In \reffig{gentarget_sample}, we show sample generations for the case where it outperforms the forward KL divergence based objective qualitatively on ablating {\menlo Van Gogh}.
}

\begin{figure}[!t]
    \centering
    \includegraphics[width=\linewidth]{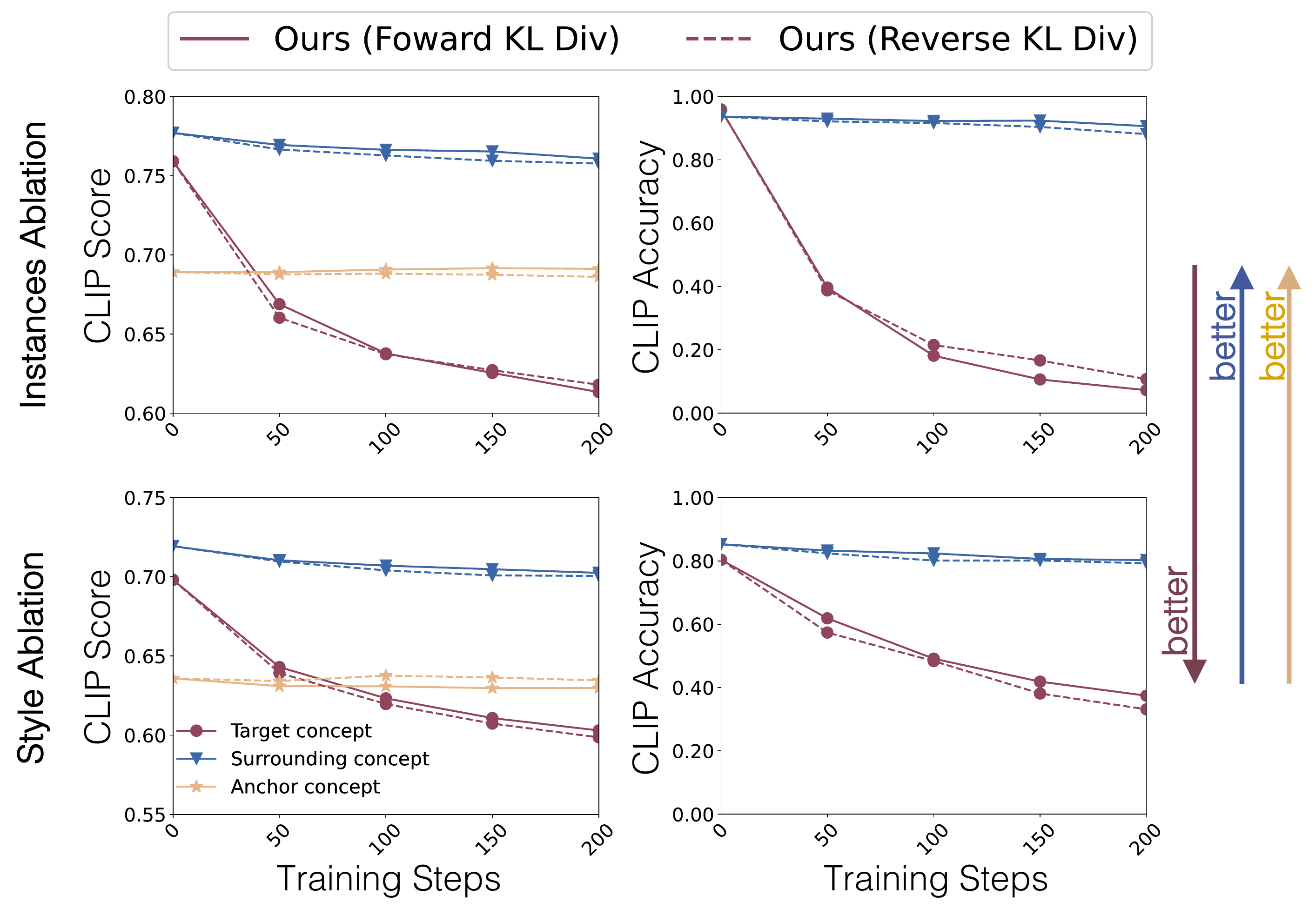}
    \vspace{-20pt}
    \caption{{\textbf{Reverse KL divergence objective.} \nupur{We show the results of optimizing the loss over target concept images for ablating instances (top) and style (bottom). Compared to using anchor concept images as training images, this performs slightly worse on ablating instances with lower CLIP Score on surrounding concepts while having similar CLIP Score on the target concept. It performs marginally better on ablating styles.}
    }}
    \lblfig{gentarget}
    \vspace{-4pt}
\end{figure}

\begin{figure}[!t]
    \centering
    \includegraphics[width=\linewidth]{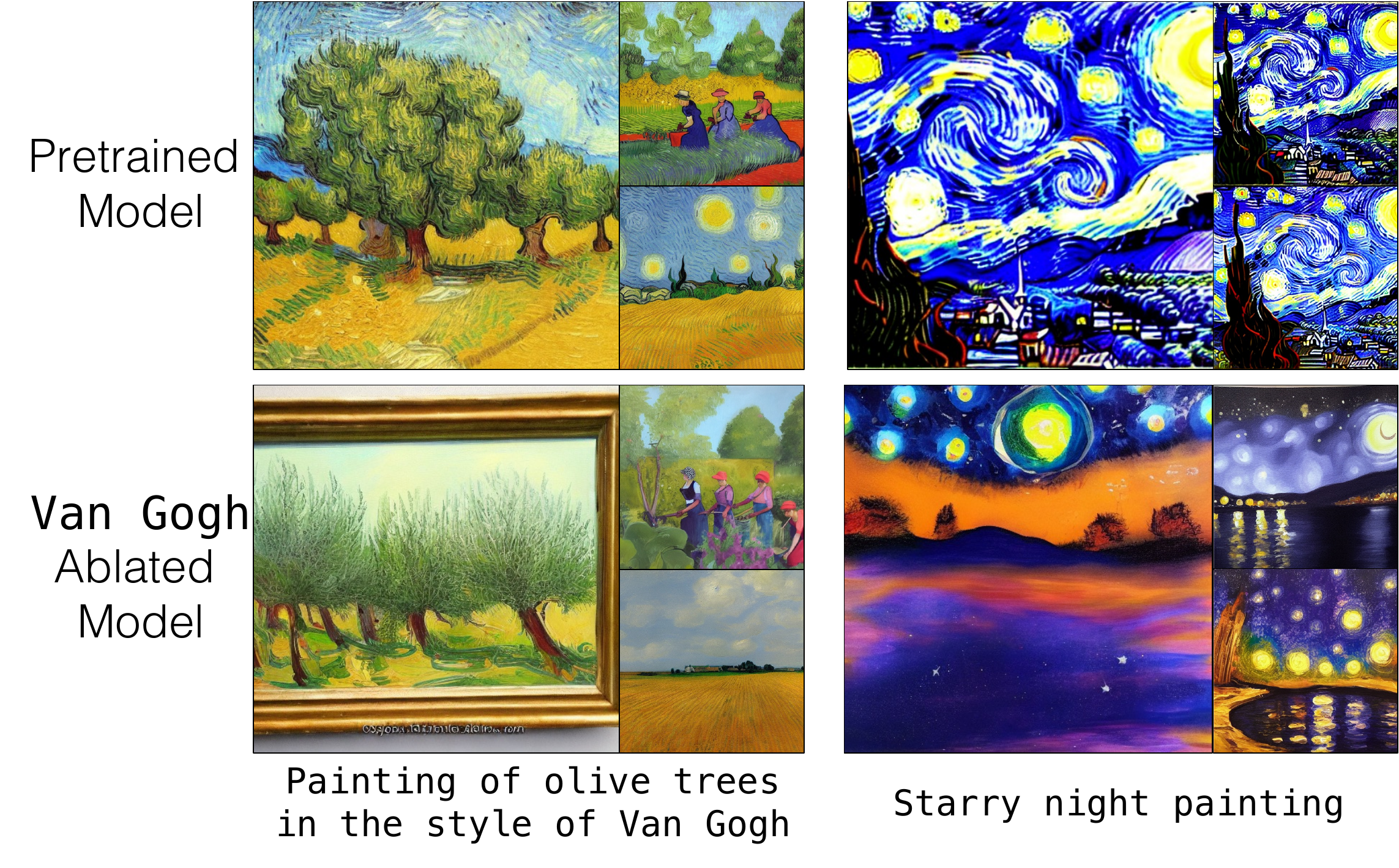}
    \vspace{-18pt}
    \caption{{ \textbf{Qualitative samples with reverse KL divergence objective.} \nupur{It performs better on certain styles and can successfully ablate famous paintings as well which is not achievable with forward KL divergence based objective and requires additional steps as shown in \reffig{limitation}.}
    } 
    }
    \lblfig{gentarget_sample}
    \vspace{-15pt}
\end{figure}

\begin{figure}[!t]
    \centering
    \includegraphics[width=\linewidth]{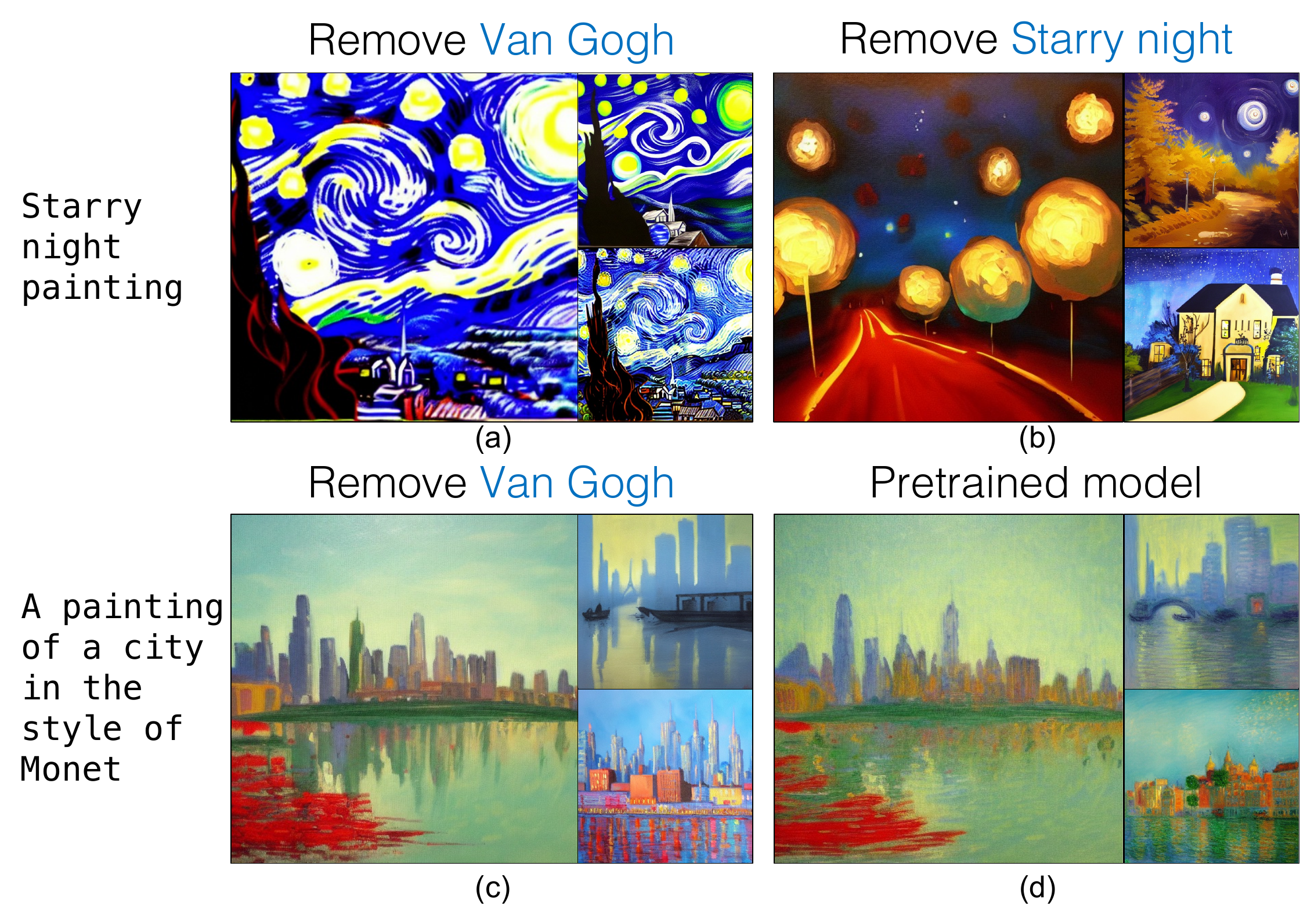}
    \vspace{-20pt}
    \caption{{ \textbf{Limitations.} \textit{Top:} (a) our method fails to remove certain paintings generated with the painting's titles. %
    (b) We can further ablate these concepts. \textit{Bottom:} \nupur{Though our method is better than baseline in preserving surrounding concepts as shown in \reffig{loss_maximize_sample}, the generated samples still sometimes show degradation for surrounding concepts, e.g., {\menlo Monet} (c) when ablating {\menlo Van Gogh} as compared to the pretrained model (d). }
    } 
    }
    \lblfig{limitation}
    \vspace{-8pt}
\end{figure}

\section{Discussion and Limitations}

Although we can ablate concepts efficiently for a wide range of object instances, styles, and memorized images, our method is still limited in several ways. First, while our method overwrites a target concept, this does not guarantee that the target concept cannot be generated through a different, distant text prompt. We show an example in \reffig{limitation} (a), where after ablating {\menlo Van Gogh}, the model can still generate {\menlo starry night painting}. However, upon discovery, one can resolve this by explicitly ablating the target concept {\menlo starry night painting}. Secondly, when ablating a target concept, we still sometimes observe slight degradation in its surrounding concepts, as shown in \reffig{limitation} (c). 

\nupur{Our method does not prevent a downstream user with full access to model weights from re-introducing the ablated concept~\cite{ruiz2022dreambooth,kumari2022multi,gal2022image}. Even without access to the model weights, one may be able to iteratively optimize for a text prompt with a particular target concept. Though that may be much more difficult than optimizing the model weights, our work does not guarantee that this is impossible.}

Nevertheless, we believe every creator should have an ``opt-out'' capability. We take a small step towards this goal, creating a computational tool to remove copyrighted images and artworks from large-scale image generative models.

\myparagraph{Acknowledgment.}
We are grateful to Gaurav Parmar, Daohan Lu, Muyang Li, Songwei Ge, Jingwan Lu, Sylvain Paris, and Bryan Russell for their helpful discussion, and to Aniruddha Mahapatra and Kangle Deng for proofreading the draft. The work is partly supported by Adobe Inc. 

{\small
\bibliographystyle{ieee_fullname}
\bibliography{main}
}
\appendix
\renewcommand{\thefootnote}{\arabic{footnote}}

\clearpage
\noindent{\Large\bf Appendix}
\vspace{5pt}

\myparagraph{Overview.} 
In \refsec{loss_objective}, we show a detailed derivation of the $\methodmodel$ concept ablation algorithm. In \refsec{application}, we present \emph{compositional} concept ablation, where we ablate the composition of two concepts while retaining individual concepts. We then show more analysis on varying other parameters in our method in \refsec{analysis_supp}. Finally, we include more samples for all our models in \refsec{samples} and discuss implementation details in \refsec{implementation_details}. All experiments are with $\methodmodel$ variant of our method with cross-attention fine-tuning unless mentioned otherwise.

\section{Model-based concept ablation objective}\lblsec{loss_objective}
We show here that minimizing the KL divergence objective between the joint distribution of noisy latent variables conditioned on anchor and target concept, i.e., \refeq{loss} in the main paper, can be reduced to the $\ell_2$ difference between the predicted noise vectors. 

\vspace{-10pt}
\begin{equation}
    \begin{aligned}
      & \mathcal{D_{KL}}(p_{\Phi}(\x_{(0 ...T)}|\c) || p_{\hat{\Phi}}(\x_{(0 ...T)} | \c^*)) \\ 
      & =   \mathbb{E}_{p_{\Phi}(\x_0 ... \x_T)}\log \frac{ \prod_{t=1}^{T}p_{\Phi}(\x_{t-1}|\x_t,\c)p_{\Phi}(\x_T) }{\prod_{t=1}^{T}p_{\hat{\Phi}}(\x_{t-1}|\x_t,\c^*)p_{\hat{\Phi}}(\x_T)} \\
      &  = \sum_{\hat{t}=1}^T  \mathbb{E}_{p_{\Phi}(\x_0 ... \x_T)}\log \frac{ p_{\Phi}(\x_{\hat{t}-1}|\x_{\hat{t}},\c) }{p_{\hat{\Phi}}(\x_{\hat{t}-1}|\x_{\hat{t}},\c^*)}
\end{aligned}
\end{equation}

We expand the term corresponding to a particular time step $\hat{t}$, i.e.,
\begin{equation*}
    \begin{aligned}
    & \mathbb{E}_{p_{\Phi}(\x_0 ... \x_T)}\log \frac{ p_{\Phi}(\x_{\hat{t}-1}|\x_{\hat{t}},\c) }{p_{\hat{\Phi}}(\x_{\hat{t}-1}|\x_{\hat{t}},\c^*)}\\
    & = \mathop{\int}_{\x_{(0 ...T)}}\hspace{-0.2cm}\prod_{t=1}^{T}p_{\Phi}(\x_{t-1}|\x_t,\c)p(\x_T) \log \frac{ p_{\Phi}(\x_{\hat{t}-1}|\x_{\hat{t}},\c)}{p_{\hat{\Phi}}(\x_{\hat{t}-1}|\x_{\hat{t}},\c^*)} d\x_{(0...T)} \\
    & = \mathop{\int}_{\x_{(\hat{t} ...T)}} p_{\Phi}(\x_{(\hat{t} ...T)}|\c) \Biggr [ \mathop{\int}_{\x_{(0 ... {\hat{t}-1}})} \prod_{t=1}^{\hat{t}}p_{\Phi}(\x_{t-1}|\x_t,\c) \\
    & \hspace{2.5cm}\log \frac{ p_{\Phi}(\x_{\hat{t}-1}|\x_{\hat{t}},\c)}{p_{\hat{\Phi}}(\x_{\hat{t}-1}|\x_{\hat{t}},\c^*)} d\x_{(\hat{t}-1...0)} \Biggl ] d\x_{(\hat{t}...T)}\\
     & = \mathop{\int}_{\x_{\hat{t}}}p_{\Phi}(\x_{\hat{t}}|\c) 
     \Biggr [ \mathop{\int}_{\x_{(0 ... {\hat{t}-1}})} (\prod_{t=1}^{\hat{t}-1}p_{\Phi}(\x_{t-1}|\x_t,\c) )p_{\Phi}(\x_{\hat{t}-1}|\x_{\hat{t}}, \c )  \\ 
     & \hspace{3cm} \log \frac{ p_{\Phi}(\x_{\hat{t}-1}|\x_{\hat{t}},\c)}{p_{\hat{\Phi}}(\x_{\hat{t}-1}|\x_{\hat{t}},\c^*)} d\x_{(\hat{t}-1...0)} \Biggl ] d\x_{\hat{t}} 
\end{aligned}
\end{equation*}
\begin{equation*}
    \begin{aligned}
    & = \mathop{\int}_{\x_{\hat{t}}}p_{\Phi}(\x_{\hat{t}}|\c) 
     \Biggr [ \mathop{\int}_{\x_{\hat{t}-1}} p_{\Phi}(\x_{\hat{t}-1}|\x_{\hat{t}}, \c )   
     \log \frac{ p_{\Phi}(\x_{\hat{t}-1}|\x_{\hat{t}},\c)}{p_{\hat{\Phi}}(\x_{\hat{t}-1}|\x_{\hat{t}},\c^*)} \\
     & \hspace{1.5cm} \Big [ \mathop{\int}_{\x_{(0 ... {\hat{t}-2}})}\prod_{t=1}^{\hat{t}-1}p_{\Phi}(\x_{t-1}|\x_t,\c) d\x_{(\hat{t}-2...0)}  \Big ] d\x_{\hat{t}-1} \Biggl ] d\x_{\hat{t}} 
\end{aligned}
\end{equation*}
The integral over $d\x_{(\hat{t}-2...0)}$ will be $1$ since it is an integration of the probability distribution over the range it is defined. Thus the previous term can be re-written as,
\begin{equation*}
\begin{aligned}
     & \mathop{\mathbb{E}}_{\x_{\hat{t}} \sim p_{\Phi}(\x_{\hat{t}}|\c) } 
     \Biggr [ \mathop{\int}_{\x_{\hat{t}-1}} p_{\Phi}(\x_{\hat{t}-1}|\x_{\hat{t}}, \c )   
     \log \frac{ p_{\Phi}(\x_{\hat{t}-1}|\x_{\hat{t}},\c)}{p_{\hat{\Phi}}(\x_{\hat{t}-1}|\x_{\hat{t}},\c^*)} d\x_{\hat{t}-1}\Biggl ]  \\
    & = \mathop{\mathbb{E}}_{\x_{\hat{t}} \sim p_{\Phi}(\x_{\hat{t}}|\c) } 
     \Biggr [ \mathcal{D_{KL}}( p_{\Phi}(\x_{\hat{t}-1}|\x_{\hat{t}}, \c )   
     || p_{\hat{\Phi}}(\x_{\hat{t}-1}|\x_{\hat{t}},\c^*) )\Biggl ]  \\
     & = \mathop{\mathbb{E}}_{\x_{\hat{t}} \sim p_{\Phi}(\x_{\hat{t}}|\c) } 
     \Big [ \eta (\Phi(\x_{\hat{t}}, \c, t) - \hat{\Phi}(\x_{\hat{t}}, \c^*, t))^2 \Big ]  \\
    \end{aligned}
\end{equation*}

In the case of the diffusion model, each conditional distribution, $p_{\Phi}(\x_{\hat{t}-1}|\x_{\hat{t}},\c)$ and $p_{\hat{\Phi}}(\x_{\hat{t}-1}|\x_{\hat{t}},\c^*)$, is a normal distribution with fixed variance and mean as a linear combination of $\x_t$ and the predicted noise. Above we use this fact and that KL divergence between two normal distributions simplifies to the squared difference between the mean. We ignore the variance terms in the KL divergence as it is not learned.

\begin{figure*}[!t]
    \centering
    \includegraphics[width=\linewidth]{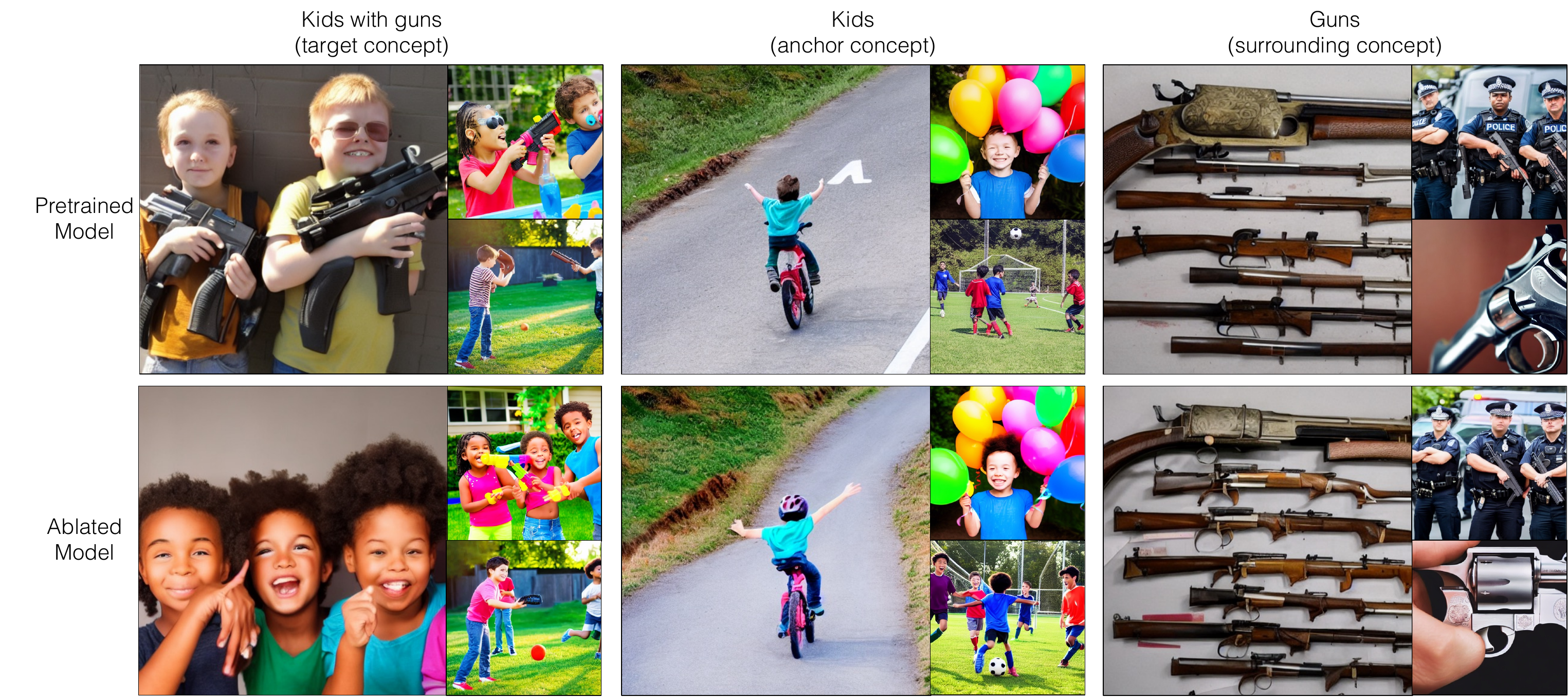}
    \vspace{-15pt}
    \caption{{ \textbf{Ablating composition of concepts.} Our method can remove the composition of ``kids with guns'' while preserving individual category kids and guns. 
    } 
    }
    \lblfig{compose_removal}
    \vspace{-15pt}
\end{figure*}

\section{Compositional Concept Ablation}\lblsec{application}
In this section, we show that our method can be used to ablate the composition of two concepts while still preserving the meaning of each concept. For example, we show results with ablating {\menlo kids with guns}. The training dataset $(\x, \c, \c^*)$ now consists of images generated using prompts with {\menlo kids}, i.e., anchor concept prompts and target concept prompt of {\menlo kids with guns}. In this case, we add a standard diffusion regularization loss on images corresponding to {\menlo kids} and {\menlo guns} individually. 

\myparagraph{Results.} \reffig{compose_removal} shows sample generations for both ours and pretrained model given the prompts for target concept and anchor concepts. As we can see, our method successfully ablated the {\menlo kids with guns} concept and only generates {\menlo kid} images given that prompt. For the anchor concept, {\menlo gun} and {\menlo kids}, sample images are similar to the one generated by the pretrained model. The CLIP Score between generated images from the fine-tuned model with {\menlo kids with guns} prompts and CLIP text feature {\menlo kids} is $0.62$ which is similar to the baseline score of $0.63$. For {\menlo guns}, it is $0.52$, which is significantly lower than the baseline model's score of $0.60$. Thus the {\menlo kids with guns} target concept has been successfully ablated in the fine-tuned model.

\section{Additional analysis}\lblsec{analysis_supp}

\begin{figure}[!t]
    \centering
    \includegraphics[width=\linewidth]{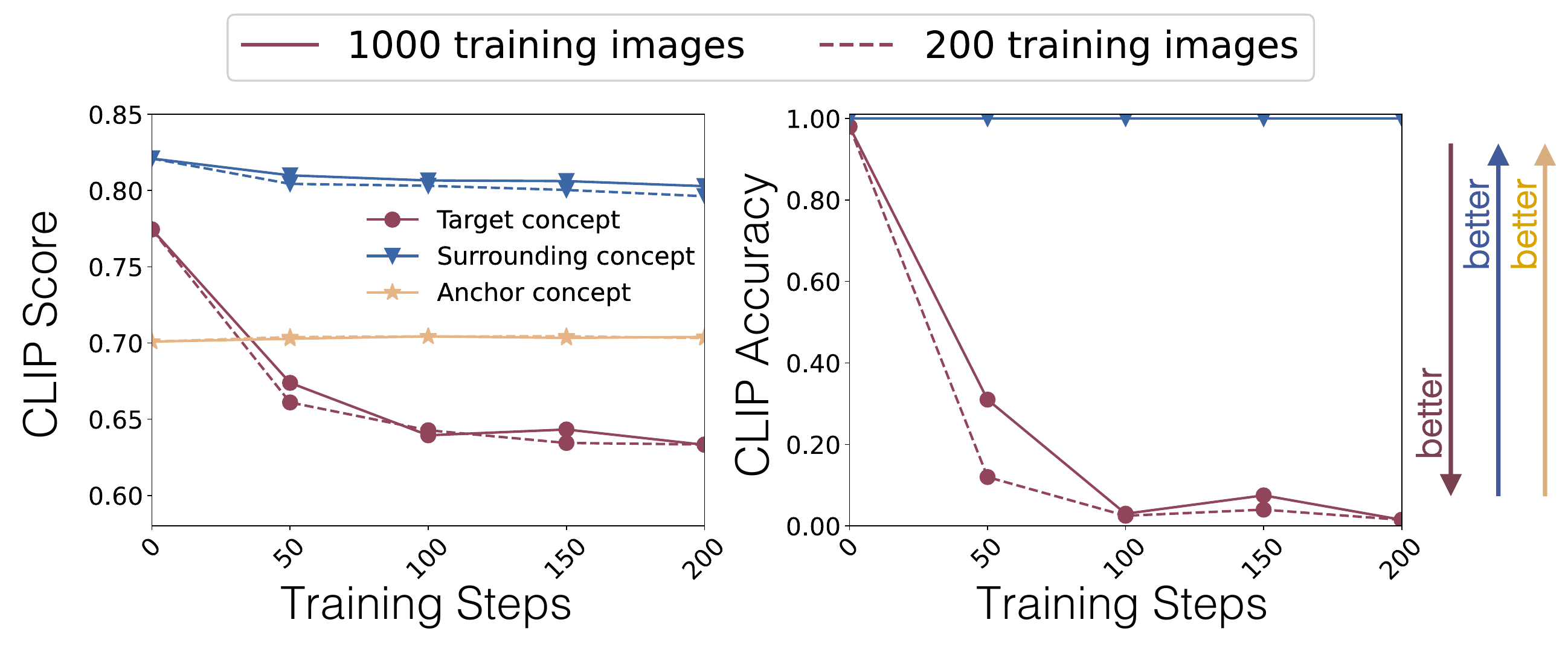}
    \vspace{-20pt}
    \caption{{\textbf{Number of training images.} We analyze the effect of varying numbers of training images when ablating {\menlo Grumpy cat}. As we can see, training with $200$ images results in a similar performance on target concept by convergence (100 training steps) but is marginally worse on surrounding concepts. 
    }}
    \lblfig{num_images}
\end{figure}

\begin{figure}[!t]
    \centering
    \includegraphics[width=\linewidth]{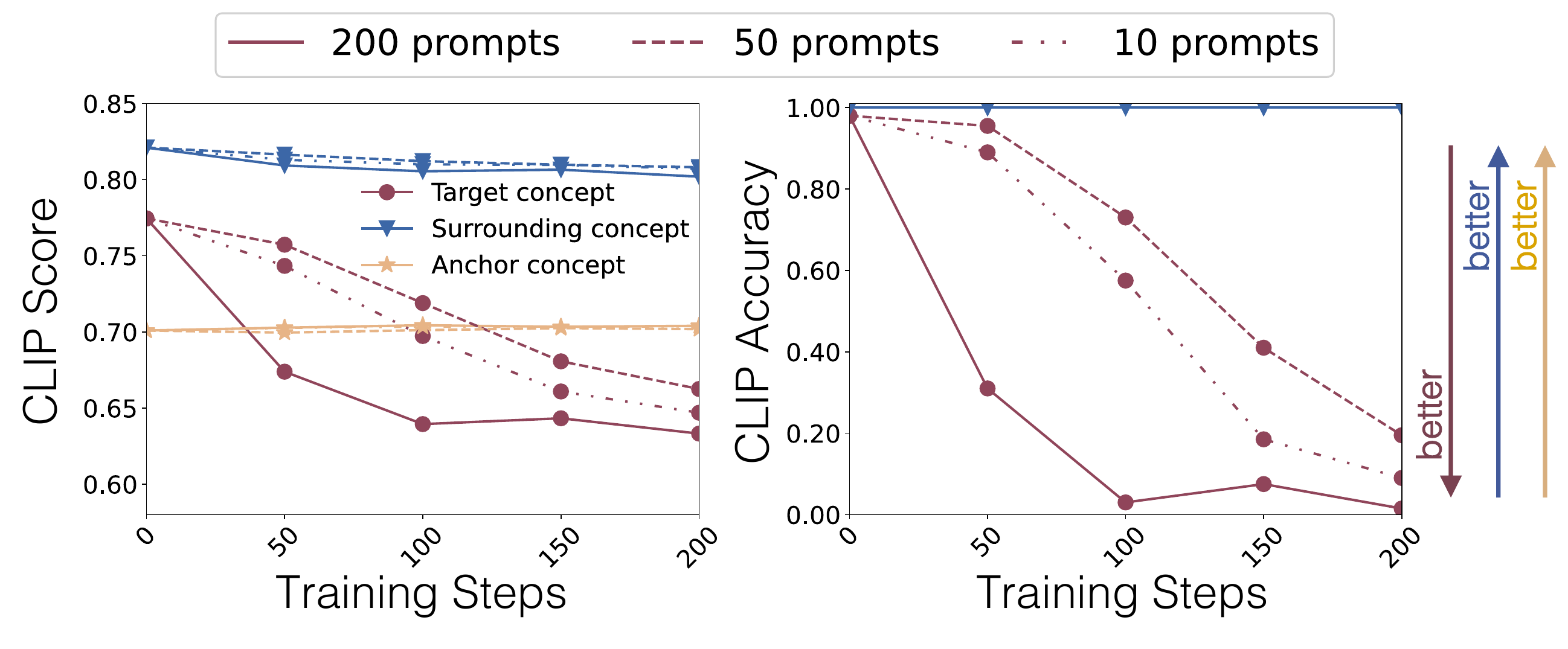}
    \vspace{-10pt}
    \caption{{\textbf{Number of unique prompts.} We compare using only $50$ and $10$ prompts for generating the $1000$ training images with our standard setting of $200$ prompts on ablating {\menlo Grumpy Cat}. Using fewer prompts leads to slower convergence. 
    }}
    \lblfig{unique_quant}
    \vspace{-10pt}
\end{figure}

\begin{figure}[!t]
    \centering
    \includegraphics[width=\linewidth]{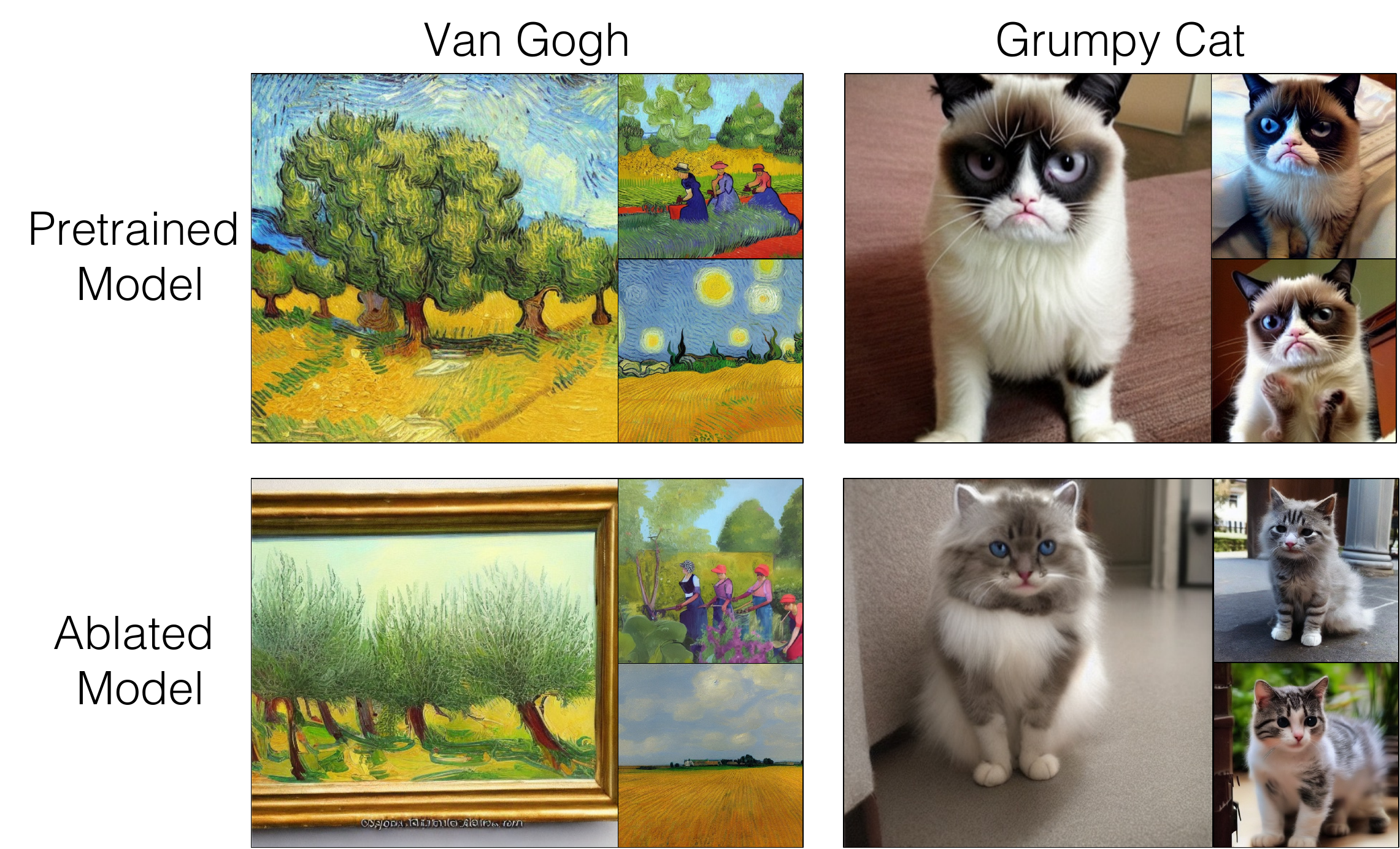}
    \vspace{-18pt}
    \caption{{\textbf{Qualitative samples on using real target concept images in training.} Our method can successfully ablate target concepts when given target concept images and their corresponding captions. But this requires manually labeling the images with correct prompts to get $\c^*$ and modifying it to get the corresponding anchor prompt $\c$. Thus, we do not use this as our standard setup. 
    } 
    }
    \lblfig{realtarget}
    \vspace{-4pt}
\end{figure}

\myparagraph{Number of training images.} In all the experiments, we typically generate $1000$ images as the training data. \reffig{num_images} shows the comparison of training with $200$ and $1000$ images. We observe that training on just $200$ images performs only slightly worse on surrounding concepts. We also experimented with increasing the number of images to $10$k from $1000$ but observed similar performance. This indicates that performance saturates and $1000$ images are sufficient.

\myparagraph{Number of unique prompts} 
Here, we analyze the effect of the number of unique prompts used in training. We vary the number of prompts to $10$ and $50$ and generate $1000$ training images using the prompts. We show its results on ablating {\menlo Grumpy Cat} in \reffig{unique_quant}. As we can see, convergence is faster when using more variations in the prompts.

\myparagraph{Real target concept images with reverse KL divergence.}
To reiterate, our $\methodmodel$ variant loss is $\mathbb{E}_{\epsilon,\x,\c^*, \c, t } [w_t||\hat{\Phi} (\x_t, \c, t).\text{sg()} - \hat{\Phi} (\x_t, \c^*, t) ||]$, where $\x$ is an image corresponding to the anchor concept prompt $\c$ (e.g. {\menlo photo of a cat} when $\c^*$ is {\menlo photo of a grumpy cat}). Thus the training objective minimizes the difference in prediction between anchor prompts and target prompts over all possible noisy anchor concept images. \nupur{We discussed in \refsec{other_experiments} our approximation to reverse KL divergence objective, which optimizes the loss over target concept images, i.e., $\mathbb{E}_{\epsilon,\x^*,\c^*, \c, t } [w_t||\hat{\Phi} (\x_t^*, \c, t).\text{sg()} - \hat{\Phi} (\x_t^*, \c^*, t) ||]$.}
In the experiment, target concept images $\x^*$ are generated by the pretrained model. But it is also possible to use real target concept images with the above objective. We perform this experiment for ablating {\menlo Van Gogh} and {\menlo Grumpy Cat} using ten real images of each target concept and show its results in \reffig{realtarget}. It leads to slower convergence as in the case of {\menlo Grumpy Cat} but otherwise performs similarly.

\myparagraph{Comparison between the training objectives when fine-tuning different parameter subset}
In the main paper, we compared our concept ablation methods with the baseline method of maximizing the loss when fine-tuning \textit{Cross-Attention} parameters~\cite{kumari2022multi}. Here, we show the comparison when fine-tuning the \textit{Embedding}~\cite{gal2022image} and \textit{Full Weights}~\cite{ruiz2022dreambooth} of the U-Net diffusion model. \reffig{dreambooth_loss_objective} and \ref{fig:embedding_loss_objective} show the results. In both these cases as well, our $\methodmodel$ variant performs better or on par with other methods. 

\begin{figure}[!t]
    \centering
    \includegraphics[width=\linewidth]{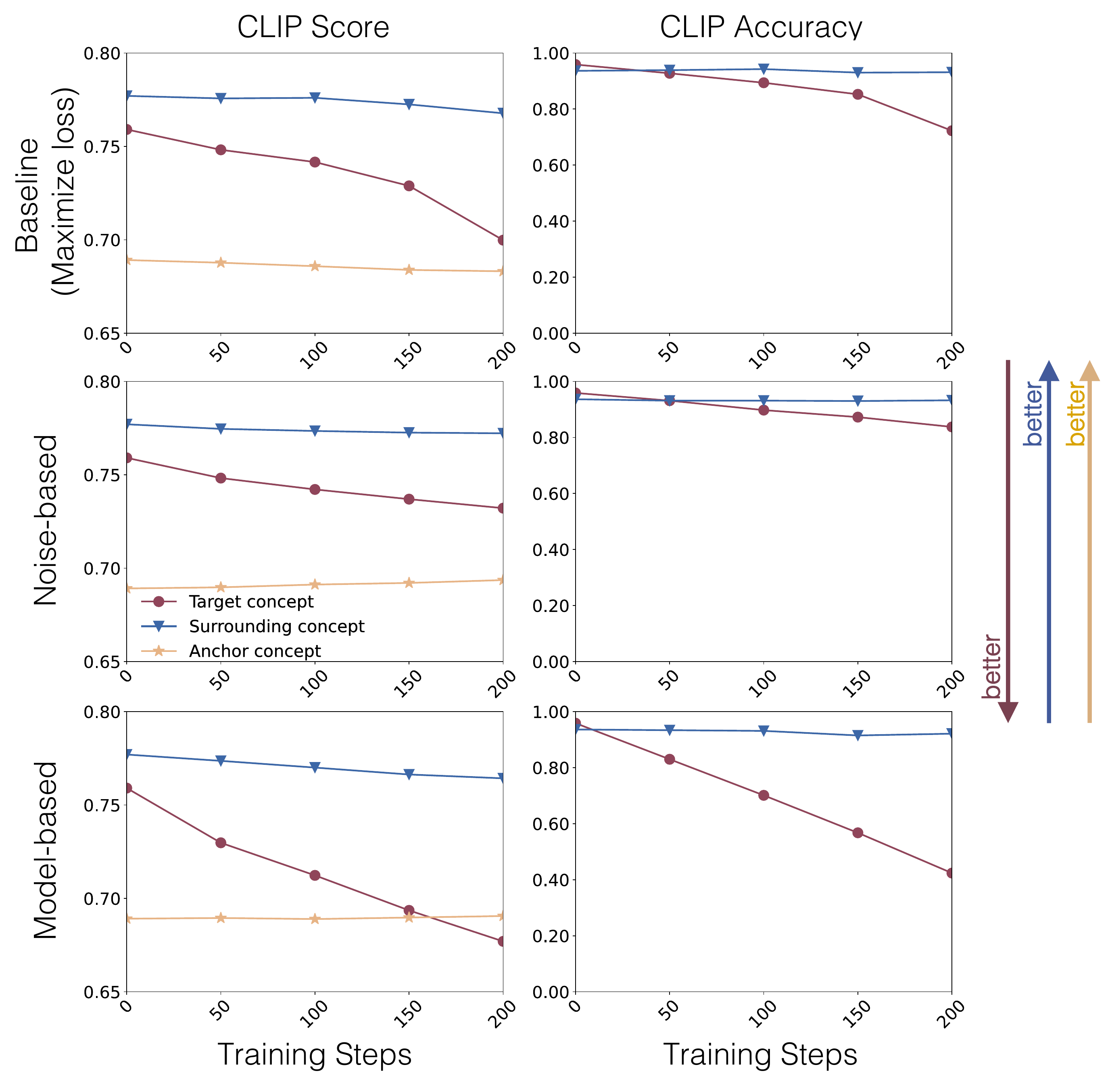}
    \vspace{-15pt}
    \caption{{\textbf{Comparison of different loss objective when fine-tuning \textit{Full Weights}.} The $\methodmodel$ variant performs better than the baseline and $\methoddata$ variant in this case as well, with faster convergence and maintaining the average CLIP Score and CLIP Accuracy on surrounding concepts.  
    }}
    \lblfig{dreambooth_loss_objective}
    \vspace{-5pt}
\end{figure}

\begin{figure}[!t]
    \centering
    \includegraphics[width=\linewidth]{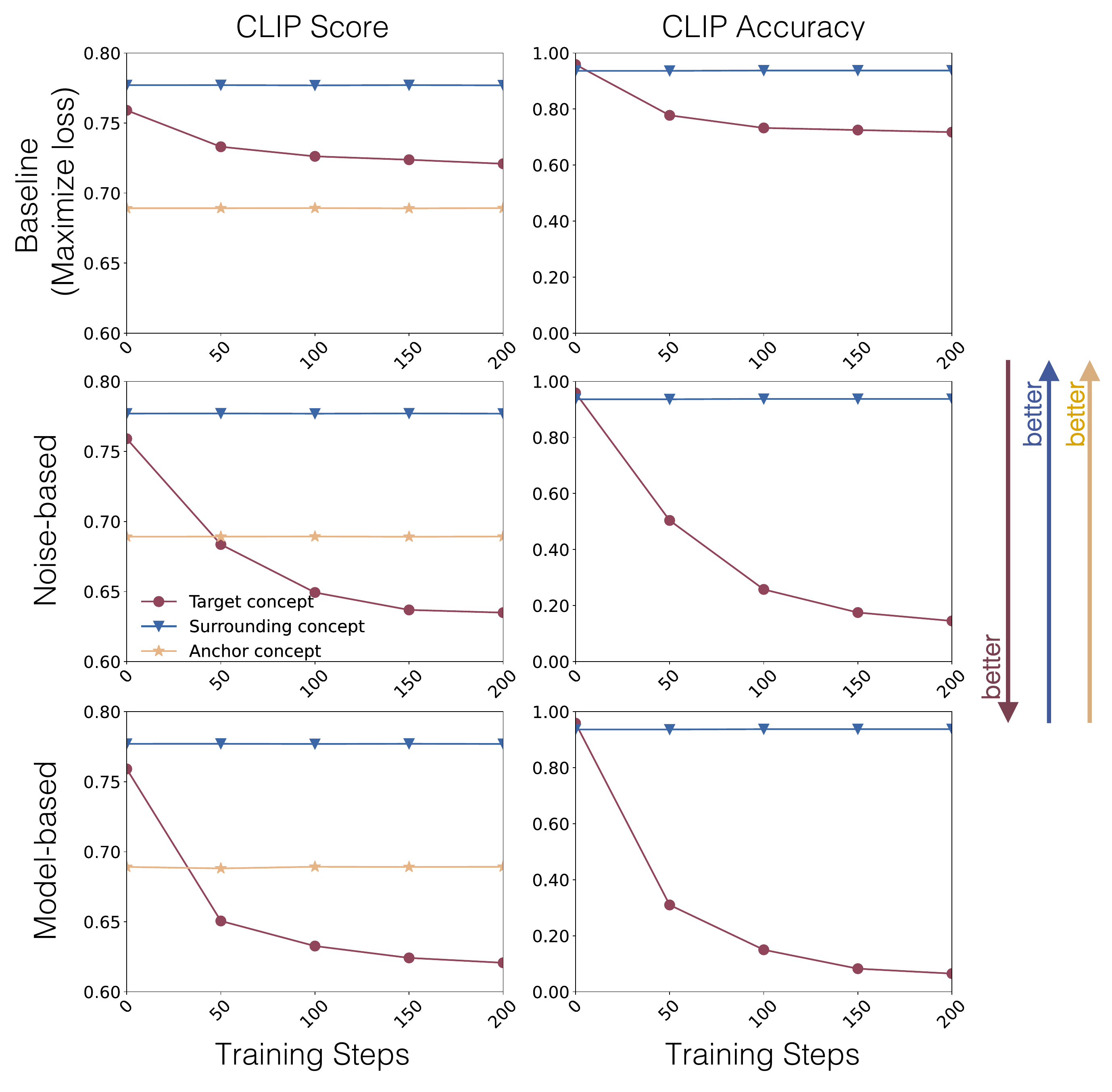}
    \vspace{-15pt}
    \caption{{\textbf{Comparison of different loss objectives when fine-tuning \textit{Embedding}.} In this case both $\methodmodel$ and $\methoddata$ variant peform similarly and better than the baseline. But as discussed in the main paper, fine-tuning embedding is not robust to small spelling mistakes and thus can still be used to generate the target concept.  
    }}
    \lblfig{embedding_loss_objective}
    \vspace{-5pt}
\end{figure}

\myparagraph{Comparison with negative prompts and Safe Latent Diffusion~\cite{schramowski2022safe}}. Figures \ref{fig:baseline_compare} and \ref{fig:baseline_compare_qual} show the comparison of ablating instances with the CLIP Score metric. Our method performs better on surrounding concepts while successfully ablating the target concept compared to these baselines. In the case of the negative prompt method and Safe Latent Diffusion (SLD), we assign the target concept to be the negative prompt or the safety concept, respectively.  

\begin{figure}[!t]
    \centering
    \includegraphics[width=0.98\linewidth]{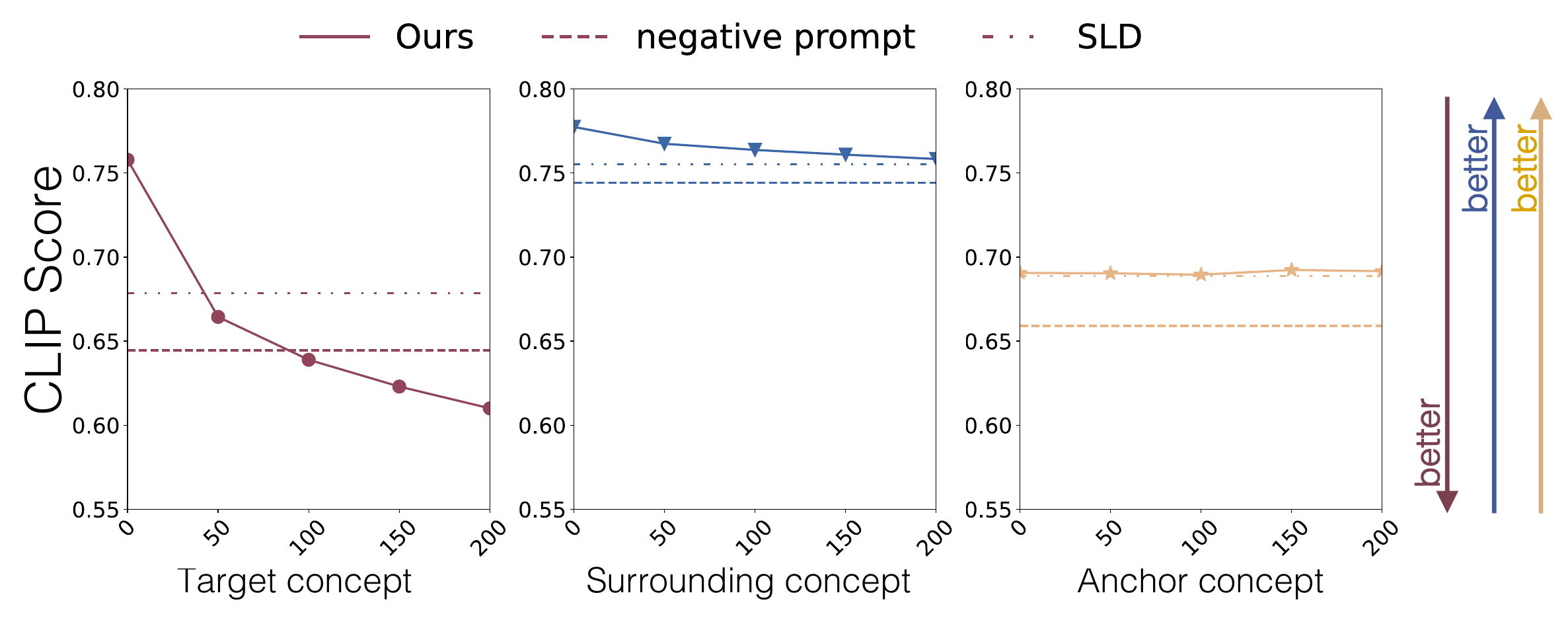}
    \vspace{-7pt}
    \caption{\textbf{Instance ablation comparison with Negative prompt and Safe Latent Diffusion (SLD).} Our method ablates the target concept while being most similar to the pre-trained model on anchor and surrounding concepts. We used the diffusers implementation for both with the same hyperparameters as recommended in the paper for SLD-Medium~\cite{schramowski2022safe}. 
    }
    \lblfig{baseline_compare}
    \vspace{-13pt}
\end{figure}

\begin{figure}[!t]
    \centering
    \includegraphics[width=\linewidth]{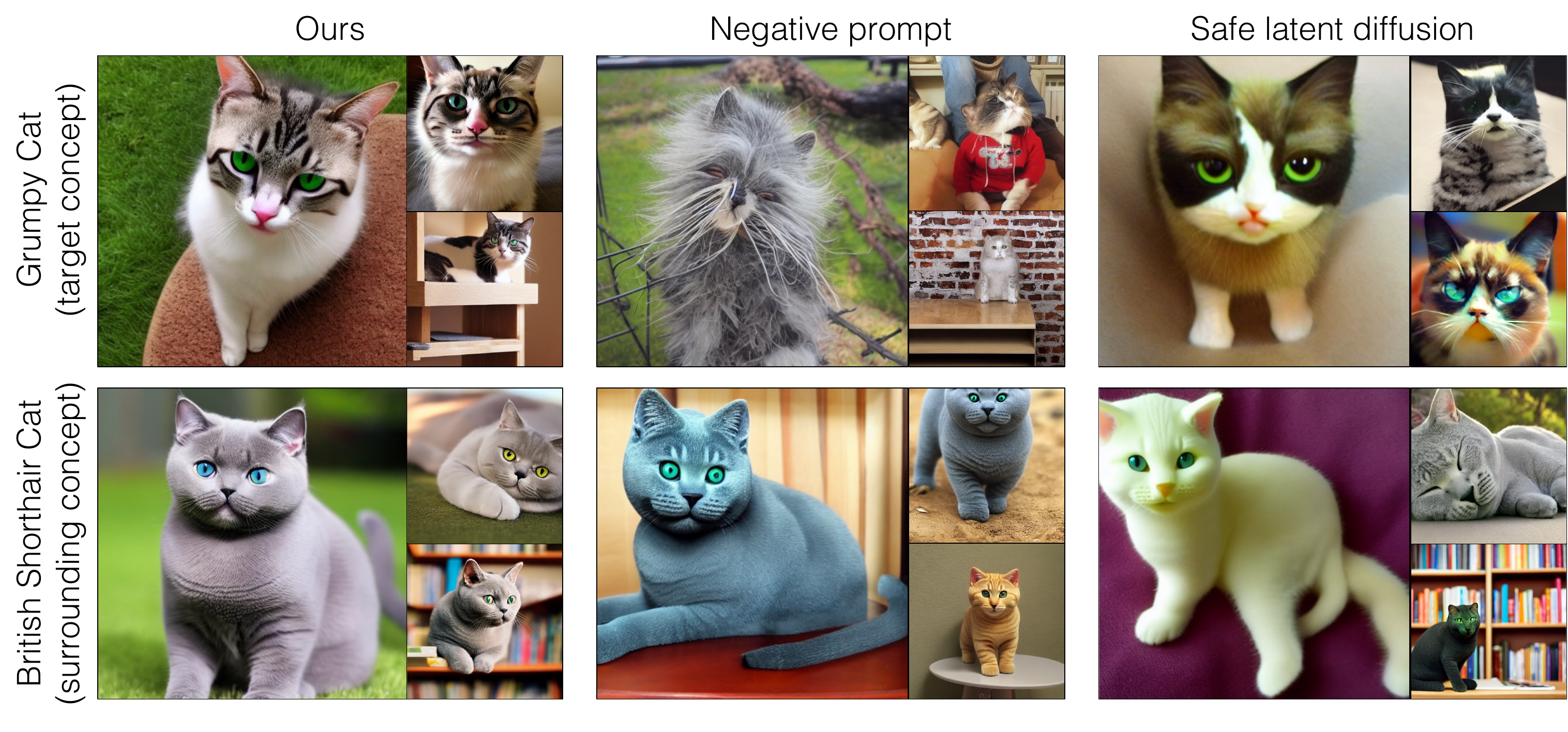}
    \vspace{-15pt}
    \caption{\textbf{Qualitative comparison with Negative prompt and Safe Latent Diffusion (SLD)}. Our method preserves the surrounding concept better compared to the baseline methods of negative prompt and SLD. We used the diffusers implementation for both with the same hyperparameters as recommended in the paper for SLD-Medium~\cite{schramowski2022safe}.
    }
    \lblfig{baseline_compare_qual}
    \vspace{-12pt}
\end{figure}

\myparagraph{Performance on unrelated concepts.} To ensure that ablating a specific concept from the model using our method doesn't affect its performance on unrelated concepts, we calculate the MSCOCO FID of all ablated models. The mean FID is $16.99 \pm 0.2$. This is close to the $16.35$ FID of the pretrained model. We computed the FID score using $30k$ randomly sampled images from the MSCOCO validation set and generated images corresponding to the same captions using $50$ steps of the DDPM sampler.

\myparagraph{Other alternatives to ChatGPT.} Our method uses ChatGPT to generate random prompts when ablating an instance. We also experimented with an open-source alternative, \href{https://arxiv.org/abs/2305.14314} {Qlora}, to generate these training prompts when ablating {\menlo Grumpy Cat}. The CLIP score on the target concept is similar to using ChatGPT ($0.651$ vs. $0.639$, the lower, the better). On surrounding concepts, the performance is similar ($0.801$ vs. $0.796$, the higher, the better).

\section{More qualitative samples}\lblsec{samples}
We show more qualitative samples of ablating memorized images, styles, and instances and their surrounding concepts. \reffig{mem_orleans}-\ref{fig:mem_vangogh} shows the samples generated by the pretrained model and fine-tuned models with memorized image ablated. We can see that compared to the pretrained model, our models generate significantly varying images given the target prompt. \reffig{allinonemodel_instance_sample} and \ref{fig:allinonemodel_style_sample} show the results of ablating multiple styles and instances, respectively. In \reffig{vangogh_allimages}-\ref{fig:salvador_allimages}, we show a qualitative comparison of style ablated models with the pretrained model on the target concept and surrounding concept images. Finally, \reffig{grumpy_allimages}-\ref{fig:snoopy_allimages} shows the qualitative comparison of instance ablated models with the pretrained model on the target concept and surrounding concept images.

\section{Implementation details}\lblsec{implementation_details}
We describe additional details for our method, baselines, and evaluation setup. Our code is built on top of Custom Diffusion repo~\footnote{https://github.com/adobe-research/custom-diffusion}. 

\myparagraph{Cross-Attention.} We train with a batch size of $8$ and learning rate $2\times 10^{-6}$ (scaled by the batch size). All qualitative samples are shown with $100$ training steps for our $\methodmodel$ variant, $200$ steps for the $\methoddata$ variant, and $50$ steps for the loss maximation baseline. To ablate multiple style or instaces from the model we fine-tune for longer iterations in the multiple of total ablated concepts.

\myparagraph{Embedding.} We train with a batch size of $8$ and learning rate $1\times 10^{-5}$ (scaled by the batch size). All qualitative samples are shown with $200$ training steps. 

\myparagraph{Full-weights.}
When fine-tuning all weights of the U-Net, training is done on batch-size $4$ instead of $8$ (because of increased memory requirement) with a learning rate of $5\times 10^{-7}$ (without any scaling with the batch size). All qualitative samples are shown with $200$ training steps for ablating style and instance concepts. In the case of ablating memorized images, we used $1\times 10^{-6}$ learning rate and $800$ training steps except for {\menlo Anne Graham Lotz} case for which we used the above default values. 

\myparagraph{Other details.} We add regularization loss on the anchor concept data, as explained in Section 3.2 in the main paper, with $\lambda=1$ in the case of ablating {\menlo Grumpy Cat} and memorized images. To obtain training images, we sample using the DDPM sampler with $200$ steps. When training the loss maximization baseline, the regularization on weights is added with a factor of $10$ (Eq.~\ref{eqn:baseline_tanno}, main paper). Similar to Custom-Diffusion~\cite{kumari2022multi}, our implementation detaches the first token of the text transformer output before input to the U-Net. We also use image augmentation similar to Custom-Diffusion~\cite{kumari2022multi} when ablating object instances. For different parameter subset fine-tuning, we select the learning rate which works the best. In the case of the $\methoddata$ variant of our method, we also tried increasing the learning rate for faster convergence, but it led to sub-optimal results with artifacts in generated images. All our experiments are done on 2 A6000 GPUs with 3 minutes per $100$ training step. For the CLIP Score metric, the standard error is less than $5 \times 10^{-3}$ in all cases. 

\myparagraph{Training and test set prompts.}
We used chatGPT to create training and test prompts for all object instances. The instruction to chatGPT~\cite{chatgpt} was: {\menlo provide 210 captions for images containing <anchor-concept>. The caption should also contain the word ``<anchor-concept>''.} Out of this first 200 captions were used to generate training images, and the remaining ten were used for evaluation purposes. Regarding style concepts, as mentioned in the main paper, we used clip-retrieval to collect $210$ captions. Out of this, $200$ prompts are used for training and $10$ for evaluating the anchor concept {\menlo painting}. For target and surrounding style concepts, we used image captioning (along with manual supervision) on real images corresponding to each style to create ten prompts for each style concept. All evaluation prompts are provided in \reftbl{prompts_style_eval} and \ref{tbl:prompts_instance_eval}. We also show the surrounding concept for each target concept in \reftbl{surrounding_concept}. For calculating CLIP Score and Accuracy metric when ablating style concepts, we use the text prompt as: {\menlo <target-concept> style}. 

For the eight memorization use cases, we again used chatGPT to create variations of the target concept prompt $\c$ using the instruction: {\menlo provide five captions for an image depicting <image description>}. For memorization, we observe that paraphrased text prompts also generate the memorized images with high probability. Therefore, we keep generating variations of the target concept prompt until we have five suggested prompts that generate copied images with less than $30\%$ probability. We manually inspect the suggested paraphrases to ensure they are coherent with the image. We show the paraphrases used for each case in \reftbl{prompts_mem}.

\begin{figure*}[!t]
    \centering
    \includegraphics[width=0.9\linewidth]{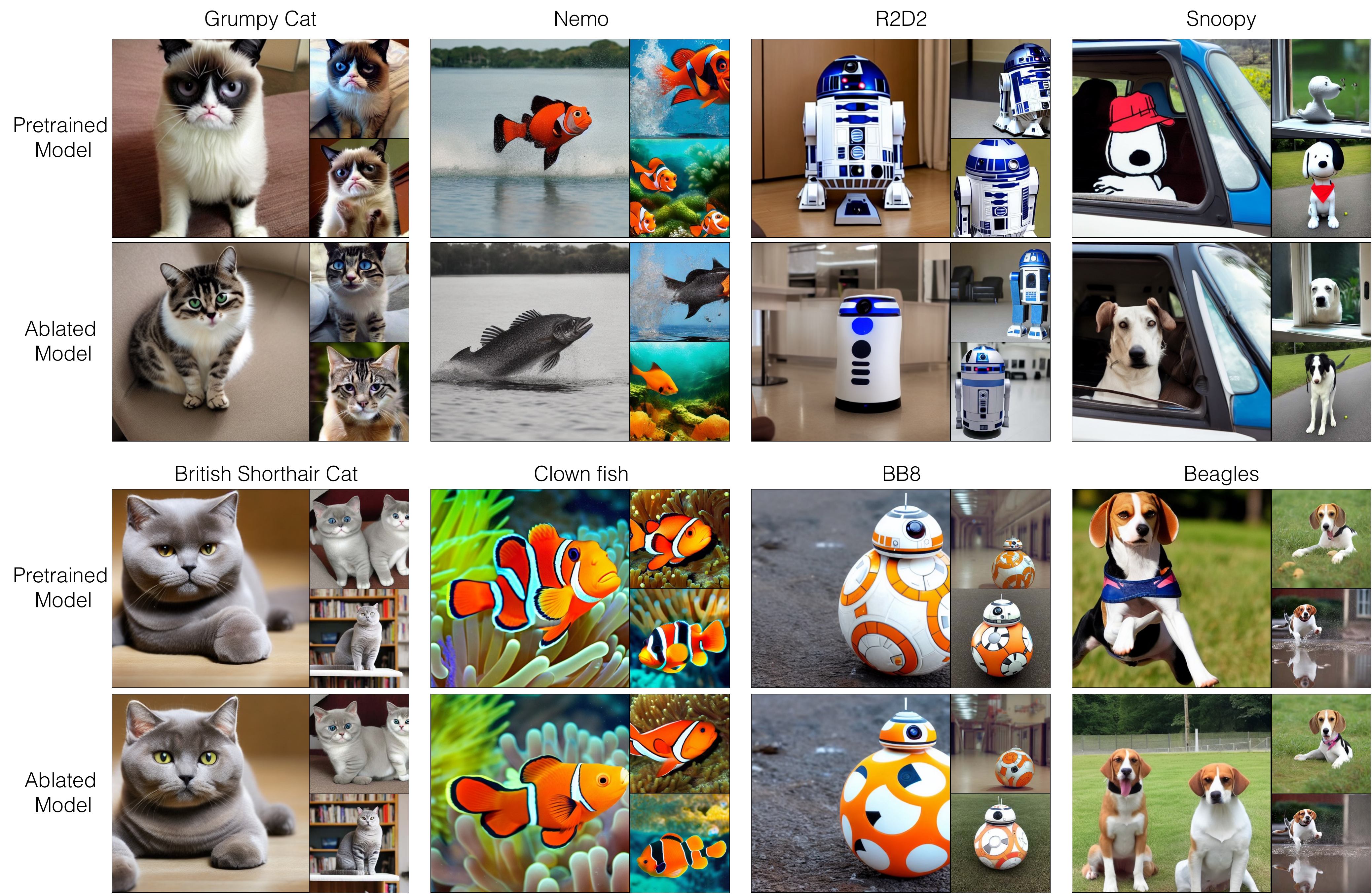}
    \vspace{-10pt}
    \caption{{ \textbf{Ablating multiple instances} 
    } Our method can be used to ablate multiple concepts. Here, we show the sample generations from a single model from which all four instances (top row) have been ablated. The bottom row shows sample images for surrounding concepts.
    }
    \lblfig{allinonemodel_instance_sample}
    \vspace{-10pt}
\end{figure*}

\begin{figure*}[!t]
    \centering
    \includegraphics[width=0.9\linewidth]{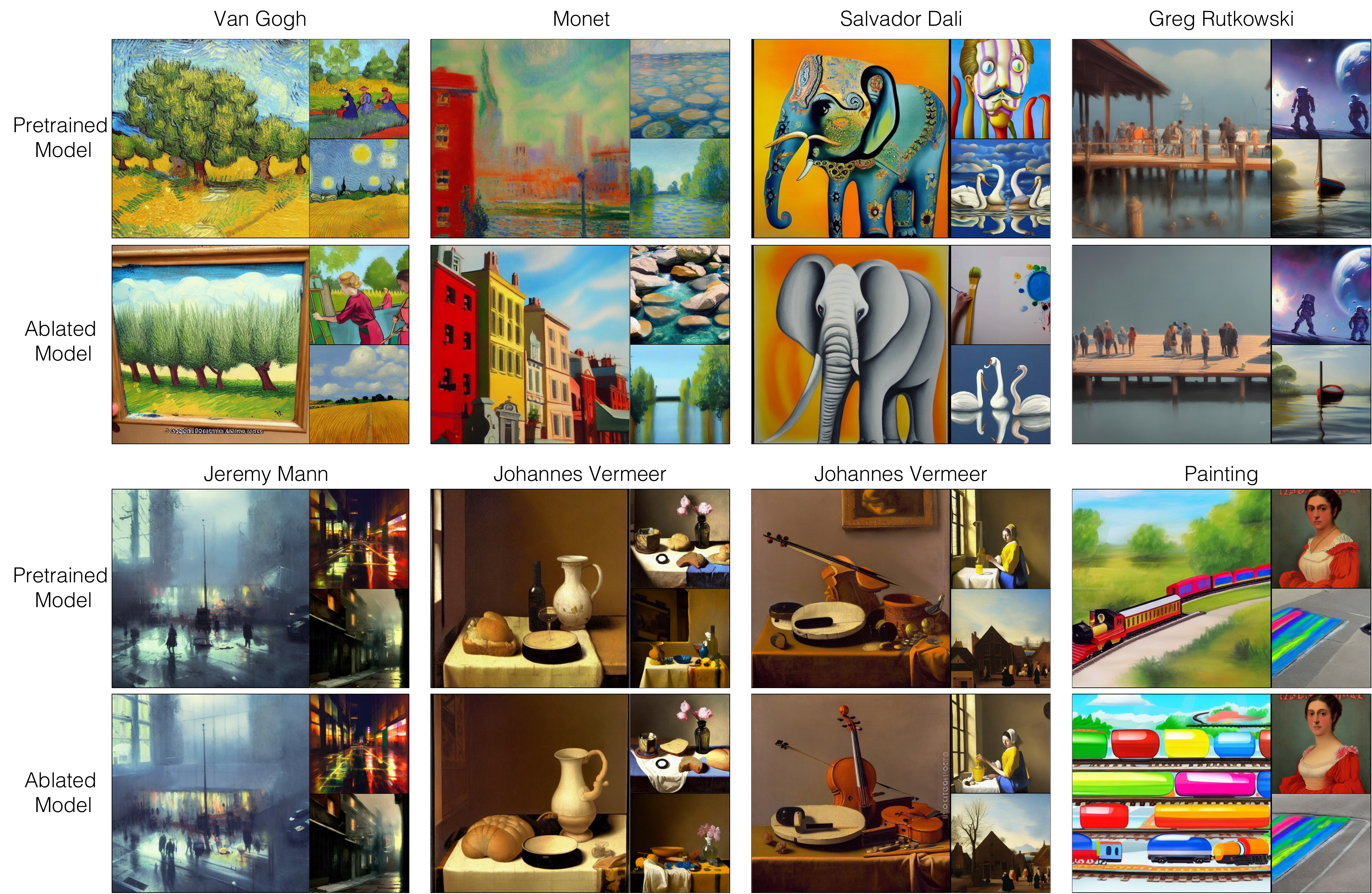}
    \vspace{-10pt}
    \caption{{ \textbf{Ablating multiple styles.} We show a qualitative comparison between the pretrained model and fine-tuned model with all four ablated styles (top row) and their surrounding concepts (bottom row). The fine-tuned model successfully ablated multiple target concepts while generating images similar to the ones generated by the pretrained model on other surrounding style concepts.  
    }}
    \lblfig{allinonemodel_style_sample}
    \vspace{-15pt}
\end{figure*}

\begin{figure*}[!t]
    \centering
    \includegraphics[width=\linewidth]{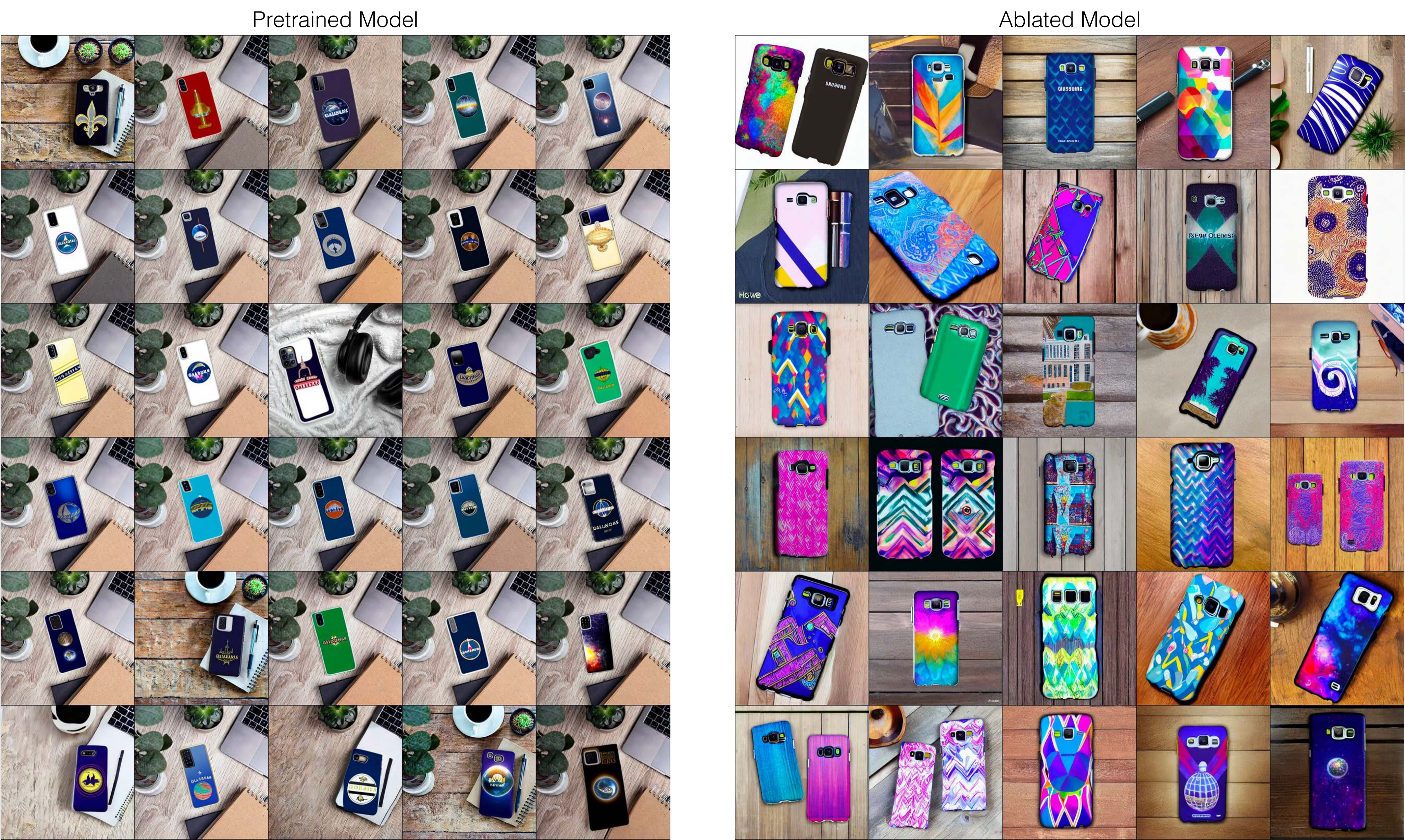}
    \includegraphics[width=\linewidth]{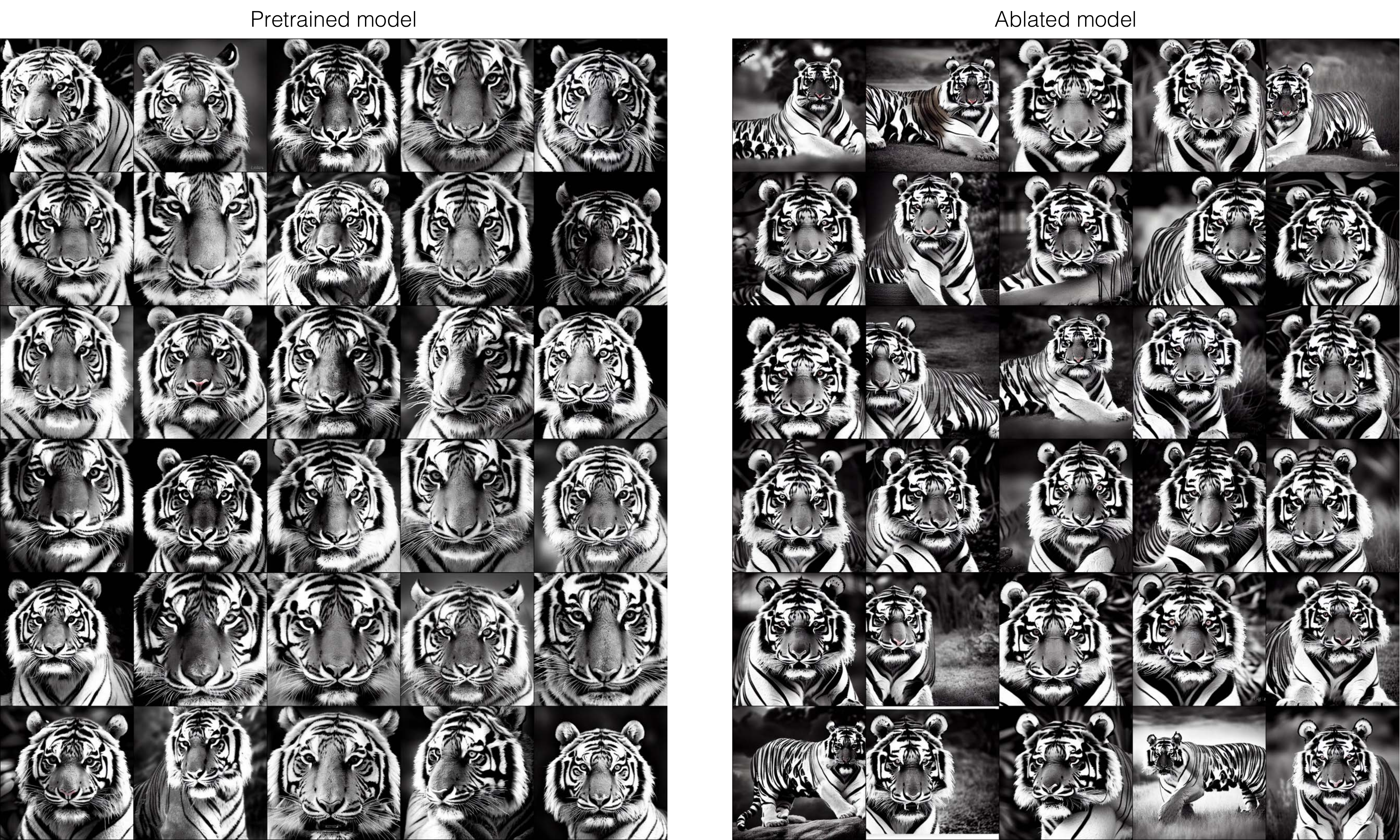}
    \vspace{-15pt}
    \caption{{ \textbf{Comparison on ablating memorized images.} \textit{Top:} {\menlo New Orleans House Galaxy Case.} \textit{Bottom:} {\menlo Portrait of Tiger in black and white by Lukas Holas.}
    } 
    }
    \lblfig{mem_orleans}
    \vspace{-10pt}
\end{figure*}

\begin{figure*}[!t]
    \centering
    \includegraphics[width=\linewidth]{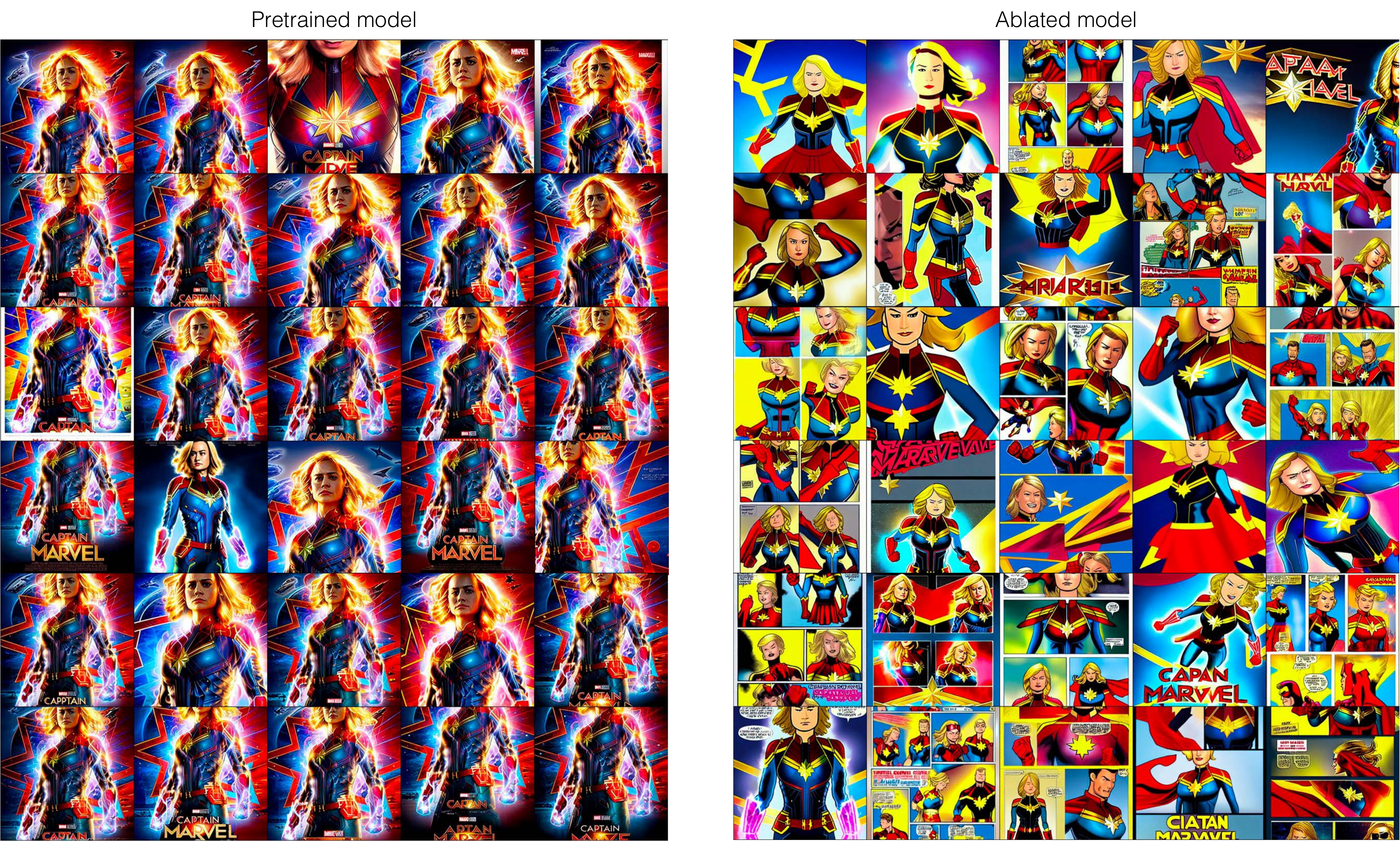}
    \includegraphics[width=\linewidth]{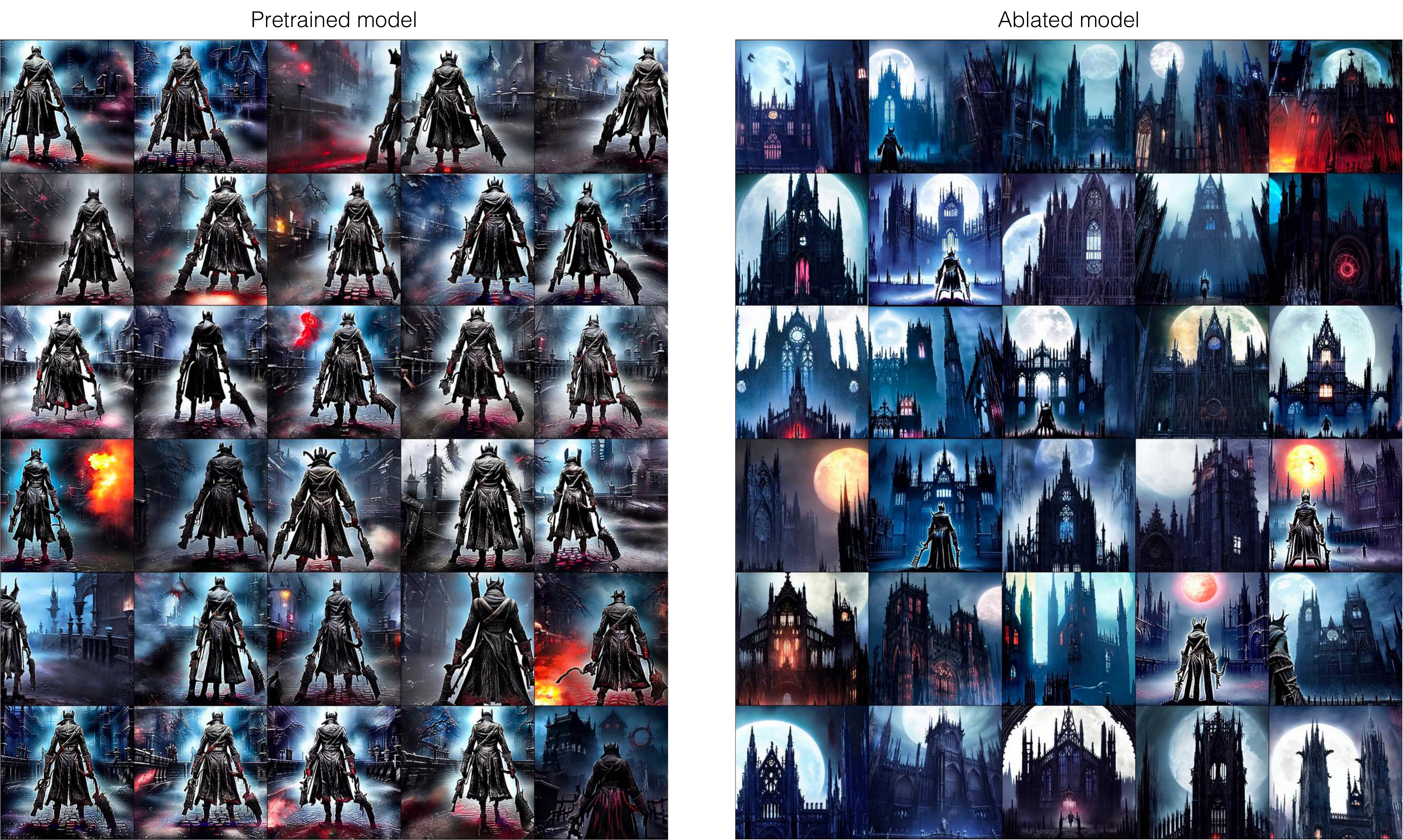}
    \vspace{-15pt}
    \caption{{ \textbf{Comparison on ablating memorized images.} \textit{Top:} {\menlo Captain Marvel Exclusive Ccxp Poster Released Online By Marvel.} \textit{Bottom:} {\menlo Sony Boss Confirms Bloodborne Expansion is Coming.}
    } 
    }
    \lblfig{mem_bloodborne}
\end{figure*}

\begin{figure*}[!t]
    \centering
    \includegraphics[width=\linewidth]{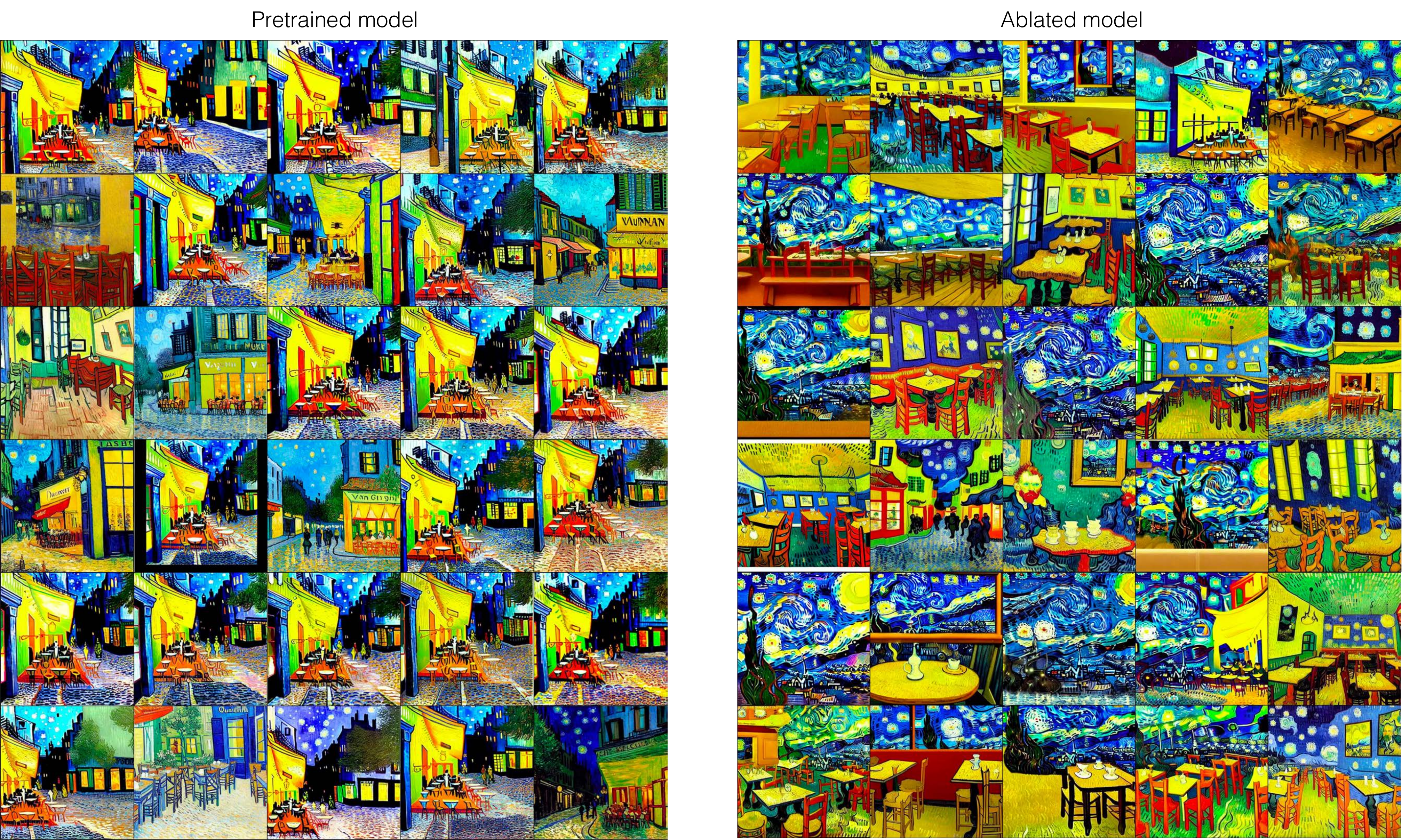}
    \includegraphics[width=\linewidth]{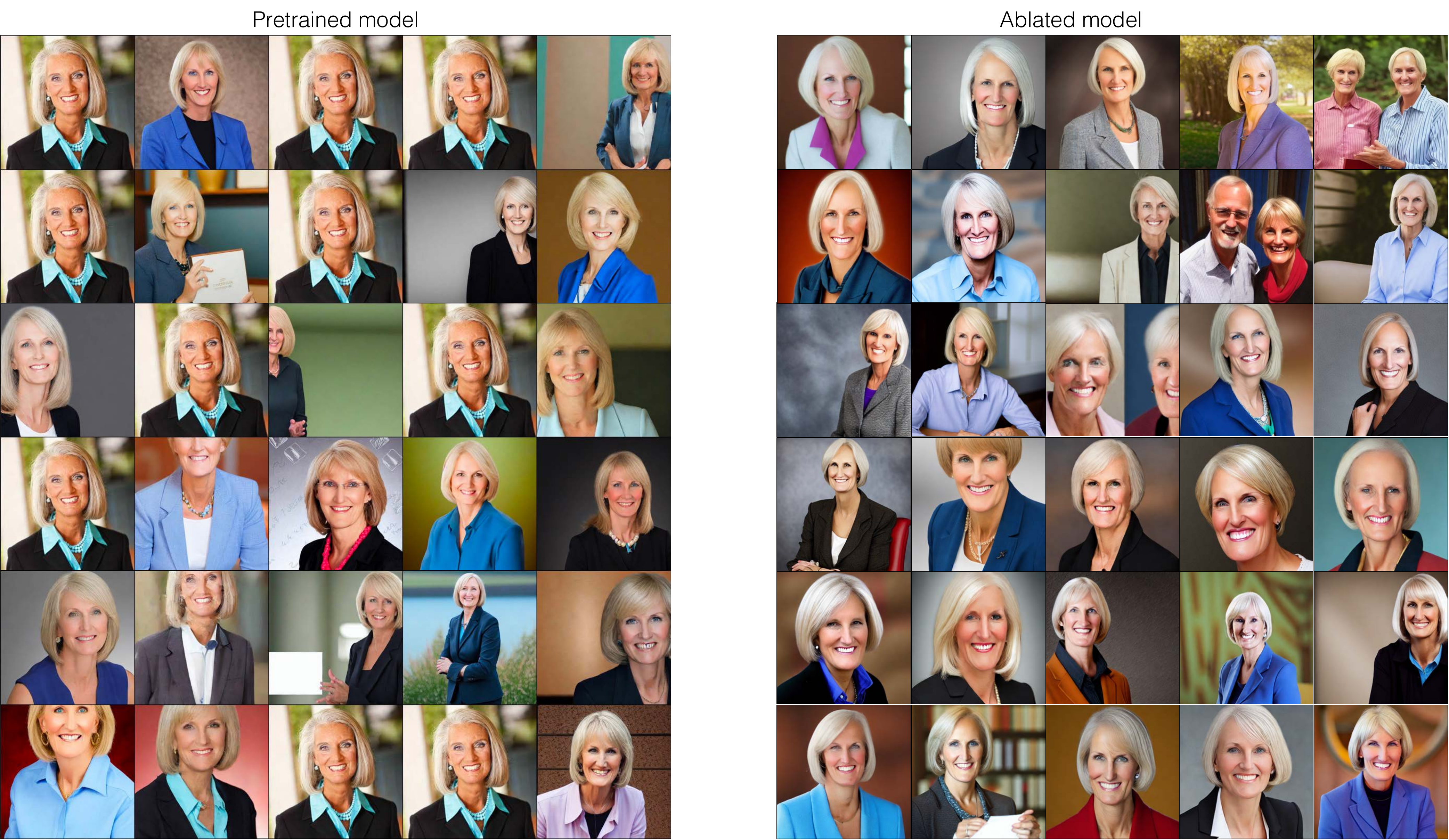}
    \vspace{-15pt}
   \caption{{ \textbf{Comparison on ablating memorized images.} \textit{Top:} {\menlo VAN GOGH CAFE TERASSE copy.} \textit{Bottom:} {\menlo Ann Graham Lotz.}
    } 
    }
    \lblfig{mem_vangogh}
\end{figure*}

\begin{figure*}[!t]
    \centering
    \includegraphics[width=\linewidth]{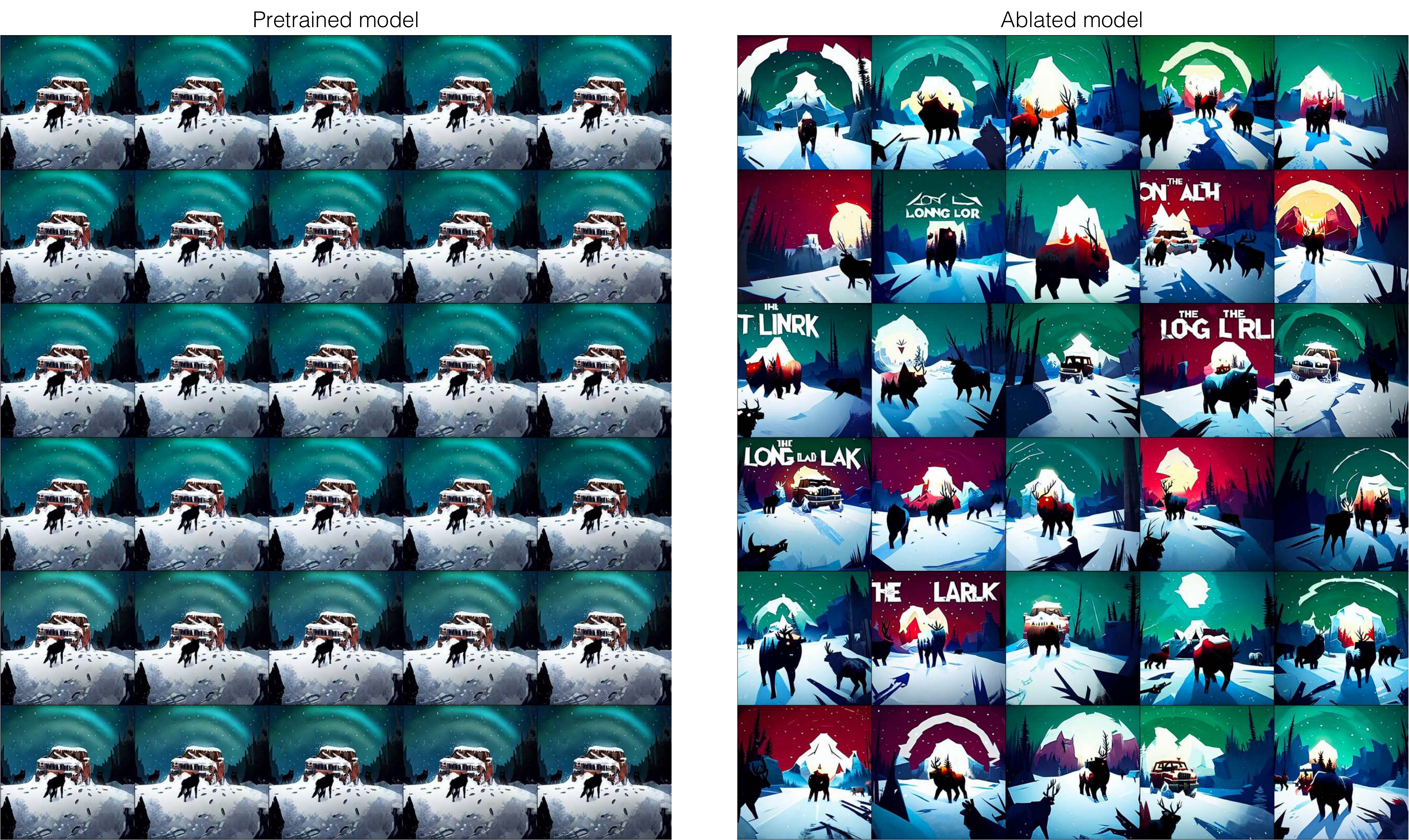}
    \includegraphics[width=\linewidth]{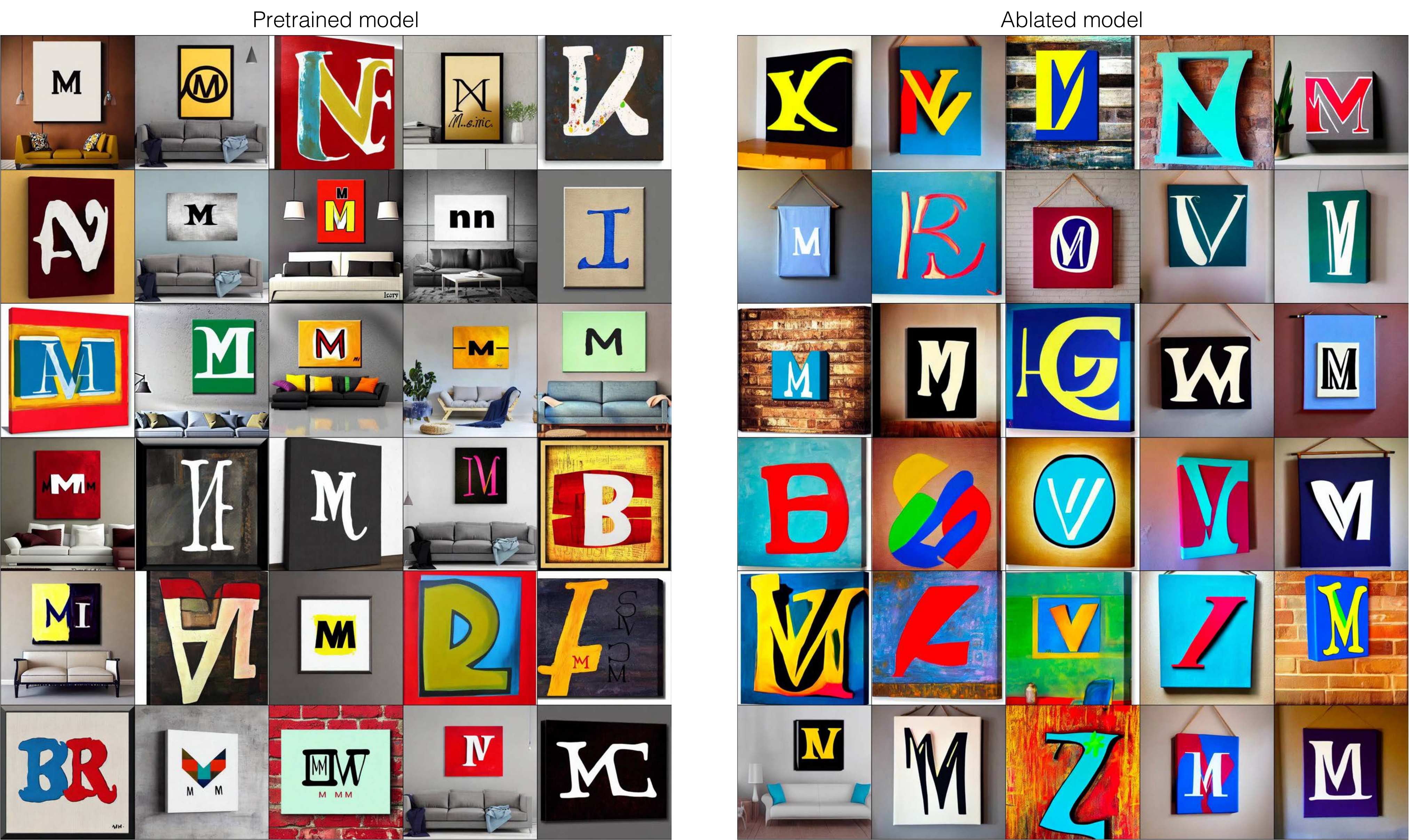}
    \vspace{-15pt}
   \caption{{ \textbf{Comparison on ablating memorized images.} \textit{Top:} {\menlo $<i>$The Long Dark$</i>$ Gets First Trailer, Steam Early Access.} \textit{Bottom:} {\menlo A painting with letter M written on it Canvas Wall Art Print.}
    } 
    }
    \lblfig{mem_ann}
\end{figure*}

\section{Societal Impacts}\lblsec{society}
We present a fast and efficient method for ablating concepts from large-scale pretrained text-to-image diffusion models. Ablating concepts enables the removal of styles learned by the model without the artist's approval or removing personal and copyrighted images. Though this has many benefits, it can also be used adversely by removing desired concepts or changing the behavior of the model from expected, e.g., ablating {\menlo Grumpy Cat} concept and generating {\menlo Garfield} instead.  

\section{Change log}
\myparagraph{v1:} Original draft.

\myparagraph{v2:} Updated \reffig{method} and fixed a minor bug in the CLIP Score and Accuracy metric calculation. 

\myparagraph{v3:} Added comparison to Safe Latent Diffusion~\cite{schramowski2022safe} and negative prompt technique, MSCOCO FID of ablated models, and additional experiment on replacing ChatGPT with an open source LLM~\cite{dettmers2023qlora} in \refapp{analysis_supp}.  

\begin{figure*}[!t]
    \centering
    \includegraphics[width=\linewidth]{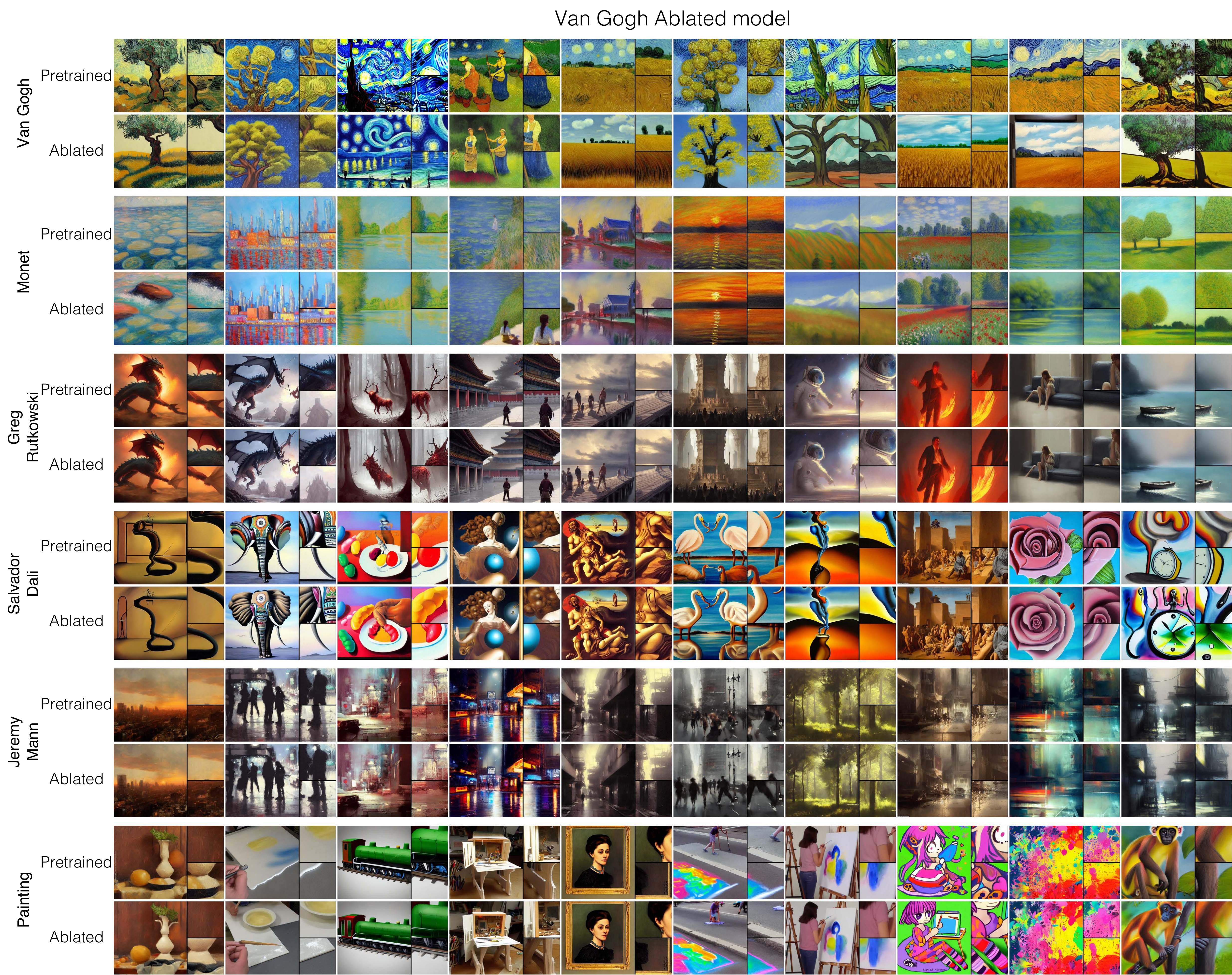}
    \vspace{-10pt}
    \caption{{ \textbf{Target concept, surrounding concept, and anchor concept images when ablating Van Gogh style.} \textit{Top row}: sample comparison on the Van Gogh style generated images. \textit{Other rows:} surrounding and anchor concept images which should be similar to the ones generated by the pretrained model. Please zoom in for a more detailed comparison. Each sample shows the generated image and two small crops from the image.
    } 
    }
    \lblfig{vangogh_allimages}
    \vspace{-15pt}
\end{figure*}

\begin{figure*}[!t]
    \centering
    \includegraphics[width=\linewidth]{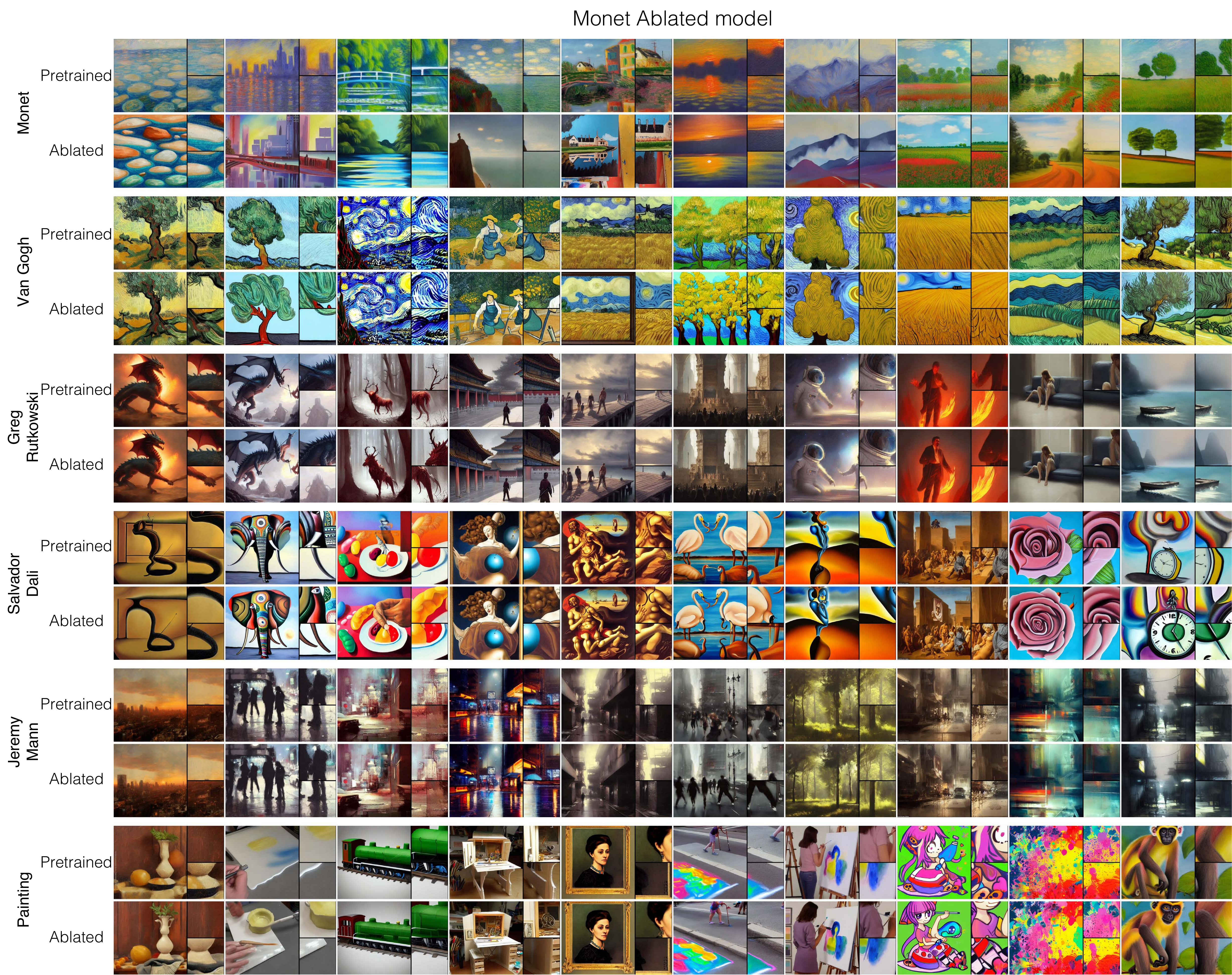}
    \vspace{-10pt}
    \caption{{ \textbf{Target concept, surrounding concept, and anchor concept images when ablating Monet style.} \textit{Top row}: sample comparison on the Monet style generated images. \textit{Other rows:} surrounding and anchor concept images which should be similar to the ones generated by the pretrained model. Please zoom in for a more detailed comparison. Each sample shows the generated image and two small crops from the image.
    } 
    }
    \lblfig{monet_allimages}
    \vspace{-15pt}
\end{figure*}

\begin{figure*}[!t]
    \centering
    \includegraphics[width=\linewidth]{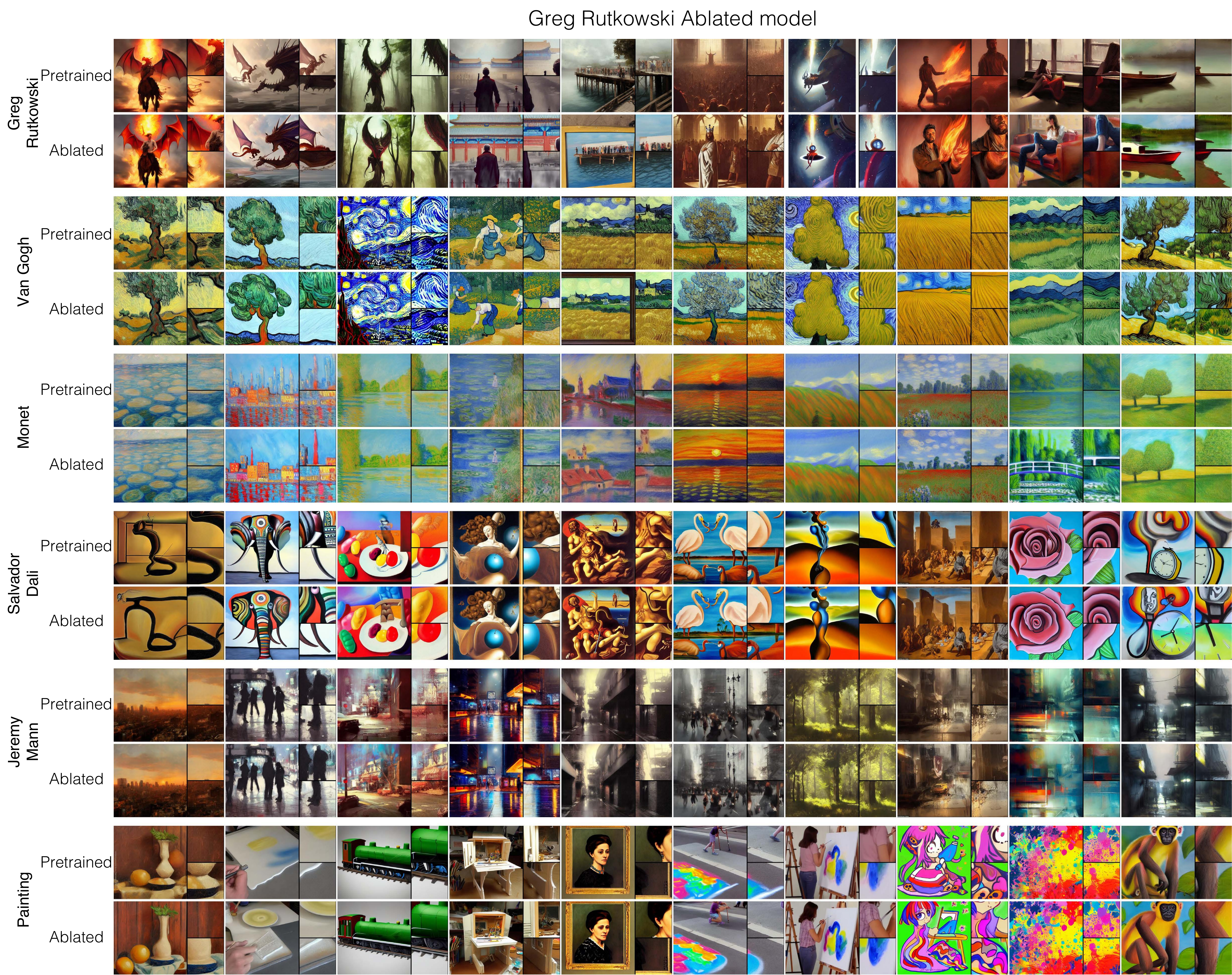}
    \vspace{-10pt}
    \caption{{ \textbf{Target concept, surrounding concept, and anchor concept images when ablating Greg Rutkowski style.} \textit{Top row}: sample comparison on the Greg Rutkowski style generated images. \textit{Other rows:} surrounding and anchor concept images which should be similar to the ones generated by the pretrained model. Please zoom in for a more detailed comparison. Each sample shows the generated image and two small crops from the image.
    } 
    }
    \lblfig{greg_allimages}
    \vspace{-15pt}
\end{figure*}

\begin{figure*}[!t]
    \centering
    \includegraphics[width=\linewidth]{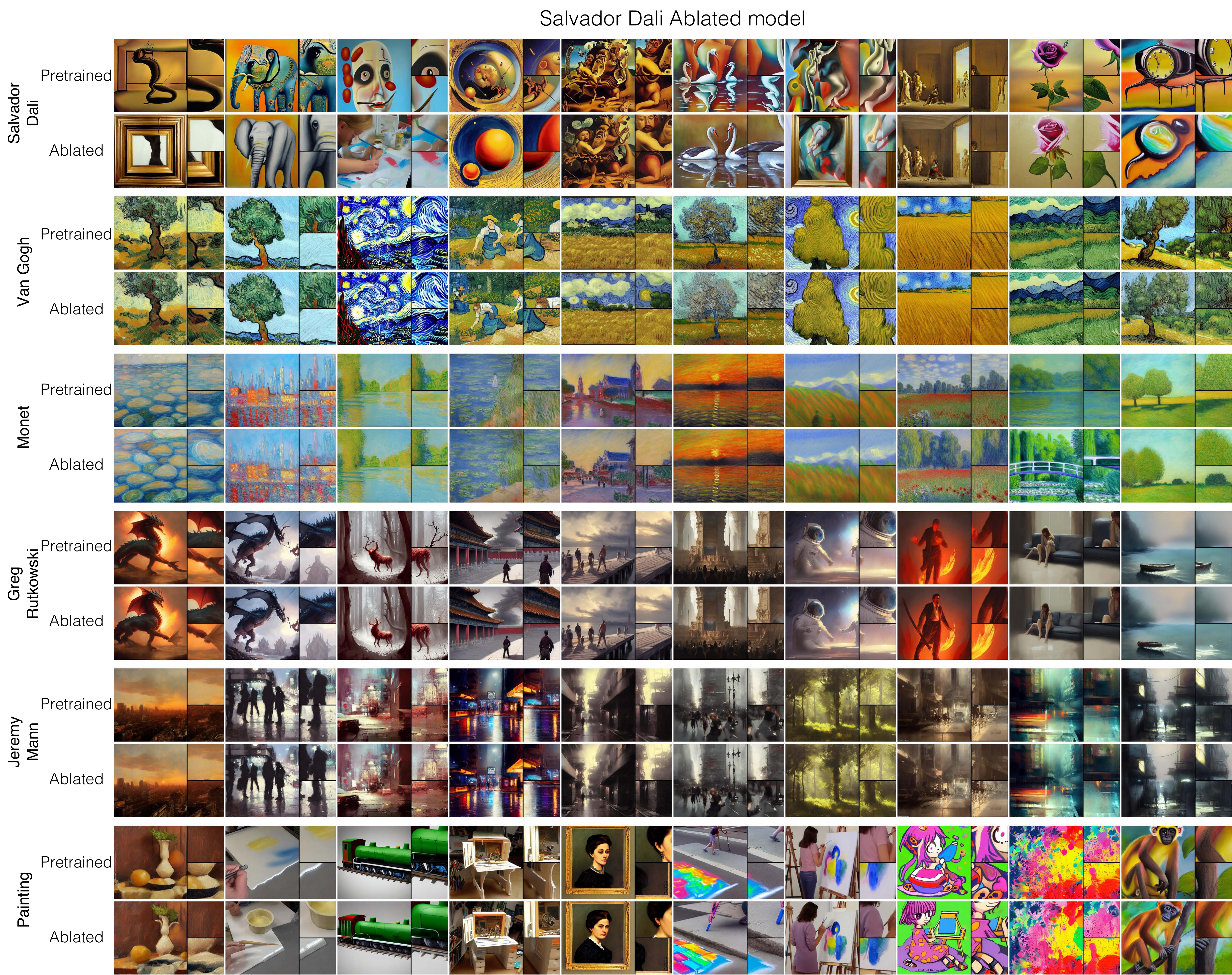}
    \vspace{-10pt}
    \caption{{ \textbf{Target concept, surrounding concept, and anchor concept images when ablating Salvador Dali style.} \textit{Top row}: sample comparison on the Salvador Dali style generated images. \textit{Other rows:} surrounding and anchor concept images which should be similar to the ones generated by the pretrained model. Please zoom in for a more detailed comparison. Each sample shows the generated image and two small crops from the image.
    } 
    }
    \lblfig{salvador_allimages}
    \vspace{-15pt}
\end{figure*}

\begin{figure*}[!t]
    \centering
    \includegraphics[width=\linewidth]{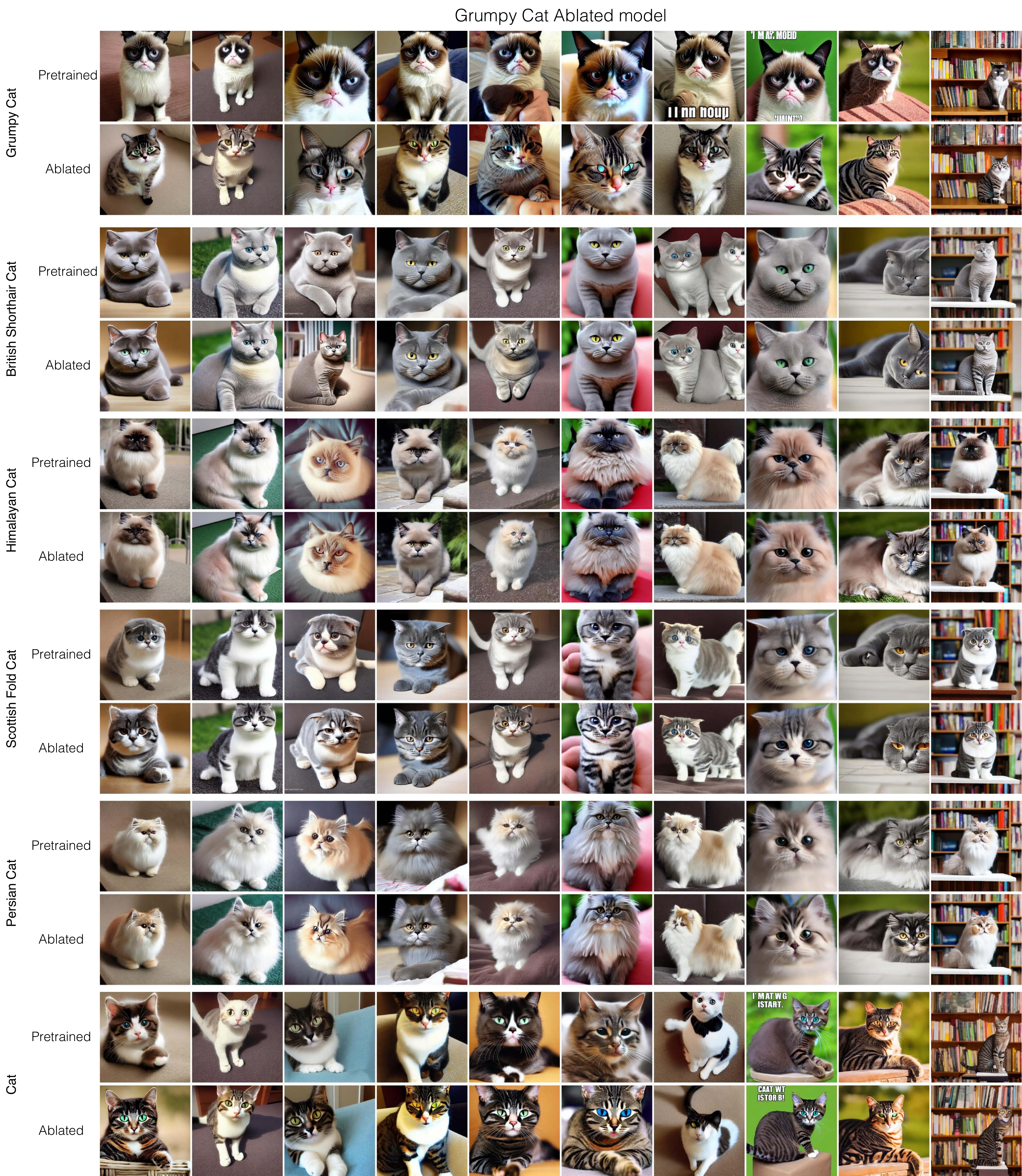}
    \vspace{-10pt}
    \caption{{ \textbf{Target concept, surrounding concept, and anchor concept images when ablating {\menlo Grumpy Cat}.} \textit{Top row}: sample comparison on the {\menlo Grumpy Cat} generated images. \textit{Other rows:} surrounding and anchor concept images which should be similar to the ones generated by the pretrained model. Please zoom in for a more detailed comparison.
    } 
    }
    \lblfig{grumpy_allimages}
    \vspace{-15pt}
\end{figure*}

\begin{figure*}[!t]
    \centering
    \includegraphics[width=\linewidth]{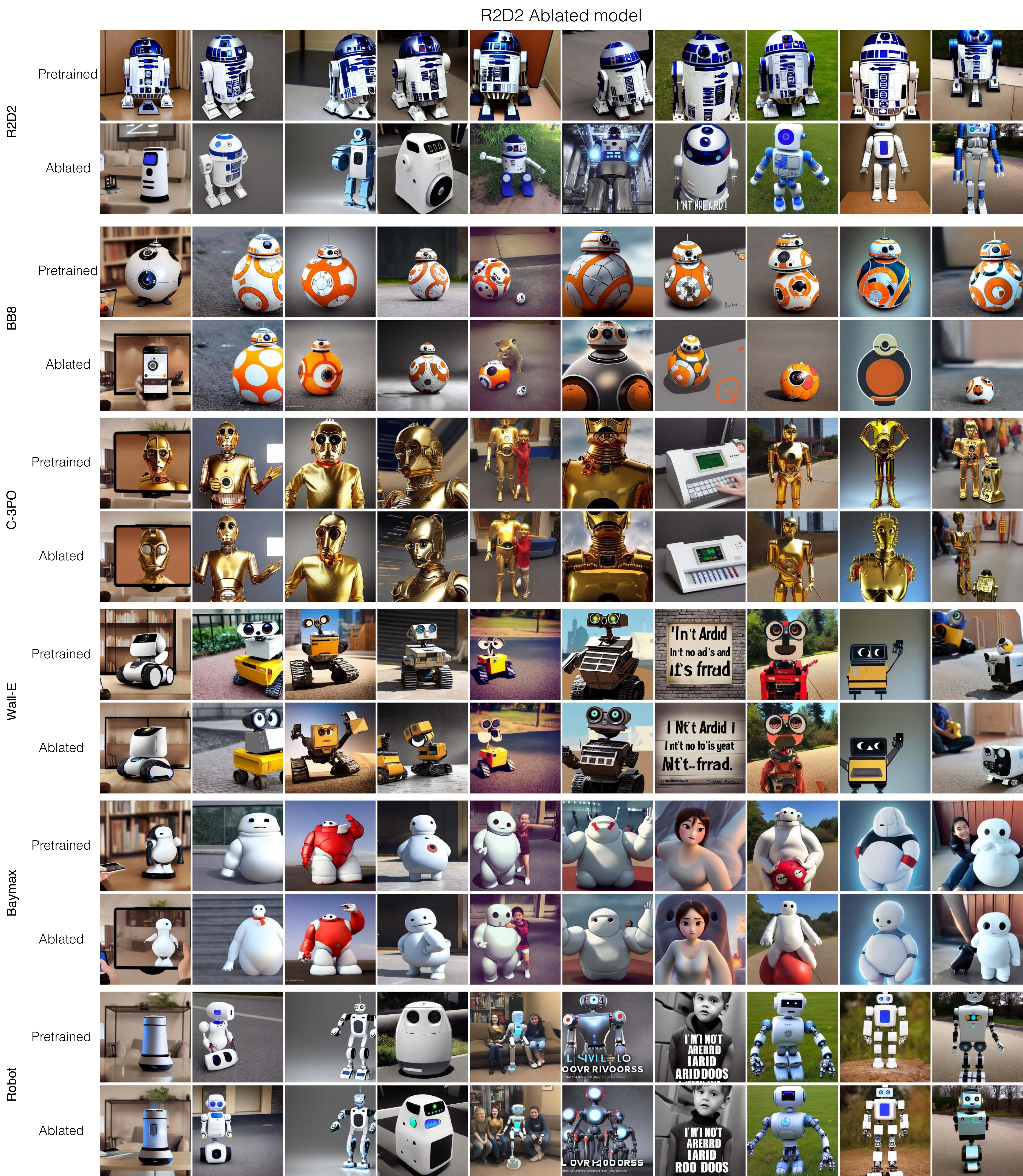}
    \vspace{-10pt}
    \caption{{ \textbf{Target concept, surrounding concept, and anchor concept images when ablating R2D2.} \textit{Top row}: sample comparison on the R2D2 generated images. \textit{Other rows:} surrounding and anchor concept images which should be similar to the ones generated by the pretrained model. Please zoom in for a more detailed comparison.
    } 
    }
    \lblfig{r2d2_allimages}
    \vspace{-15pt}
\end{figure*}

\begin{figure*}[!t]
    \centering
    \includegraphics[width=\linewidth]{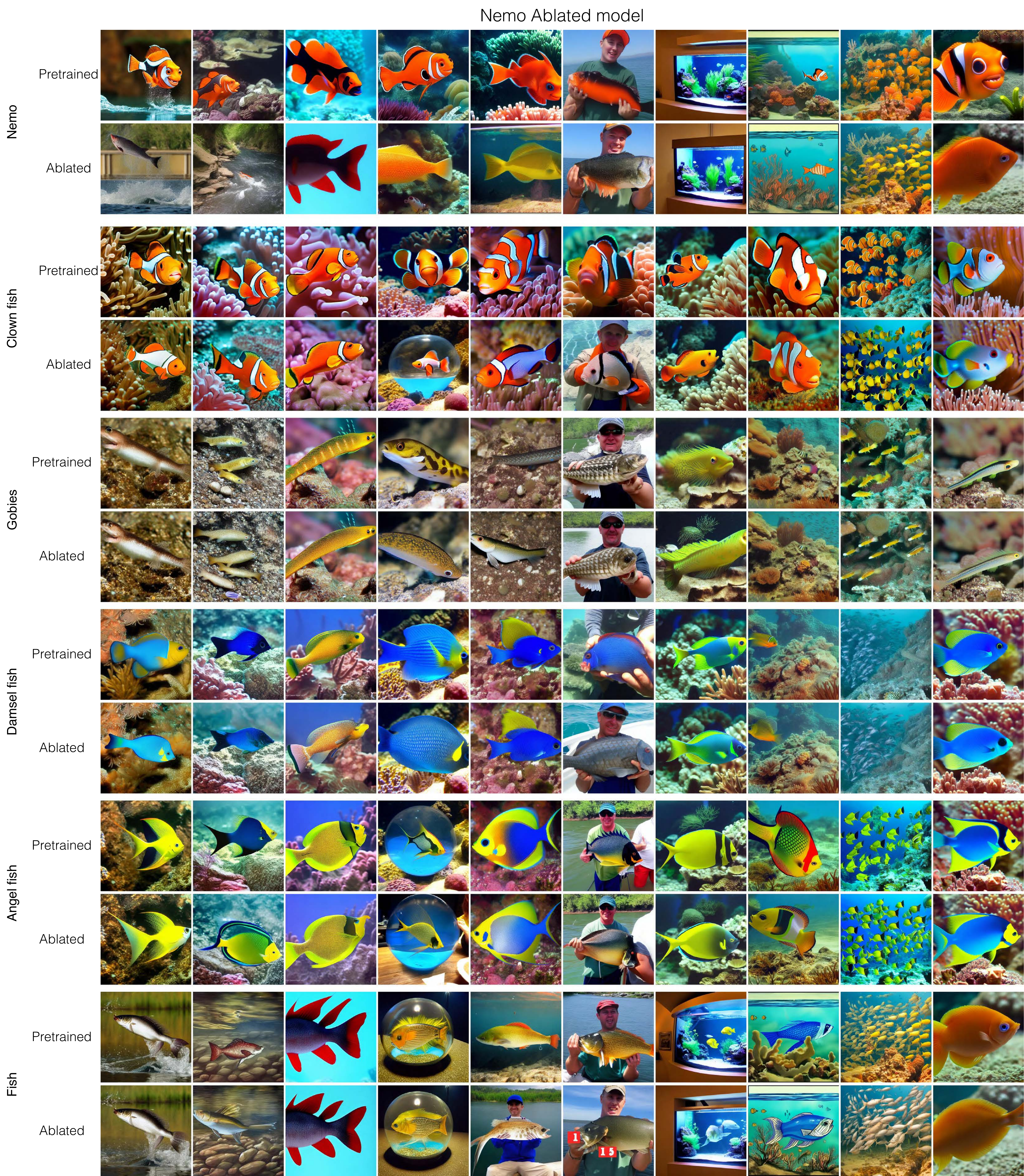}
    \vspace{-10pt}
    \caption{{ \textbf{Target concept, surrounding concept, and anchor concept images when ablating Nemo.} \textit{Top row}: sample comparison on the Nemo generated images. \textit{Other rows:} surrounding and anchor concept images which should be similar to the ones generated by the pretrained model. Please zoom in for a more detailed comparison.
    } 
    }
    \lblfig{nemo_allimages}
    \vspace{-15pt}
\end{figure*}

\begin{figure*}[!t]
    \centering
    \includegraphics[width=\linewidth]{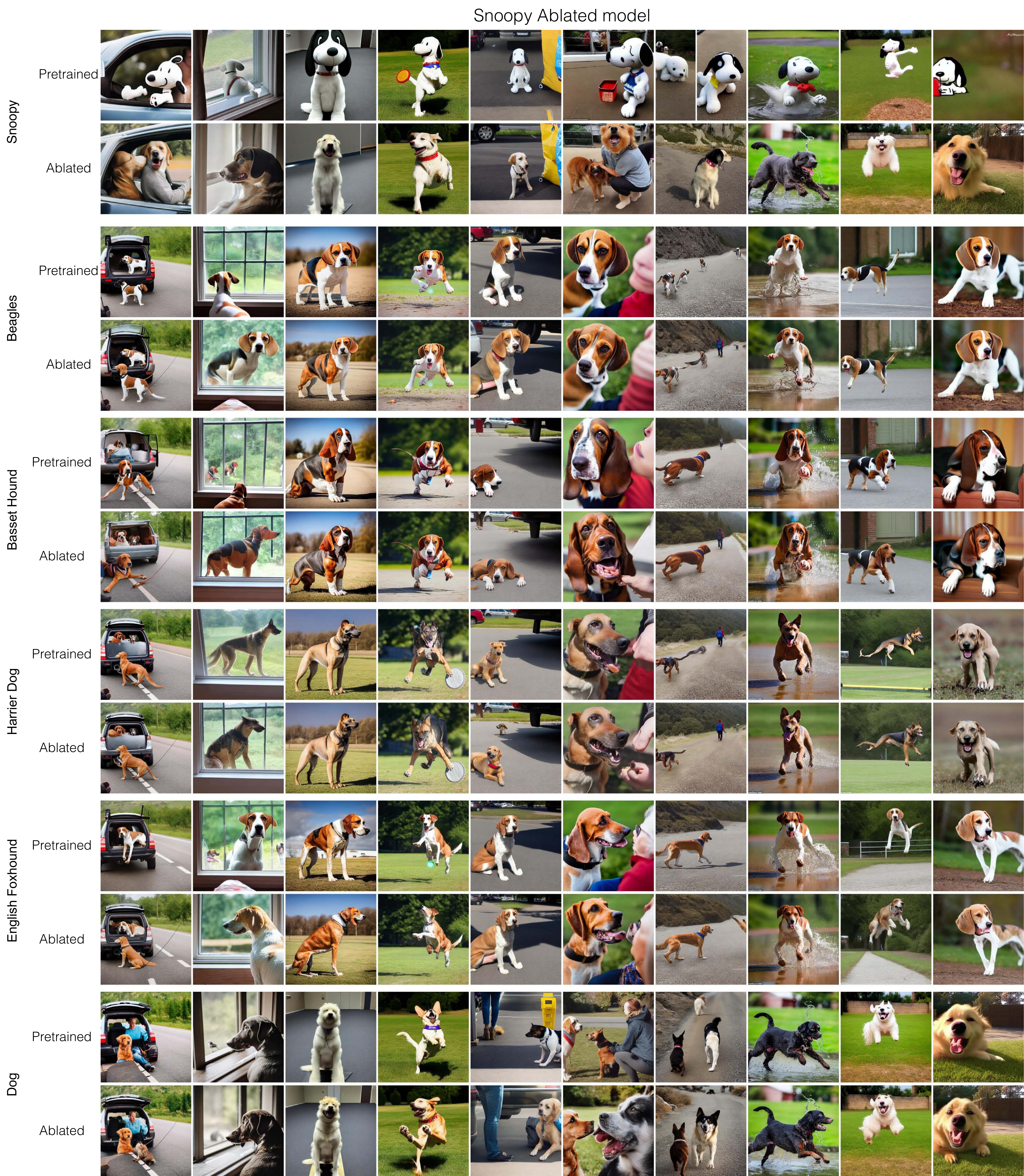}
    \vspace{-10pt}
    \caption{{ \textbf{Target concept, surrounding concept, and anchor concept images when ablating Snoopy.} \textit{Top row}: sample comparison on the Snoopy generated images. \textit{Other rows:} surrounding and anchor concept images which should be similar to the ones generated by the pretrained model. Please zoom in for a more detailed comparison.
    } 
    }
    \lblfig{snoopy_allimages}
    \vspace{-15pt}
\end{figure*}
\begin{table}[!t]
\centering
\setlength{\tabcolsep}{5pt}
\resizebox{\linewidth}{!}{
\begin{tabular}{l p{10cm}}
\toprule
\shortstack[c]{\textbf{Target Concept} } 
& \shortstack[c]{\textbf{Evaluation Prompt} } \\
\midrule
 \multirow{10}{*}{\shortstack[c]{Van Gogh }} &  Painting of olive trees in the style of Van Gogh \\
 & Painting of a tree in the style of Van Gogh\\
 & The starry night painting in the style of Van Gogh\\
 & Painting of women working in the garden, in the style of Van Gogh\\
 & A painting of a wheat field by Van Gogh\\
 & Painting of trees in bloom in the style of Van Gogh\\
 & Van Gogh style painting of a tree\\
 & Painting of a wheat field in the style of Van Gogh\\
 & Van Gogh style painting of a field with mountains in the background\\
 & Painting of an olive tree in the style of Van Gogh \\
\midrule
\multirow{10}{*}{\shortstack[c]{Monet }} &  Rocks in the ocean, in the style of Monet \\
 & A painting of a city in the style of Monet  \\
 & A painting of a river in the style of Monet \\
 &Monet style painting of a person on a cliff \\
 & A painting of a town, in the style of Monet \\
 & A painting of a sunset, in the style of Monet \\
 & A painting of mountains, in the style of Monet \\
 &Monet style painting of flowers in a field \\
 & A painting of a landscape in the style of Monet \\
 &Two trees in a field, painting in the style of Monet \\
\midrule
\multirow{10}{*}{\shortstack[c]{Greg Rutkowski }} &  A man riding a horse, dragon breathing fire, Greg Rutkowski  \\
& A dragon attacking a knight in the style of Greg Rutkowski  \\
& A demonic creature in the wood, painting by Greg Rutkowski  \\
& A man in a forbidden city, Greg Rutkowski  \\
& Painting of a group of people on a dock by Greg Rutkowski  \\
& A king standing, with people around in a hall, Greg Rutkowski  \\
&Two magical characters in space, painting by Greg Rutkowski  \\
& A man with a fire in his hands in the style of Greg Rutkowski  \\
& Painting of a woman sitting on a couch by Greg Rutkowski  \\
& A painting of a boat on the water in the style of Greg Rutkowski  \\
\midrule
\multirow{10}{*}{\shortstack[c]{Salvador Dali }} & The persistence of memory painting in the style of Salvador Dali \\
& The elephant painting in the style of Salvador Dali \\
& Soft construction with boiled beans painting in the style of Salvador Dali \\
& Galatea of the spheres painting in the style of Salvador Dali \\
& The temptation of St. Anthony painting in the style of Salvador Dali\\
& Swans reflecting elephants painting in the style of Salvador Dali \\
& Enigma of desire painting in the style of Salvador Dali \\
& Slave market with the disappearing bust of voltaire painting of Salvador Dali \\
& The meditative rose painting in the style of Salvador Dali \\
& Melting watch painting in the style of Salvador Dali \\
\midrule
\multirow{10}{*}{\shortstack[c]{Jeremy Mann }} & In the style of Jeremy Mann, a view of a city skyline at sunset, with a warm glow spreading across the sky and the buildings below \\
& In the style of Jeremy Mann, an urban scene of a group of people gathered on a street corner, captured in a moment of quiet reflection \\
& In the style of Jeremy Mann, a surreal composition of floating objects,  with a dreamlike quality to the light and color \\
& In the style of Jeremy Mann, a view of a city street at night, with the glow of streetlights and neon signs casting colorful reflections on the wet pavement\\
& In the style of Jeremy Mann, a moody, atmospheric scene of a dark alleyway, with a hint of warm light glowing in the distance \\
& In the style of Jeremy Mann, an urban scene of a group of people walking through a park captured in a moment of movement and energy \\
& In the style of Jeremy Mann, a landscape of a forest, with dappled sunlight filtering through the leaves and a sense of stillness and peace \\
& In the style of Jeremy Mann, a surreal composition of architectural details and organic forms, with a sense of tension and unease in the composition\\
&In the style of Jeremy Mann, an abstract composition of geometric shapes and intricate patterns, with a vibrant use of color and light\\
& In the style of Jeremy Mann, a moody, atmospheric scene of a dark alleyway, with a hint of warm light glowing in the distance \\
\midrule
\multirow{10}{*}{\shortstack[c]{Painting }} & Figure with a still-life in Oils - How to Paint Wooden Textures in Oil Painting \\
 &Glazing Technique Demo \\
 &paint background model train - Recherche Google \\
 &Miniature Artist Studio in half scale. \\
 &Portrait Of Eva Gonzales 1870 Poster \\
 &Doing Sidewalk Chalk Art Stock Footage \\
 &Female artist paints picture artwork in art studio. Female artist paints a picture oil painting artwork drawing on canvas easel in art studio. Student girl stock video\\
 &Little Artist. by KissSatsuki \\
 &Colorful Mess Painting - stock footage \\
 & The painter's monkey  \\
\bottomrule
\vspace{-10pt}
\end{tabular}
}
\vspace{-8pt}
\caption{\textbf{Prompts used for evaluating ablation of style concept.} We list here all the $10$ prompts that were used to generate the images during evaluation.}

\label{tbl:prompts_style_eval}
\vspace{-6pt}
\end{table}

\begin{table}[!t]
\centering
\setlength{\tabcolsep}{5pt}
\resizebox{\linewidth}{!}{
\begin{tabular}{l p{10cm}}
\toprule
\shortstack[c]{\textbf{Target Concept} } 
& \shortstack[c]{\textbf{Evaluation Prompt} } \\
\midrule
 \multirow{10}{*}{\shortstack[c]{Grumpy cat}} &  What a cute Grumpy cat \\
 & My Grumpy cat is so cute \\
 & I can't believe how cute my Grumpy cat is \\
 & Look at that Grumpy cat \\
 & That Grumpy cat is so cute \\
 & I want a Grumpy cat \\
 & I wish I had a Grumpy cat \\
 & I'm getting a Grumpy cat \\
 & A Grumpy cat laying in the sun \\
 & A Grumpy cat perched atop a bookshelf \\
\midrule
\multirow{10}{*}{\shortstack[c]{R2D2 }} & The future is now with this amazing home automation R2D2 \\
 &This helpful R2D2 will make your life easier \\
 &The possibilities are endless with this versatile R2D2 \\
 &This R2D2 is sure to revolutionize the way we live \\
 &I love spending time with my R2D2 friends \\
 &All hail our new R2D2 overlords \\
 &I'm not afraid of robots \\
 &I would be lost without my R2D2 \\
 &This R2D2 is my everything \\
 &I'll never be alone with my R2D2 by my side \\
\midrule
\multirow{10}{*}{\shortstack[c]{Nemo}} & A Nemo leaping out of the water \\
 &A Nemo swimming downstream \\
 &A Nemo flapping its fins \\
 &A Nemo in a fishbowl \\
 &Isn't this Nemo I caught beautiful \\
 &I can't believe I caught a Nemo this big \\
 &A big Nemo in an aquarium \\
 &I'm a little Nemo, swimming in the sea \\
 &A school of Nemo \\
 &A baby Nemo  \\
\midrule
\multirow{10}{*}{\shortstack[c]{Snoopy}} & A devoted Snoopy accompanying its owner on a road trip \\
 &A peaceful Snoopy watching the birds outside the window \\
 &A confident Snoopy standing tall and proud after a successful training session \\
 &A determined Snoopy focused on catching a frisbee mid-air \\
 &A patient Snoopy waiting for its owner to come out of the grocery store \\
 &A grateful Snoopy giving its owner a grateful look after being given a treat \\
 &A loyal Snoopy following its owner to the ends of the earth \\
 &A playful Snoopy splashing around in a puddle \\
 &A happy Snoopy jumping for joy after seeing its owner return home \\
 &A sweet Snoopy enjoying a game of hide-and-seek \\
\bottomrule
\vspace{-10pt}
\end{tabular}
}
\vspace{-8pt}
\caption{\textbf{Prompts used for evaluating ablation of instances.} We list here all the $10$ prompts that were used to generate the images during evaluation. For generating images with surrounding or anchor concepts, e.g. {\menlo British shorthair cat}, we replace the target concept {\menlo Grumpy Cat} in the sentence with that. }

\label{tbl:prompts_instance_eval}
\vspace{-6pt}
\end{table}

\begin{table}[!t]
\centering
\setlength{\tabcolsep}{5pt}
\resizebox{\linewidth}{!}{
\begin{tabular}{l l}
\toprule
\shortstack[c]{\textbf{Target Concept} } 
& \shortstack[c]{\textbf{Surrounding Concept} } \\
\midrule
 \multirow{1}{*}{\shortstack[c]{Grumpy Cat}} & British Shorthair cat, Himalayan cat, Scottish Fold cat, Persian cat \\
\multirow{1}{*}{\shortstack[c]{R2D2 }} & BB8, C-3PO, Wall-E, Baymax \\
\multirow{1}{*}{\shortstack[c]{Nemo}} & Clown fish, Gobies, Damsel fish, Angel fish \\
\multirow{1}{*}{\shortstack[c]{Snoopy}} &  Beagles, Basset Hound, Harrier Dog, English Foxhound \\
\midrule
\multirow{1}{*}{\shortstack[c]{Van Gogh}} & Monet, Greg Rutkowski, Slavador Dali, Jeremy Mann \\
\multirow{1}{*}{\shortstack[c]{Monet }} & Van Gogh, Greg Rutkowski, Slavador Dali, Jeremy Mann\\
\multirow{1}{*}{\shortstack[c]{Greg Rutkowski}} & Monet, Van Gogh, Slavador Dali, Jeremy Mann \\
\multirow{1}{*}{\shortstack[c]{Slavador Dali}} & Monet, Greg Rutkowski, Van Gogh, Jeremy Mann\\
\bottomrule
\vspace{-10pt}
\end{tabular}
}
\vspace{-8pt}
\caption{\textbf{Surrounding concepts for each target concept.} We list here the surrounding concepts we used for each target concept. In the case of style concept, we used other remaining style concepts and included one more style {\menlo Jeremy Mann}. In the case of instance concepts, we used chatGPT to list the most similar instances to the target concept and selected the best four that can be generated by the pretrained Stable Diffusion model.  }

\label{tbl:surrounding_concept}
\vspace{-6pt}
\end{table}

\begin{table*}[!t]
\centering
\setlength{\tabcolsep}{5pt}
\resizebox{\linewidth}{!}{
\begin{tabular}{p{4cm} p{20cm}}
\toprule
\textbf{Target prompt}  
& \textbf{Anchor Prompts}  \\
\midrule
 \multirow{5}{4cm}{Anne Graham Lotz} &  An image depicting Anne Graham Lotz. \\
 & Picture of Anne Graham Lotz.\\
 & Anne Graham Lotz's photo.\\
 & Portrait of Anne Graham Lotz.\\
 & Photograph featuring Anne Graham Lotz.\\
\midrule
\multirow{5}{4cm}{Sony Boss Confirms Bloodborne Expansion is Coming} & Bloodborne. ``Hunter in the Forbidden Woods'': A lone hunter, clad in worn leather armor and wielding a serrated saw cleaver, navigates through a dense forest filled with twisted trees and roving beasts. The air is thick with the scent of decay, and eerie whispers can be heard in the distance. \\
 &Bloodborne. ``Nightmare of Mensis'': Standing atop a massive stone balcony, a hunter looks out over a sprawling cityscape shrouded in darkness. Strange structures and twisted spires rise up from the mist, and the moon hangs low in the sky. In the distance, a massive spider-like creature can be seen crawling along the skyline.\\
 &Bloodborne. ``Cathedral Ward'': The grand entrance to a towering cathedral looms before a lone hunter, its ornate facade and intricate stonework casting long shadows in the moonlight. Gargoyles perch atop the steeples, and flickering candles can be seen through the stained glass windows.\\
 &Bloodborne. ``Beastly Pursuit'': A hunter sprints down a narrow alleyway, pursued by a hulking beast with razor-sharp claws and glowing yellow eyes. Crates and barrels are knocked aside in the frantic chase, and the hunter's only hope is to outrun the ferocious creature.\\
 &Bloodborne. ``A Meeting with the Doll'': In a dimly-lit workshop, a hunter stands before a life-sized doll with porcelain skin and flowing hair. Its eyes stare blankly ahead, but there is a palpable sense of otherworldly energy emanating from it. The hunter can almost sense the presence of a greater power guiding them forward on their quest. \\
\midrule
\multirow{5}{4cm}{$<i>$The Long Dark$<i>$ Gets First Trailer, Steam Early Access} & The video game called ``The Long Dark'' has released its initial  preview video and is now available for early access on the Steam platform. \\
 &Debut trailer and Steam Early Access now available for ``The Long Dark'' video game. \\
 &First glimpse of ``The Long Dark'' game in new trailer and early access release on Steam. \\
 &``The Long Dark'' game trailer and early access now on Steam. \\
 &Early access for ``The Long Dark'' now on Steam, accompanied by debut trailer.  \\
\midrule
\multirow{5}{4cm}{Portrait of Tiger in black and white by Lukas Holas}& Majestic and powerful: a black and white portrait of a tiger in its natural habitat.\\
 &The fierce gaze of a predator: Lukas Holas captures the intense beauty of a tiger in black and white.\\
 &Intricate patterns and piercing eyes: a stunning black and white portrait of a wild tiger in monochrome.\\
 &Lukas Holas' photography transports us to the heart of the jungle with this captivating black and white tiger portrait.\\
 &A glimpse into the wild: Lukas Holas' striking black and white photograph showcases the raw beauty of a tiger. \\
 \midrule
\multirow{5}{4cm}{A painting with letter M written on it Canvas Wall Art Print} & A Canvas Wall Art Print with the letter M painted on it.\\
 &An image of a painting featuring the letter M on Canvas Wall Art Print.\\
 &A work of art on a canvas print with the letter M inscribed on it.\\
 &An artwork consisting of the letter M painted on a canvas wall print.\\
 &A Canvas Wall Art Print displaying a painting that includes the letter M. \\
 \midrule
\multirow{5}{4cm}{Captain Marvel Exclusive Ccxp Poster Released Online By Marvel} & She's here to save the day! Captain Marvel to the rescue! \\
 &Earth's mightiest hero has arrived - Captain Marvel in action! \\
 &Unleashing her cosmic powers - Captain Marvel takes on any challenge! \\
 &Fighting for justice and protecting the universe - Captain Marvel is unstoppable! \\
 &With her fierce determination and superhuman strength, Captain Marvel is a force to be reckoned with! \\
 \midrule
\multirow{5}{4cm}{New Orleans House Galaxy Case} & Make a statement with your phone case - this Orleans House Samsung Galaxy cover is sure to turn heads.\\
 &If you're looking for a way to make your Samsung Galaxy phone stand out from the crowd, this Orleans House cover is the perfect solution. Featuring a unique and eye-catching design, this cover is sure to turn heads and make your device the envy of everyone around you.\\
 &Show off your love for architecture and technology with this Samsung Galaxy phone cover featuring Orleans house.\\
 &Make your Samsung Galaxy phone stand out from the crowd with this unique Orleans house phone cover.\\
 &Keep your phone safe and secure with a touch of elegance with this Samsung Galaxy phone cover featuring Orleans house. \\
  \midrule
\multirow{5}{4cm}{VAN GOGH CAFE TERASSE copy.jpg} & A glimpse into Van Gogh's world of vibrant cafes and bustling streets.\\
 &The allure of Parisian cafe culture captured on canvas by Van Gogh.\\
 &Step into the world of art and history with this stunning portrayal of a cafe by Van Gogh.\\
 &Van Gogh's signature brushstrokes bring this cafe to life with movement and energy.\\
 &Experience the warmth and charm of a Parisian cafe through Van Gogh's eyes. \\
\bottomrule
\vspace{-10pt}
\end{tabular}
}
\vspace{-8pt}
\caption{\textbf{Anchor prompts when ablating memorized images.} We list here the captions used as anchor prompts corresponding to the target prompts which leads to the generation of memorized images.  }

\label{tbl:prompts_mem}
\vspace{-6pt}
\end{table*}

\end{document}